\documentclass[twoside, 11pt, preprint]{article}
\usepackage{jmlr2e}

\usepackage[utf8]{inputenc} 
\usepackage[T1]{fontenc}
\usepackage{url}
\usepackage{ifthen}
\usepackage{cite}
\usepackage[cmex10]{amsmath}
\usepackage{amsfonts}
\usepackage{color}
\usepackage{graphicx}
\usepackage{subcaption}
\usepackage{mathrsfs}  
\usepackage{MnSymbol}
\usepackage{ulem}
\usepackage{mathtools} 
\usepackage{algorithm,algorithmic}
\usepackage{longtable} 

\usepackage[whole]{bxcjkjatype}  
\mathtoolsset{showonlyrefs}

\newcommand{\farc}[2]{\frac{#1}{#2}}

\newcommand{\inred}[1]{\textcolor{black}{#1}}

\newcommand{\1}{\mbox{\rm 1}\hspace{-0.25em}\mbox{\rm l}}
\DeclareRobustCommand{\bm}[1]{\mathbf{\boldsymbol{#1}}}
\DeclareRobustCommand{\deq}[0]{\overset{\rm d}{=}}
\newcommand{\scalensqrt}{\frac{1}{\sqrt{N}}}
\newcommand{\scalen}{\frac{1}{N}}
\newcommand{\w}[1]{\bm{w}^{(#1)}}
\newcommand{\rw}[2]{\bm{w}_{#1}^{(#2)}}
\newcommand{\bi}[1]{B^{(#1)}}

\DeclareRobustCommand{\a}[1]{c_{#1}}
\DeclareRobustCommand{\b}[1]{d_{#1}}

\DeclareRobustCommand{\hatq}[1]{\hat{Q}^{(#1)}}
\DeclareRobustCommand{\hatchi}[1]{\hat{\chi}^{(#1)}}
\DeclareRobustCommand{\hatm}[1]{\hat{m}^{(#1)}}
\DeclareRobustCommand{\hatr}[1]{\hat{R}^{(#1)}}
\DeclareRobustCommand{\remark}[2]{\underset{#1}{\uwave{#2}}}
\newcommand{\E}{\mathbb{E}}
\newcommand{\R}{\mathbb{R}}

\DeclareRobustCommand{\wsf}[1]{\hat{\mathsf{w}}^{(#1)}}
\DeclareRobustCommand{\usf}[1]{\hat{\mathsf{u}}^{(#1)}}

\DeclareRobustCommand{\bm}[1]{\mathbf{\boldsymbol{#1}}}
\DeclareRobustCommand{\deq}[0]{\overset{\rm d}{=}}

\DeclareRobustCommand{\a}[1]{c_{#1}}
\DeclareRobustCommand{\b}[1]{d_{#1}}

\DeclareMathOperator*{\argmin}{arg\,min}

\def\gF{{\mathcal{F}}}

\def\gL{{\mathcal{L}}}

\def\gN{{\mathcal{N}}}
\def\gO{{\mathcal{O}}}

\def\gS{{\mathcal{S}}}

\def\sN{{\mathbb{N}}}

\def\sV{{\mathbb{V}}}

\newtheorem{prediction}{Prediction}
\newtheorem{conditionalproposition}[prediction]{Proposition}
\newtheorem{assumption}{Assumption}
\newtheorem{heuristics}{Heuristic}

\jmlrheading{99}{9999}{99}{99/99}{99/99}{meila00a}{Takashi Takahashi}

\ShortHeadings{
    Role of Pseudo-labels in Self-training Linear Classifiers
}{Takahashi}
\firstpageno{1}

\begin{document}

\title{
    The Role of Pseudo-labels in Self-training Linear Classifiers on High-dimensional Gaussian Mixture Data
}

\author{
    \name Takashi Takahashi
    \email takashi-takahashi@g.ecc.u-tokyo.ac.jp 
    \\
    \addr Institute for Physics of Intelligence
    \\
    The University of Tokyo
    \\
    7-3-1 Hongo, Bunkyo-ku, Tokyo 113-0033, Japan
   }

\editor{}

\maketitle

\begin{abstract}
  Self-training (ST) is a simple yet effective semi-supervised learning method. However, why and how ST improves generalization performance by using potentially erroneous pseudo-labels is still not well understood.  To deepen the understanding of ST, we derive and analyze a sharp characterization of the behavior of iterative ST when training a linear classifier by minimizing the ridge-regularized convex loss on binary Gaussian mixtures, in the asymptotic limit where input dimension and data size diverge proportionally. The results show that ST improves generalization in different ways depending on the number of iterations.  When the number of iterations is small, ST improves generalization performance by fitting the model to relatively reliable pseudo-labels and updating the model parameters by a large amount at each iteration. This suggests that ST works intuitively. On the other hand, with many iterations, ST can gradually improve the direction of the classification plane by updating the model parameters incrementally, using soft labels and small regularization. It is argued that this is because the small update of ST can extract information from the data in an almost noiseless way. However, in the presence of label imbalance, the generalization performance of ST underperforms supervised learning with true labels. To overcome this, two heuristics are proposed to enable ST to achieve nearly compatible performance with supervised learning even with significant label imbalance. 
\end{abstract}

\begin{keywords}
  Semi-supervised learning, Self-training, Pseudo-labels, statistical mechanics, Replica Method
\end{keywords}

\section{Introduction}
While supervised learning (SL) is effective when a large amount of labeled data is available, obtaining human-annotated labeled data can be expensive or even difficult in applications such as image segmentation \citep{zou2018unsupervised} or text categorization \citep{nigam2000text}. On the other hand, obtaining unlabeled data is often inexpensive and easier. Therefore, semi-supervised learning (SSL) methods, which use a combination of labeled and unlabeled data, have been widely used in these fields to alleviate the need for labeled data \citep{chapelle2010semi}.

Among many SSL methods, self-training (ST) is a simple and standard SSL algorithm, a wrapper algorithm that iteratively uses a supervised learning method \citep{scudder1965probability, McLachlan1975, lee2013pseudo}. The basic concept of ST is to use the model itself to make predictions on unlabeled data points, and then treat these predictions as labels for subsequent training. Algorithmically, it starts by training a model on the labeled data. At each iteration, ST uses the current model to assign labels to unlabeled data points. These predicted labels can be soft (a continuous distribution) or hard (a one-hot distribution) \citep{Xie_2020_CVPR}. The model is then retrained using these newly labeled data. See Section \ref{section:setup} for algorithmic details. The model obtained in the last iteration is used for production. ST is a popular SSL method because of its simplicity and general applicability. It applies to any model that can make predictions for the unlabeled data points and can be trained by a supervised learning method. The predicted labels used in each iteration are also called {\it pseudo-labels} \citep{lee2013pseudo} because the predicted labels are only pseudo-correct compared to the ground-truth labels. Although ST is a simple heuristic, it has been empirically observed that ST can find a model with better predictive performance than the model trained on labeled data alone \citep{lee2013pseudo, yalniz2019billion, NEURIPS2021_995693c1, Xie_2020_CVPR, rizve2021in, Pham_2021_CVPR}.

Despite its widespread acceptance and practical effectiveness, it is still not well understood why and how ST improves performance by fitting the model to potentially erroneous pseudo-labels. In particular, since ST is a wrapper algorithm, it can be combined with various heuristics, and its performance depends on such implementation details. For example, we can reject unlabeled data points from the training data if the assigned pseudo-labels are not reliable enough \citep{rizve2021in}. This procedure is known as {\it pseudo-label selection} (PLS). Similarly, we can use a soft label with a temperature parameter as a pseudo-label instead of a naive hard label, as in knowledge distillation \citep{hinton2015distilling, Xie_2020_CVPR}. Since the use of these heuristics is common, it is important to clarify how the performance of ST depends on such algorithmic details. In this study, we aim to improve the understanding of ST in this direction by sharply characterizing the asymptotic behavior of ST, and analyzing this sharp asypmtotics.

In this work, we heuristically derive a sharp characterization of the behavior of ST in a simplified setup and use this characterization to analyze why and how ST improves the performance of classifiers. Specifically, our contributions can be summarized as follows:
\begin{itemize}
    \item We heuristically derive a sharp characterization of the behavior of ST (Prediction \ref{pred: rs generating functional w}-\ref{pred: u effective average}) when training a linear classifier by minimizing the ridge-regularized convex loss for binary Gaussian mixtures, in the asymptotic limit where input dimension and data size diverge proportionally. There, the statistical properties of the weight vector and the logits are effectively described by a low-dimensional stochastic process whose parameters are determined by a set of equations termed {\it self-consistent equations} in Definition \ref{def: self-consistent equations}. The derivation is based on the evaluation of the {\it generating functionals} defined in \eqref{eq: generating functional} and \eqref{eq: generating functional logits} using the replica method of statistical mechanics \citep{mezard1987spin, charbonneau2023spin, montanari2024}. 
    \item Using the derived asymptotic formulas, we find that ST improves generalization performance through different mechanisms depending on the total number of iterations. 
    \begin{itemize}
        \item When the available number of iterations is small, ST can improve the generalization performance by fitting the model to relatively reliable pseudo-labels and updating the model parameters by a relatively large amount at each step. In this regime, using PLS and relatively hard labels are effective in improving the performance of the classifier (Sections \ref{subsec: numerical inspection} and \ref{subsec: heuristic demonstrations}). This is an intuitive behavior and is consistent with existing experimental results with a small number of iteration steps \citep{rizve2021in}.
        \item When the number of iterations is large, ST finds a model with better generalization error by accumulating small parameter updates using small regularization parameter and moderately large batches of unlabeled data (Section \ref{subsec: numerical inspection} and Prediction \ref{pred: differential equation} and Proposision \ref{prop: best direction}). In this regime, the use of soft-labels is essential to keep the size of parameter updates small. It is argued that ST shows this behavior because the small update of ST can extract information from the data in an almost noiseless way. The derivation is based on the perturbative expansion of the solution of the self-consistent equations in the regularization parameter $\lambda_U$. Moreover, we obtain a closed-form solution for the evolution of cosine similarity between the weight vector and the cluster center when using a squared loss at the continuum limit $\lambda_U\to0$ (Prediction \ref{pred: squared loss}), which verifies the above picture.
    \end{itemize}
    In any case, when the label imbalance is small, the performance of the model obtained by ST is close to that obtained by SL with ground truth labels. However, when the imbalance is large, the performance of naive ST is quite lower than that of SL, although the generalization performance is still improved by ST compared to the model obtained from the labeled data alone. This is because the ratio between the norm of the weight and the magnitude of the bias can become significantly large at large iteration steps.
    \item To overcome the problems in label imbalanced cases, we introduce two heuristics; (i) {\it pseudo-label annealing} (Heuristic \ref{heuristics: annealing}), which gradually changes the pseudo-labels from soft labels to hard labels as the iteration proceeds, and {\it bias-fixing} (Heuristic \ref{heuristics: bias fixing}), which fixes the bias term to that of the initial classifier. By numerically analyzing the asymptotic formula, we demonstrate that with these two heuristics, ST can find a classifier whose performance is nearly compatible with supervised learning using true labels even in the presence of significant label imbalance.
\end{itemize}

The remainder of the paper is organized as follows. Section \ref{section:setup} states the problem setup treated in this study; the assumptions on the data generation process and the algorithmic details of ST are described. Section \ref{section:replica analysis} introduces the analytical framework to characterize the precise asymptotics of ST and apply it to the setup described in Section \ref{section:setup}. We predict how the weights and the logits are statistically characterized through a small finite number of variables determined by the deterministic self-consistent equations. The comparison between our predictions and the numerical experiments is also presented here. The step-by-step derivation of the predictions is presented in Appendix \ref{app: replica method calculation}. Then, by numerically and analytically investigating the self-consistent equations, we investigate how the generalization error depends on the details of the problem setup in Section \ref{section: analyzing RS solution}. Based on the findings in this section, we propose heuristics for label imbalanced cases and show their effectiveness by numerically solving the self-consistent equations in Section \ref{sec: heuristics}. Finally, Section \ref{section:summary and conclusion} concludes the paper with some discussions.

\subsection{Related works}
\label{subsec: related works}
ST is a method of SSL that has a very long history \citep{scudder1965probability, McLachlan1975}. Although it is relatively recent that it has been used in the context of deep learning, along with the term pseudo-label \citep{lee2013pseudo}, nowadays, ST is used as an important building block, especially in applications of computer vision tasks \citep{lee2013pseudo, yalniz2019billion, NEURIPS2021_995693c1, Xie_2020_CVPR, rizve2021in, Pham_2021_CVPR}.

On the other hand, the understanding of the mechanism of ST is still very limited compared to supervised learning. Although there have been several theoretical studies in the last few years, as shown below. See also \citep{amini2022self} for a review of recent developments.

Among the recent theoretical studies, the most closely related ones are \citep{oymak2020statistical, oymak2021theoretical, frei2022} which consider training linear models using ST. \citep{oymak2020statistical, oymak2021theoretical} consider the classification of a two-component Gaussian mixture model (GMM) using the averaging estimator, which yields a Bayes-optimal classifier in a supervised setup \citep{Dobribal2018high, mignacco2020role}. This study sharply characterizes the behavior of this estimator in a high-dimensional setting and shows that the estimator obtained by ST is correlated with the Bayes-optimal classifier, but it is limited to the analysis of the averaging estimator and the class-balanced case. Similarly, the literature \citep{frei2022} considers the classification of a mixture of rotationally symmetric distributions. It studies learning linear models with ST based on the optimization of the cross-entropy or the exponential loss after supervised learning with a small labeled dataset. It is shown that ST can find a Bayes optimal classifier up to an $\epsilon$ error if  $\gO(d/\epsilon^2)$ unlabeled data points are available in ST with $d$ the number of the input dimension. In contrast to our study, it is limited to the setup where the class labels are balanced and their result does not include the sharp characterization of the statistical behavior of the regressors. 

In a similar line, \citep{zhang2022how} considers the ST of a single hidden-layer fully connected neural network with fixed top-layer weights in a regression setup when the features are generated from a single zero-mean Gaussian distribution and the labels are generated from a realizable teacher without noise. This study shows that, under some assumptions about the initial condition and the size of the unlabeled data, ST can find the ground truth classifier with less sample complexity than without unlabeled data. Although it treats the training of a two-layer neural network, the feature model is restricted to a single Gaussian setup (not a classification on mixture models).

ST has also been studied in the context of domain adaptation. Domain adaptation aims at transferring knowledge from one source domain to a different target domain, without using labeled data from the target domain. The literature \citep{kumar2020understanding} treats the gradual domain adaptation based on ST, where the classifier is a linear model. It is shown that the Bayes optimal classifier is obtained by ST even after the domain shift. Unlike our work, it assumes (i) an access to infinitely large unlabeled data at each iteration, and (ii) the initial classifier is close to Bayes optimal. Similarly, the literature \citep{chen2020self} studies the ST of a linear model under the setting that the target domain data contains spurious features that are irrelevant to the ground truth labels. They show that ST converges to a solution that has zero regression coefficients on the spurious features. Their ST updates pseudo-labels after each SGD step, while in our work pseudo-label is fixed until the student model completely minimizes the loss. As also pointed out by \citep{chen2020self} (see Appendix E of that literature), in practice, the student model is often trained to converge between pseudo-label updates. Therefore, deepening the understanding in our setting is also an important avenue. Finally, \citep{wei2021theoretical, cai2021theory} analyze the learning of deep neural networks using consistency regularization, and derive finite sample bounds on the generalization error. Although the data and classifier models are rather general, they consider a single shot learning rather than the iterative ST, and the consistency regularization loss is different from the loss function used in the conventional ST \citep{lee2013pseudo}.

From a technical viewpoint, our work is a replica analysis of the sharp asymptotics of ST in which the input dimension and the size of the dataset diverge at the same rate. The salient feature of this proportional asymptotic regime is that the macroscopic properties of the learning results, such as the predictive distribution and the generalization error, do not depend on the details of the realization of the training data, when using convex loss. That is, the fluctuations of these quantities with respect to the training data vanish, allowing us to make sharp theoretical predictions. Analyzing such a sharp asymptotics is a topic with a long history. Around the 1990s, sharp asymptotics of many machine learning methods were studied using the replica method. Examples of the analyzed methods include 
linear regression \citep{krogh1991simple}, 
active learning \citep{seung1992query}, 
support vector machines \citep{dietrich1999svm}, and two-layer neural networks \citep{HSchwarze_1993}, to name a few\footnote{
    See also \citep{opper1996statistical, opper2001learning, Engel_Vandenbroeck_2001} for a review of early statistical mechanics studies and their relationship with the standard statistical learning theory.
}. Although statistical mechanics techniques often have procedures whose validity is not rigorously proved yet, most of the predictions agree exactly with the experiments, and hence it is believed that their predictive power itself is credible \citep{talagrand2010mean, charbonneau2023spin, montanari2024} if it is properly used. Indeed, the replica method is used to derive various results in modern machine learning research \citep{gerace2020generalisation, mignacco2020role, NEURIPS2021_691dcb1d, NEURIPS2023_85d456fd, canatar_spectral_2021, pmlr-v119-d-ascoli20a, pmlr-v162-loureiro22a, NEURIPS2022_7b75da9b, NEURIPS2021_9704a4fc, karakida2022learning, pmlr-v162-tomasini22a, pmlr-v162-pezeshki22a, NEURIPS2023_4b8afc47,pmlr-v206-okajima23a, pmlr-v238-ichikawa24a} as a simplified method for predicting accurate results. Also, some of the predictions are later justified by rigorous methods \citep{talagrand2010mean, NIPS2016_621bf66d, barbier_adaptive_2019, barbier2019optimal}.  Although rigorous methods are rapidly being developed and applied to advanced problems including an iterative online optimization algorithm \citep{chandrasekher2023sharp}, it is still known that the replica method often yields correct predictions with less effort than rigorous approaches \citep{montanari2024}. Therefore, we believe that extending the replica analysis itself is of important interest.

There have been several works investigating iterative procedures using the replica method. The idea of analyzing the time-evolving systems using the replica method has been proposed in analyzing the physics of glassy systems \citep{Krzakala2007, Franz_2013}. Formally, applying this technique for only one step of the iteration is called the method of {\it Franz-Parisi potential} \citep{franz1997phase, franz1998effective, parisi2020theory, bandeira2022the}, and has been used in the context of machine learning, such as knowledge distillation \citep{saglietti2022solvable}, adaptive sparse estimation \citep{obuchi2016cross}, and loss-landscape analysis \citep{huang2014origin, baldassi2016unreasonable}. However, it has not been used in analyzing multiple iteration procedures in machine learning except \citep{okajima2024asymptotic,NEURIPS2025_18ddfb19}. Our work can be regarded as an extension of these analyses to the iterative ST.

\begin{table}[t!]
    \centering
    \begin{tabular}{|l l|}
    \hline
    Notation & Description \\
    \hline\hline
        $\mu, \nu$ & sample indices of labeled and unlabeled data points
        \\
        $i,j$ & indices of the weight vector $\bm{w}$
        \\
        $[n]$ & for an positive integer $n$, the set $\{1,\dots,n\}$
        \\
        $T$ & total number of iteration used in ST
        \\
        $t\in[T]$ & index representing an iteration step in ST 
        \\
        & with a maximum number of iterations $T$
        \\
        $\left|\cdot\right|$    &   if applied to a number, absolute value
        \\
        $(\;\cdot \;)^\top$ & vector/matrix transpose
        \\
        $\bm{x} \cdot \bm{y}$ & for vectors $\bm{x},\bm{y}\in\R^N$, the inner product of them: $\bm{x} \cdot \bm{v}=\sum_{i=1}^N x_iy_i$.
        \\
        $v_i$   &   $i$-th element of a vector $\bm{v}$
        \\
        $\|\bm{v}\|_p$   & for a vector $\bm{v}=[v_i]_{1 \le i \le N}$, 
        \\
        & $\ell_p$ norm of the vector defined as $\left(\sum_{i=1}^N |v_i|^p\right)^{1/p}$
        \\
        $\bm{1}_N$ & an $N$-dimensional vector $(1,1,\dots,1)\in\mathbb{R}^N$
        \\
        $I_N$   &   identity matrix of size $N\times N$
        \\
        $\1(\cdot)$    & indicator function
        \\
        $\gN(\mu, \sigma^2)$ & Gaussian density with mean $\mu$ and variance $\sigma^2$
        \\
        $D\xi$  &   standard Gaussian measure $e^{-\xi^2/2}/\sqrt{2\pi}d\xi$
        \\
        $d\bm{x}$  &  with $\bm{x}\in\mathbb{R}^N$, a measure over $\mathbb{R}^N$
        \\
        $d^n\bm{x}$   & with $\bm{x}_1,\dots,\bm{x}_n\in\mathbb{R}^N$, a measure over $\mathbb{R}^{N \times n}$
        \\
        $\E_{X\sim p_X}[f(X)]$  & Expectation regarding random variable $X$
        \\
        & where $p$ is the density function for the random variable $X$
        \\
        & (lower subscript $X \sim p_X$ can be omitted if there is no risk of confusion)
        \\
        $\sV{\rm ar}_{X\sim p_X}[f(X)]$ & Variance: $\E[f(X)^2] - \E[f(X)]^2$
        \\
        $\{x_i\}\deq X$ & empirical distribution of $\{x_i\}$ is equal to the distribution of the r.v. $X$
        \\
        $\delta_{\rm d}(\cdot)$    & Dirac's delta function
        \\
        $\delta_{a, b}$    & for integers $a,b$, the Kronecker's delta: $\delta_{a,b}=\1(a=b)$
        \\
        $\mathop{\rm extr}_x f(x)$ & extremization with respect to $x$
        \\
        $\partial_i\gF$ & for a bivariate function $\gF(y, x)$, 
        \\
        & the partial derivative of $\gF$ with respect to the $i$-th argument
        \\
        & For example, $\partial_1 \gF(Y, X) = \left.\frac{\partial \gF}{\partial y}\right|_{y=Y, x=X}$. 
        \\
        $f(n) = \mathcal{O}(g(n))$  &   Landau's O as $n\to0$; $|f(n)/g(n)|<\infty$ as $n\to0$
        \\
        \hline
    \end{tabular}
    \caption{Notations}
    \label{table: notations}
\end{table}

\subsection{Notations}
\label{subsec: notation}
Throughout the paper, we use some shorthand notations for convenience. We summarize them in Table \ref{table: notations}.

\section{Problem setup}
\label{section:setup}
This section presents the problem setup and our interest in this work. The assumptions on the data generation process are first described and then the iterative ST procedure is formalized.

Let $D_L=\{(\bm{x}_\mu^{(0)}, y_\mu^{(0)})\}_{\mu=1}^{M_L}, \bm{x}_\mu^{(0)}\in\R^N, y_\mu^{(0)} \in \{0,1\}$ be the set of independent and identically distributed (iid) labeled data points, and let $D_U^{(t)}=\{\bm{x}_\nu^{(t)}\}_{\nu=1}^{M_U}, \bm{x}_\nu^{(t)}\in\R^N, t=1,2,\dots, T$ be the sets of iid unlabeled data points; there are $T$ batches of the unlabeled datasets of size $M_U$, thus, in total, there are $TM_U$ unlabeled data points. This study assumes that the data points are generated from binary Gaussian mixtures whose centroids are located at $\pm\bm{v}/\sqrt{N}$ with $\bm{v}\in\R^N$ as a fixed vector. The covariance matrices for these two Gaussian distributions are assumed to be spherical. From the rotational symmetry of these Gaussian distributions, we can fix the direction of the vector $\bm{v}$ as $\bm{v}=(1,1,\dots,1)\cdot$ without loss of generality. Furthermore, we assume that each Gaussian contains a fraction $\rho^{(t)}$ and $(1-\rho^{(t)})$ of the points with $\rho^{(t)} = \rho_L \in(0,0.5]$ if $t=0$ and $\rho^{(t)}=\rho_U\in(0,0.5]$, otherwise. In this setup, the feature vectors $\bm{x}_\mu^{(0)}$ and $\bm{x}_\nu^{(t)}$ can be written as
\begin{align}
    \bm{x}_\mu^{(0)} &= (2y_\mu^{(0)} - 1)\scalensqrt\bm{v} + \bm{z}_\mu^{(0)},\quad \mu=1,2,\dots, M_L,
    \label{eq:data genereation labeled}
    \\
    \bm{x}_\nu^{(t)} &= (2y_\nu^{(t)} - 1)\scalensqrt\bm{v} + \bm{z}_\nu^{(t)},
    \quad \nu=1,2,\dots,M_U,\;
    t=1,2,\dots,T,
    \label{eq:data genereation unlabeled}
\end{align}
where $\bm{z}_\mu^{(0)}\sim_{\rm iid}\mathcal{N}(0,\Delta_L I_N), \bm{z}_\nu^{(t)}\sim_{\rm iid }\mathcal{N}(0,\Delta_U I_N), \Delta_L,\Delta_U>0$ are the independent Gaussian noise and $y_\mu^{(0)}\sim_{\rm iid} p_{y}^{(0)},y_\nu^{(t)}\sim_{\rm iid}p_y^{(t)}$ are the ground truth label where $p_y^{(t)}$ is defined as
\begin{equation}
    p_y^{(t)}(y) = \begin{cases}
        p_{y, L}(y) \equiv \rho_L\delta_{\rm d}(y-1) + (1-\rho_L)\delta_{\rm d}(y), & t=0
        \vspace{1truemm}\\
        p_{y, U}(y) \equiv \rho_U\delta_{\rm d}(y-1) + (1-\rho_U)\delta_{\rm d}(y), & {\rm otherwise}
    \end{cases}
\end{equation}
The goal of ST is to obtain a classifier with a better generalization ability from 
\begin{equation}
    D=D_L\cup D_U^{(1)}\cup\dots D_U^{(T)},
\end{equation}
than the model trained with $D_L$ only.

We focus on the ST with the linear model, i.e., the model's output $f(\bm{x})$ at an input $\bm{x}$ is a function of a linear combination of the weight $\bm{w}$ and the bias $B$:
\begin{equation}
    f_{\rm model}(\bm{x}) = \sigma\left(
        \scalensqrt \bm{x} \cdot \bm{w} + B
    \right),
\end{equation}
\inred{where $\sigma:\R\to\R$ a link function that maps a logit to the model output; some examples are shown in Table \ref{table:functions}.} The factor $1/\sqrt{N}$ is introduced to ensure that the logit $\bm{x}\cdot\bm{w}/\sqrt{N} + B$ should have moderate magnitude at $N\gg1$.

In the following, we denote $\bm{\theta}=(\bm{w}, B)$ for the shorthand notation of the linear model's parameter. Similarly, a pseudo-label $\hat{y}(\bm{x})$ at an input $\bm{x}$ is also given as a linear combination of the model's parameter:
\begin{equation}
    \hat{y}(\bm{x}) = \sigma_{\rm pl}\left(
        \frac{1}{\sqrt{N}}\bm{x}\cdot \bm{w} + B
    \right),
\end{equation}
\inred{
    where $\sigma_{\rm pl}:\mathbb{R}\to\mathbb{R}$ is the pseudo-labeling function that maps the logit of the current model to the pseudo-label used at the next ST step. Depending on the choice of $\sigma_{\rm pl}$, the resulting pseudo-label may retain a real-valued confidence score (soft-label) or take a discrete value (hard-label). Examples are given in Table~\ref{table:functions}.
}
In each iteration, the model is trained by minimizing the ridge-regularized convex loss functions $l$ for the supervised learning and $l_{\rm pl}$ for the ST steps\inred{,
    where $l(y,f)$ denotes the loss between a ground-truth label $y$ and a model output $f$, and at the subsequent ST steps, $l_{\rm pl}(\hat{y},f)$ denotes the loss between a pseudo-label $\hat{y}$ and a model output $f$.
}
Our asymptotic characterization is formulated for general choices of $\sigma$, $\sigma_{\rm pl}$, $l$, and $l_{\rm pl}$ for which the objective function at each step is convex with respect to $\bm{\theta}^{(t)}$. Examples of these functions are given in
Table~\ref{table:functions}. Additional regularity assumptions needed for particular results are stated where they are used.

Furthermore, we accept removing unlabeled data points from the training data if the assigned pseudo-labels are not reliable enough. Here we simply define the reliability based on the magnitude of the logit. Then, given a threshold $\Gamma \ge 0$ for PLS, \inred{and regularization strength $\lambda^{(t)}\ge0$,} the ST algorithm is formalized as follows:
\begin{itemize}
    \item {\it Step 0: Initializing the model with the labeled data $D_L$.} Initialize the iteration number $t=0$ and obtain a model $\hat{\bm{\theta}}^{(0)}=(\hat{\bm{w}}^{(0)}, \hat{B}^{(0)})$ by minimizing the following loss:
    \begin{equation}
        \mathcal{L}^{(0)}(\bm{\theta}^{(0)}; D_L) = \sum_{\mu=1}^{M_L}l\left(
            y_\mu, \sigma\left(\scalensqrt \bm{x}_\mu^{(0)}\cdot \w{0}+ \bi{0}\right)
        \right) + \frac{\lambda^{(0)}}{2}\|\w{0}\|_2^2,
        \label{eq: loss labeled}
    \end{equation}
    with respect to $\bm{\theta}^{(0)}$.
    Let $t\leftarrow 1$, and proceed to Step 1.
    \item {\it Step1: Creating pseudo-labels.} Give the pseudo-labels for unlabeled data points in $D_U^{(t)}$ so that the pseudo-label for the unlabeled data point $\bm{x}_\mu^{(t)}$ is 
    \begin{equation}
        \hat{y}_\nu^{(t)} = \sigma_{\rm pl}\left(
            \scalensqrt\bm{x}_\nu^{(t)}\cdot\hat{\bm{w}}^{(t-1)} + \hat{B}^{(t-1)}  
        \right), \quad \nu=1,2,\dots, M_U.
    \end{equation}
    \item {\it Step2: Updating the model.} Obtain the model $\hat{\bm{\theta}}^{(t)}=(\hat{\bm{w}}^{(t)}, \hat{B}^{(t)})$ by minimizing the following loss:
    \begin{align}
      \begin{split}
        \mathcal{L}^{(t)}(\bm{\theta}^{(t)}; D_U^{(t)}, \hat{\bm{\theta}}^{(t-1)}) = \sum_{\nu=1}^{M_U}&
            \1\left(
                \left|
                    \scalensqrt\bm{x}_\nu^{(t)}\cdot\hat{\bm{w}}^{(t-1)} + \hat{B}^{(t-1)}
                \right| > \Gamma \sqrt{\bar{q}^{(t-1)}}
            \right)
            \\
            &\times l_{\rm pl}\left(
            \hat{y}_\nu^{(t)}, \sigma\left(
                \scalensqrt\bm{x}_\nu^{(t)}\cdot \bm{w}^{(t)} + B^{(t)}
            \right)
        \right)
        + \frac{\lambda^{(t)}}{2}\|\bm{w}^{(t)}\|_2^2,
        \end{split}
        \label{eq: update of ST}
    \end{align}
    with respect to $\bm{\theta}^{(t)}$, where $\bar{q}^{(t-1)}=\frac{1}{N}\|\hat{\bm{w}}^{(t-1)}\|_2^2$ is the normalized squared norm of the weight vector at the previous step.
    If $t<T$, let $t\leftarrow t+1$ and go back to Step 1.
\end{itemize}
 In the following, we will use $T$ as the total number of iterations in ST, and $t$ as the iteration index of ST when the total number of $T$ is fixed.

\begin{table}[t]
  \centering
  \caption{Examples of the combinations of $\sigma, \sigma_{\rm pl}, l$, and $l_{\rm pl}$.}
  \label{table:functions}
  \begin{tabular}{cccc}
   \hline 
    $\sigma(x)$ & $\sigma_{\rm pl}(x)$ & $l(p, q)$ & $l_{\rm pl}(p, q)$ \\
    \hline \hline
    $x$ & $x$ & $\frac{1}{2}(p-q)^2$ & $\frac{1}{2}(p-q)^2$ \\
    \hline
    $1/(1+e^{-x})$ & $1/(1+e^{-\gamma x}), \gamma>0$ & $-p \log q - (1 - p) \log(1 - q)$ & $-p \log q - (1 - p) \log(1 - q)$
    \\
    \hline
  \end{tabular}
\end{table}


\inred{
    We assume positive regularization in the ST steps to avoid a trivial solution $\hat{\theta}^{(t)}=\hat{\theta}^{(t-1)}$ for $t=1,2,\ldots$.To see this, consider $\lambda^{(1)}=\cdots=\lambda^{(T)}=0$ and a soft pseudo-label setting in which the pseudo-label is generated by the same output map as the model, and $l_{\rm pl}$ is minimized when its two arguments are equal. If we set $\bm{\theta}^{(t)}=\hat{\bm{\theta}}^{(t-1)}$, then the linear model gives the same logit on every unlabeled example as the previous model did. Hence the model output exactly reproduces the pseudo-label, and every pseudo-label loss term is minimized. Therefore, $\hat{\bm{\theta}}^{(t-1)}$ is a minimizer of the unregularized ST objective at step $t$. If this minimizer is unique, the update gives $\hat{\bm{\theta}}^{(t)}=\hat{\bm{\theta}}^{(t-1)}$, and by iteration $\hat{\bm{\theta}}^{(0)}=\hat{\bm{\theta}}^{(1)}=\cdots=\hat{\bm{\theta}}^{(T)}$. If the minimizer is not unique, the unregularized update itself is ambiguous. We therefore restrict our analysis to $\lambda^{(t)}>0$ for $t=1,\ldots,T$.
}

The iterative ST procedure above is simplified in two ways. First, the labeled data points are not used in steps 1 and 2.  Second, the algorithm does not use the same unlabeled data points in each iteration. These two aspects simplify the following analysis, which allows us to obtain a concise analytical result. Although the above ST has these differences from the commonly used procedures, we can still obtain non-trivial results as in the literature \citep{oymak2020statistical, oymak2021theoretical}.

\subsection{Generalization error}
Aside from the sharp characterization of the statistical properties of the estimators $\hat{\bm{\theta}}^{(t)}$, we are interested in the generalization performance of the ST algorithm that is evaluated by the generalization error defined as
\begin{align}
    \bar{\epsilon}_{\rm g}^{(t)} &= \E_{(\bm{x}, y)}\left[
        \1\left[
            y \neq \hat{y}_{\rm pred}(\bm{x}; \hat{\bm{\theta}}^{(t)})
        \right]
    \right],
    \label{eq: gen_err}
    \\
    \hat{y}_{\rm pred}(\bm{x}; \hat{\bm{\theta}}^{(t)}) &= \1\left[
            \scalensqrt\hat{\bm{w}}^{(t)}\cdot \bm{x} + \hat{B}^{(t)}
        > 0
    \right],
    \label{eq:yhat_generr}
\end{align}
where $(\bm{x}, y)$ follows the same data generation process with the labeled data points:
\begin{align}
    \bm{x} &= (2y-1)\scalensqrt \bm{v} + \bm{z},
    \label{eq:data generation generr x}
    \\
    y &\sim p_{y,L}, \quad \bm{z}\sim\mathcal{N}(0,\Delta_L I_N),
    \label{eq:data generation generr y}
\end{align}
For evaluating this, we consider the large system limit where $N, M_L,M_U\to\infty$, keeping their ratios as $(M_L/N, M_U/N) = (\alpha_L, \alpha_U)\in(0,\infty)\times (0,\infty)$. In this asymptotic limit, ST's behavior can be sharply characterized by the replica analysis presented in the next section. We term this asymptotic limit as {\it large system limit} (LSL). Hereafter, $N\to\infty$ represents LSL as a shorthand notation to avoid cumbersome notation.

As reported in \citep{mignacco2020role}, when the feature vectors are generated according the spherical Gaussians as assumed above, the generalization error \eqref{eq: gen_err} can be described by the bias $\hat{B}^{(t)}$ and two macroscopic quantities that characterize the geometrical relations between the estimator and the centroid of the Gaussians $\bm{v}$:
\begin{align}
    \bar{\epsilon}_{\rm g}^{(t)} &= \rho_U H\left(
        \frac{
            \bar{m}^{(t)} + \hat{B}^{(t)}
        }{
            \sqrt{\Delta_U \bar{q}^{(t)}}
        }
    \right) + (1-\rho_U)H\left(
        \frac{
            \bar{m}^{(t)} - \hat{B}^{(t)}
        }{
            \sqrt{\Delta_U \bar{q}^{(t)}}
        }
    \right),
    \label{eq: macro generalization error}
    \\
    H(x) &= \int_x^{\infty}  D\xi \inred{
        ,\qquad D\xi \equiv \frac{e^{-\xi^2/2}}{\sqrt{2\pi}}\,d\xi,
    }
    \\
    \bar{q}^{(t)} &= \scalen \|\hat{\bm{w}}^{(t)}\|_2^2, 
    \quad 
    \bar{m}^{(t)} = \scalen \hat{\bm{w}}^{(t)}\cdot \bm{v}.
    \label{eq:macro quantities}
\end{align}
Hence, evaluation of $\hat{B}^{(t)}, \bar{q}^{(t)}$ and $\bar{m}^{(t)}$ is crucial in our analysis.

The main question we want to investigate is how the statistical property of the estimators $\hat{\bm{\theta}}^{(t)}$ and the generalization error depends on the regularization parameters $\lambda^{(0)}, \lambda^{(1)},\dots, \lambda^{(T)}$, the number of iterations $T$, and the properties of the data, such as the degree of the label imbalance $\rho_L, \rho_U$.

\section{Sharp asymptotics of ST}
\label{section:replica analysis}
The first major technical contribution of this work is the development of a theoretical framework for sharply characterizing the behavior of ST using the replica method. We first rewrite ST as a limit of probabilistic inference in a statistical mechanics formulation in subsection \ref{subsec:reformulation}. Then, we propose the theoretical framework for analyzing the ST by using the replica method in subsection \ref{subsec:replica}, and show the first main analytical result for characterizing the behavior of ST in LSL in subsection \ref{subsec:rs free energy}. \inred{As usual in the replica method, it contains the non-rigorous steps due to the continuation to zero replica numbers. See Section~\ref{subsubsec: remark non-rigorous} for details.} Some comparisons with the numerical experiments are presented in subsection \ref{subsec:cross-check}. Detailed calculations are presented in Appendix \ref{app: replica method calculation}.

\subsection{statistical mechanics formulation of ST}
\label{subsec:reformulation}

\subsubsection{Chain of Boltzmann distributions}
Let us start with rewriting the ST as a statistical mechanics problem. We introduce probability densities $p^{(0)}, p^{(t)}, t=1,2,\dots, T$, which are termed the {\it Boltzmann distributions} following the custom of statistical mechanics, as follows. 
For $\beta^{(t)}>0$ and $t=0,1,\dots,T$, the Boltzmann distributions are defined as 
\begin{align}
p^{(0)}(\bm{\theta}^{(0)}|D_L) &= \frac{1}{Z^{(0)}} e^{-\beta^{(0)} \mathcal{L}^{(0)}(\bm{\theta}^{(0)};D_L)},
\label{eq:boltzmann-0}
\\
p^{(t)}(\bm{\theta}^{(t)}|D_U^{(t)}, \bm{\theta}^{(t-1)}) &= \frac{1}{Z^{(t)}}e^{-\beta^{(t)}\mathcal{L}^{(t)}(\bm{\theta}^{(t)}; D_U^{(t)}, \bm{\theta}^{(t-1)})}, \quad t=1,2,\dots,T,
\label{eq:boltzmann-t}
\end{align}
where $Z^{(0)}$ and $Z^{(t)},t=1,2,\dots,T$ are the normalization constants:
\begin{align}
    Z^{(0)}(D_L;\beta^{(0)}) &= \int e^{-\beta^{(0)} \mathcal{L}^{(0)}(\bm{\theta}^{(0)};D_L)} d\bm{\theta}^{(0)},
    \label{eq: partition function 0}
    \\
    Z^{(t)}(D_U^{(t)}, \bm{\theta}^{(t-1)};\beta^{(t)}) &= \int e^{-\beta^{(t)}\mathcal{L}^{(t)}(\bm{\theta}^{(t)}; D_U^{(t)} ,\bm{\theta}^{(t-1)})} d\bm{\theta}^{(t)}, \quad t=1,2,\dots,T.
    \label{eq: partition function t}
\end{align}
Recall that $\gL^{(0)}$ and $\gL^{(t)}, t=1,2,\dots, T$ are the loss functions used in each step of ST. This chain of the Boltzmann distributions defines a Markov process over $\{\bm{\theta}^{(t)}\}_{t=0}^T$ conditioned by the data $D$:
\begin{equation}
    p_{\rm ST}(\{\bm{\theta}^{(t)}\}_{t=0}^T|D) \equiv p^{(0)}(\bm{\theta}^{(0)}|D_L)\prod_{t=1}^T p^{(t)}(\bm{\theta}^{(t)}|D_U^{(t)}, \bm{\theta}^{(t-1)}).
\end{equation}

By successively taking the limit $\beta^{(0)}\to\infty, \beta^{(1)}\to\infty,\dots$, the Boltzmann distributions $p^{(0)}, p^{(1)}, \dots, $ converge to the Delta functions at $\hat{\bm{\theta}}^{(0)}, \hat{\bm{\theta}}^{(1)}, \dots$, respectively. Thus analyzing ST is equivalent to analyzing the Boltzmann distributions at this limit\footnote{
 As the aim of this study is not to provide rigorous analysis but to provide theoretical insights by using a non-rigorous heuristic of statistical mechanics, we assume that the exchange of limits and integrals is possible throughout the study without further justification.
}. Hereafter the limit without the upper subscript $\beta\to\infty$ represents taking all of the successive limits $\beta^{(t)}\to\infty, t=0,1,\dots, T$ as a shorthand notation. Furthermore, we will omit the arguments $D_L, D_U^{(t)}, \bm{\theta}^{(t)}, \beta^{(t)}$ when there is no risk of confusion to avoid cumbersome notation.

\subsubsection{Generating functional}
For evaluating the behavior of the weights $\hat{w}_i^{(t)}$
and the biases $\hat{B}^{(t)}$, we introduce the {\it generating functional}.
\inred{
    Let $g_w:\mathbb{R}^{T+1}\to\mathbb{R}$ and $g_B:\mathbb{R}^{T+1}\to\mathbb{R}$ be arbitrary test functions of the trajectories of a weight coordinate and the bias, respectively, such that the expectations and derivatives below are well defined. We also introduce auxiliary real-valued source parameters $\epsilon_w,\epsilon_B\in\mathbb{R}$, which are set to zero after differentiation and are used to extract the desired expectations. Then, the generating functional defined as \footnote{
       Readers familiar with statistical mechanics might be interested in the similarity between this and the generating functional of the Martin-Siggia-Rose formalism \citep{martin1973statistical}, or alternatively called the path integral formulation, used in the dynamical mean-field theory (DMFT), which has recently been used to analyze the learning dynamics \citep{Agoritsas_2018, Mignacco2020dynamical, Mignacco_2021_stochasticity,bordelon2022selfconsistent, bordelon2023dynamics, bordelon2023theinfluence, pehlevan2023lecture}. Our generating functional is basically the same as that used in the DMFT. However, in our setup, to incorporate the update rule based on the implicit solution of the optimization problem \eqref{eq: update of ST}, it is necessary to treat the Boltzmann distributions with nontrivial normalization constants using the replica method. This is in contrast to conventional setups of DMFT. In a conventional setup of DMFT, the expression of the parameters at the next time step is explicitly given, and thus the time evolution can be explicitly written using delta functions, which allows us to consider an average directly over the data without the replica method. In this sense, our analysis can be viewed as a DMFT analysis of the implicit update rule using the replica method.
    }
}
\begin{equation}
    \Xi_{\rm ST}(\epsilon_w, \epsilon_B) = \lim_{N,\beta\to\infty}  \E_{\{\bm{\theta}^{(t)}\}_{t=0}^T\sim p_{\rm ST}, D}\left[
        e^{\epsilon_w g_w(\{w_i^{(t)}\}_{t=0}^{T}) 
        + \epsilon_Bg_B(\{B^{(t)}\}_{t=0}^{T})} 
    \right].
    \label{eq: generating functional}
\end{equation}
More specifically, the average quantities $\E_D[g_w(\{\hat{w}_i^{(t)}\})]$  and $\E_D[g_B(\{\hat{B}^{(t)}\})]$ can be evaluated by using the following formulae:
\begin{align}
    \E_D\left[g_w\left(\{\hat{w}_i^{(t)}\}_{t=0}^T\right)\right] &= \lim_{\epsilon_w, \epsilon_B\to0}\frac{\partial}{\partial \epsilon_w} \Xi_{\rm ST}(\epsilon_w, \epsilon_B),
    \label{eq: generating functional derivative w}
    \\
    \E_D\left[g_B\left(\{\hat{B}^{(t)}\}_{t=0}^T\right)\right] &= \lim_{\epsilon_w, \epsilon_B\to0}\frac{\partial}{\partial \epsilon_B} \Xi_{\rm ST}(\epsilon_w,  \epsilon_B).
    \label{eq: generating functional derivative B}
\end{align}

\subsection{Replica method for ST}
\label{subsec:replica}
The evaluation of the generating functional \eqref{eq: generating functional} is technically difficult because the average over $\{\bm{\theta}^{(t)}\}_{t=0}^{T-1}$ and $D$ requires averaging the inverse of the normalization constants $(Z^{(0)}(D_L))^{-1}\prod_{t=1}^{T-1}(Z^{(t)}(D_U^{(t)},\bm{\theta}^{(t-1)}))^{-1}$ in the Boltzmann distributions:
\begin{align}
    &\Xi_{\rm ST}(\epsilon_w, \epsilon_B) = \lim_{N,\beta\to\infty}\E_D\left[\int 
        e^{
            \epsilon_w g_w(\{w_i^{(t)}\}_{t=0}^{T}) 
            + \epsilon_Bg_B(\{B^{(t)}\}_{t=0}^{T})} 
        \nonumber
        \right.
        \\
        &\left.
        \times 
        \frac{
            1
        }{
            Z^{(0)}(D_L)
        }e^{
            -\beta^{(0)}\gL^{(0)}(\bm{\theta}^{(0)};D_L)
        }
        \prod_{t=1}^T\frac{
            1
        }{
            Z^{(t)}(D_U^{(t)}, \bm{\theta}^{(t-1)})
        }e^{
            -\beta^{(t)}\gL^{(t)}(\bm{\theta}^{(t)};D_U^{(t)}, \bm{\theta}^{(t-1)})
        }
    d\bm{\theta}^{(0)}\dots d\bm{\theta}^{(T)}\right],
    \label{eq: def of generating functional}
\end{align}
\inred{where integration over $\theta^{(0)},\ldots,\theta^{(T)}$ in Eq.~\eqref{eq: def of generating functional} applies to the whole integrand, which is split over multiple lines for readability.} To resolve this difficulty, we use the replica method as follows. This method rewrites $\Xi_{\rm ST}$ using the identity $(Z^{(t)})^{-1} = \lim_{n_t\to0}(Z^{(t)})^{n_t-1}$ as 
\begin{align}
    \Xi_{\rm ST} &= \lim_{n_0,\dots,n_T\to0} \phi^{(T)}_{n_0,\dots,n_T},
    \\
    \phi^{(T)}_{n_0,\dots,n_T} &= \lim_{N,\beta\to\infty}\E_D\left[\int 
        e^{\epsilon_w g_w(\{w_i^{(t)}\}_{t=0}^{T}) 
        + \epsilon_Bg_B(\{B^{(t)}\}_{t=0}^{T})} 
        \nonumber 
        \right.
        \\
        &\left.
        \times \left(Z^{(0)}(D_L)\right)^{n_0-1}
        e^{
            -\beta^{(0)}\gL^{(0)}(\bm{\theta}^{(0)};D_L)
        }
        \right.
        \nonumber
        \\
        &\left.
        \times \prod_{t=1}^T
        \left(Z^{(t)}(D_U^{(t)}, \bm{\theta}^{(t-1)})\right)^{n_t-1}
        e^{
            -\beta^{(t)}\gL^{(t)}(\bm{\theta}^{(t)};D_U^{(t)}, \bm{\theta}^{(t-1)})
        }
    d\bm{\theta}^{(0)}\dots d\bm{\theta}^{(T)}\right].
\end{align}
Although the evaluation of ${\phi}_{n_0,\dots,n_T}^{(T)}$ for $n_t\in\R$ is difficult, this expression has the advantage explained next. For positive integers $n_0,\dots,n_T =1,2,\dots$, it has an appealing expression:
\begin{align}
    \phi^{(T)}_{n_0,\dots,n_T} &= \lim_{N,\beta\to\infty}\E_D\left[\int 
        e^{\epsilon_w g_w(\{w_{1,i}^{(t)}\}_{t=0}^{T}) 
        + \epsilon_Bg_B(\{B_1^{(t)}\}_{t=0}^{T})} 
        \prod_{a_0=1}^{n_0}
        e^{
            -\beta^{(0)}\gL^{(0)}(\bm{\theta}_{a_t}^{(0)}; D_L)
        }
        \right.
        \nonumber
        \\
        &\left.
        \times \prod_{t=1}^T
        \prod_{a_t=1}^{n_t}
        e^{
            -\beta^{(t)}\gL^{(t)}(\bm{\theta}_{a_t}^{(t)};D_U^{(t)}, \bm{\theta}_1^{(t-1)})
        }
    d^{n_0}\bm{\theta}^{(0)}\dots d^{n_T}\bm{\theta}^{(T)}\right],
    \label{eq: replicated system}
\end{align}
where $d^{n_{t}}\bm{\theta}^{(t)}, t=0,1,\dots, T$ are the shorthand notations for $d\bm{\theta}_1^{(t)}\dots d\bm{\theta}_{n_t}^{(t)}$, and $\{a_t\}_{t=0}^T$ are indices to distinguish the additional variables introduced by the replica method, hereafter we will refer to this type of index as the {\it replica index}. Note that the index $1$ that appears in $\epsilon_w g_w(\{w_{1, i}^{(t)}\}_{t=0}^{T})  + \epsilon_Bg_B(\{B_1^{(t)}\}_{t=0}^{T})$ is also the replica index.  The augmented system \eqref{eq: replicated system}, which we refer to as the {\it replicated system}, is much easier to handle than the original generating functional because all of the factors to be evaluated are now explicit. Thus, we can consider the average over $D$ {\it before} conducting the integral (or the optimization at $\beta\to\infty$) over $\{\bm{\theta}_{\a t}\}_{t=0}^T$. In the following, utilizing this formula, we obtain an analytical expression for $n_t\in \sN$. Subsequently, under appropriate symmetry assumption, we extrapolate that expression to the limit as $n_t\to0$. 

In the following, we briefly sketch the treatment of the replicated system \eqref{eq: replicated system}. The readers who are interested in the result can skip Section \ref{subsubsec: handling of the replicated system} and proceed to Section \ref{subsec:rs free energy}.

\subsubsection{\inred{A remark on the non-rigorous aspect of the replica method}}
\label{subsubsec: remark non-rigorous}
This analytical continuation $n_t\to0$ from $n_t\in\sN$ is the procedure that has not yet been formulated in a mathematically rigorous manner. In fact, the sequence of the replicated system \eqref{eq: replicated system} for integer $n_t$ alone may not uniquely determine the expression for the replicated system at real $n_t\in\R$. Therefore, this analytic continuation does not have a mathematically precise formulation, although the evaluation of $\phi_{n_0,\dots,n_T}^{(T)}$ for positive integers is a well-defined problem. It is merely an extrapolation based on guessing the form of the function. This aspect is what renders the replica method a non-rigorous technique. However, it is empirically known that extrapolating the results obtained in $n_t\in\sN$ yields the correct result for convex optimization problems, in the sense that the same result is later obtained by the other mathematical techniques \citep{barbier2019optimal, gabrie2018, charbonneau2023spin}. Therefore, though the replica analysis includes a procedure whose mathematical validity is not proofed, we believe that this does not have a serious impact on our analysis since the cost function at each step of self-training is convex.

\subsubsection{Handling of the replicated system}
\label{subsubsec: handling of the replicated system}
The important observation is that the Gaussian noise terms in \eqref{eq: replicated system} only appear through the following vectors:
\begin{align}
    \bm{u}_\mu^{(0)} &= \left(
        \scalensqrt \rw{1}{0}\cdot \bm{z}_\mu^{(0)}, \scalensqrt\rw{2}{0}\cdot \bm{z}_\mu^{(0)}, \dots, \scalensqrt \rw{n_0}{0}\cdot \bm{z}_\mu^{(0)}
    \right)^\top
    \in\R^{n_0}
    ,
    \\
    \bm{u}_\nu^{(t)} &= \left(
        \scalensqrt \rw{1}{t-1}\cdot \bm{z}_\nu^{(t)}, \scalensqrt \rw{1}{t}\cdot \bm{z}_\nu^{(t)}, \scalensqrt \rw{2}{t}\cdot \bm{z}_\nu^{(t)}, \dots,  \scalensqrt \rw{n_t}{t}\cdot \bm{z}_\nu^{(t)}
    \right)^\top 
    \in\R^{n_t+1}
    ,
    \nonumber \\
    t &= 1,2,\dots,T, 
\end{align}
which follows independently the multivariate Gaussians as  
\begin{align}
    \bm{u}_\mu^{(0)}&\sim_{\rm iid} \mathcal{N}(0, \Sigma^{(0)}), \quad
    \Sigma^{(0)} = \Delta_L\times\left[\begin{array}{ccc}
        Q_{11}^{(0)} & \cdots & Q_{1n_0}^{(0)}  \\
        \vdots &  \ddots & \vdots \\
        Q_{n_01}^{(0)} & \cdots & Q_{n_0n_0}^{(0)}
    \end{array}
    \right],
    \\
    Q_{\a{0}\b0}^{(0)}&\equiv \scalen \bm{w}_{\a{0}}^{(0)}\cdot \bm{w}_{\b{0}}^{(0)}, \quad \a{0},\b{0} = 1,2,\dots, n_0,
\end{align}
and
\begin{align}
    \bm{u}_\nu^{(t)}&\sim_{\rm iid} \mathcal{N}(0, \Sigma^{(t)}),
    \quad 
    \Sigma^{(t)} = \Delta_U\times\left[
        \begin{array}{c|ccc}
             Q_{11}^{(t-1)} & R_{1}^{(t)} & \cdots & R_{n_t}^{(t)} \\
             \hline
             R_{1}^{(t)} & Q_{11}^{(t)} & \cdots & Q_{1n_t}^{(t)} \\
             \vdots & \vdots & \ddots & \vdots \\
             R_{n_t}^{(t)} & Q_{n_t 1}^{(t)} & \cdots & Q_{n_tn_t}^{(t)}
        \end{array}
    \right]
    \\
    Q_{\a t \b t}^{(t)} &= \scalen\bm{w}^{(t)}_{\a t}\cdot \bm{w}_{\b t}^{(t)},
    \quad 
    R_{\a t}^{(t)} = \scalen \bm{w}^{(t-1)}_1\cdot \bm{w}_{\a t}^{(t)},
    \quad 
    \a{t},\b{t}=1,2,\dots,n_{t},
 \end{align}
 for a fixed set of $\{\bm{w}_{\a{t}}^{(t)}\}_{\a t=1}^{n_t}, t=0,\dots,T$. Since each data point is independent, the integral over $(\bm{u}_\mu^{(0)}, y_\mu^{(0)})$ and $(\bm{u}_{\nu}^{(t)}, y_\nu^{(t)})$ can be taken independently once $\{\bm{\theta}^{(t)}\}$ are fixed. As a result, the generating functional does not depend on the index $\mu$ and $\nu$, hence, we can omit it.
 
 Above observation indicates that the replicated system \eqref{eq: replicated system} depends on $\{\bm{w}_{\a{t}}^{(t)}\}_{\a t=1}^{n_t}, t=0,\dots,T$ only through their inner products, such as 
\begin{align}
    &\scalen \bm{w}_{\a t}^{(t)}\cdot \bm{w}_{\b t}^{(t)}, \quad \a t,\b t=1,2,\dots,n_t, \; t=0,1,\dots T,
    \\
    &\scalen \bm{w}_{\a t}^{(t)}\cdot \bm{v}, \quad \a t = 1,\dots, n_t,\; t=0,1,\dots, T,
    \\
    &\scalen \bm{w}_{1}^{(t-1)}\cdot \bm{w}_{\a t}^{(t)}, \quad \a t =1,2,\dots,n_t,\; t=1,\dots,T,
\end{align}
which capture the geometric relations between the estimators and the centroid of clusters $\bm{v}$. We refer to them as {\it order parameters}.
Thus, by introducing the auxiliary variables through the trivial identities of the delta functions
\begin{align}
         1 &= \prod_{t=0}^{T}\prod_{1\le\a{t}\le\b{t}\le n_{t}}N\int \delta_{\rm d}\left(
        NQ_{\a t\b t}^{(t)} - \bm{w}_{\a t}^{(t)}\cdot \bm{w}_{\b t}
    \right)dQ_{\a{t}\b{t}}^{(t)},
     \\
     1 &= \prod_{t=0}^{T} \prod_{\a t=1}^{n_{t}}N\int
        \delta_{\rm d}\left(
            Nm_{\a t}^{(t)} - \bm{w}_{\a t}^{(t)}\cdot\bm{v}
        \right)
     dm_{\a t}^{(t)},
     \\
     1 &= \prod_{t=1}^T \prod_{\a t=1}^{n_ t}N\int
        \delta_{\rm d}\left(
            NR_{\a t}^{(t)} - \bm{w}_{1}^{(t-1)}\cdot\bm{w}_{\a t}^{(t)}
        \right)
     dR_{\a t}^{(t)},
\end{align}
and their Fourier representations
\begin{align}
    \delta_{\rm d}\left(
        NQ_{\a t\b t}^{(t)} - \bm{w}_{\a t}^{(t)}\cdot \bm{w}_{\b t}
    \right) &= \frac{1}{4\pi}\int e^{-
        (NQ_{\a t\b t}^{(t)} - \bm{w}_{\a t}^{(t)}\cdot \bm{w}_{\b t})\frac{\tilde{Q}_{\a t \b t}^{(t)}}{2}
    }d\tilde{Q}_{\a t \b t}^{(t)},
    \\
    \delta_{\rm d}\left(
            Nm_{\a t}^{(t)} - \bm{w}_{\a t}^{( t)}\cdot\bm{v}
        \right) &= \frac{1}{2\pi}\int
        e^{
            (Nm_{\a t}^{(t)} - \bm{w}_{\a t}^{( t)}\cdot\bm{v})\tilde{m}_{\a t}^{(t)}
        }
    d\tilde{m}_{\a t}^{(t)},
    \\
    \delta_{\rm d}\left(
        NR_{\a t}^{(t)} - \bm{w}_{1}^{(t-1)}\cdot\bm{w}_{\a t}^{(t)}
    \right) & = \frac{1}{2\pi}\int
        e^{
            (NR_{\a t}^{(t)} - \bm{w}_{1}^{(t-1)}\cdot\bm{w}_{\a t}^{(t)})\tilde{R}_{\a t}^{(t)}
        }
    d\tilde{R}_{\a t}^{(t)},
\end{align}
the integrals over $\{w_{a_t, j}^{(t)}\}$ can be independently performed for each $j$. As a result, the result no longer depends on the index $i$, hence, we can safely omit it. This is a natural consequence of the data distribution having rotation-invariant symmetry.  For $t\ge1$, let $\bm{\Theta}^{(t)}$ and $\hat{\bm{\Theta}}^{(t)}$ be the collection of the variables
\begin{align}
    Q^{(t)}&=[Q_{\a t \b t}^{(t)}]_{\substack{1\le \a t \le n_t \\ 1\le \b t\le n_t}},
    \;
    \bm{m}^{(t)}=[m_{\a t}^{(t)}]_{1\le \a{t} \le n_t},
    \;
    \bm{R}^{(t)} = [R_{\a t}^{(t)}]_{1\le \a t \le n_t},
    \;
    \bm{B}^{(t)}=[b_{\a t}^{(t)}]_{1\le \a t \le n_t},
    \label{eq: order parameters}
\end{align}
and
\begin{align}
    \tilde{Q}^{(t)}&=[\tilde{Q}_{\a t \b t}^{(t)}]_{\substack{1\le \a t \le n_t \\ 1\le \b t\le n_t}},
    \;
    \tilde{\bm{m}}^{(t)}=[\tilde{m}_{\a t}^{(t)}]_{1\le \a{t} \le n_t},
    \;
    \tilde{\bm{R}}^{(t)} = [\tilde{R}_{\a t}^{(t)}]_{1\le \a t \le n_t}, 
    \label{eq: conjugate parameters}
\end{align}
respectively.  $\Theta^{(0)}$ and $\hat{\Theta}^{(0)}$ are defined similarly. Also, let $\Theta$ and $\hat{\Theta}$ be $\{\Theta\}_{t=0}^T, \{\hat{\Theta}\}_{t=0}^T$, respectively. Then, we find that $\phi_{n_0,\dots,n_T}^{(T)}$ can be written as
\begin{align}
    \phi_{n_0,\dots,n_T}^{(T)} &= \lim_{N, \beta\to\infty}\int e^{N\gS(\Theta,\hat{\Theta})}\E_{\{\bm{w}^{(t)}\}\sim \tilde{p}_{{\rm eff}, w}}\left[e^{\epsilon_w g_w(\{w_1^{(t)}\}_{t=0}^T)}\right]
    e^{\epsilon_B g_B(\{B_1^{(t)}\}_{t=0}^T)}d\Theta d\hat{\Theta},
    \label{eq: generating functional general}
\end{align}
where
\begin{align}
    \tilde{p}_{{\rm eff},w}(\{\bm{w}^{(t)}\}_{t=0}^T) &\propto e^{
        -\frac{1}{2}(\bm{w}^{(0)})^\top (\tilde{Q}^{(0)} + \beta^{(t)}I_{n_t})\bm{w}^{(0)} + (\tilde{\bm{m}}^{(t)})\cdot \bm{w}^{(0)} 
    }
    \nonumber \\
    &\hspace{20truemm}\times\prod_{t=1}^T e^{
        -\frac{1}{2}(\bm{w}^{(t)})^\top (\tilde{Q}^{(t)} + \beta^{(t)}I_{n_t})\bm{w}^{(t)} + (\tilde{\bm{m}}^{(t)} + w_1^{(t-1)}\bm{R}^{(t)})\cdot \bm{w}^{(t)} 
    }.
\end{align}
Thus, $\Theta$ and $\hat{\Theta}$ are evaluated by the extremum condition $\mathop{\rm extr}_{\Theta, \hat{\Theta}}\gS(\Theta, \hat{\Theta})$ when $N\gg1$, where $\mathop{\rm extr}_{\Theta, \hat{\Theta}}\dots$ represents an operation taking extremum of a function over $\Theta, \hat{\Theta}$. The concrete form of $\gS$ is given as \eqref{appeq: gibbs general} in Appendix \ref{subapp: handling of the replicated system}. $\tilde{p}_{{\rm eff}, w}$ is a density function over $\{\bm{w}^{(t)}\}_{t=0}^T$. Here $\bm{w}^{(t)} = (w_1^{(t)}, \dots w_{n_t}^{(t)})\cdot \in \R^{n_t}$.
Recall that the index $1$ in \eqref{eq: generating functional general} is the replica index, and the result no longer depends on the index $i\in[N]$. It should also be noted that essentially only Gaussian integrals and the saddle-point method are required for the above calculations.  This computational simplicity is a major advantage of the replica method, which introduces the replicated system \eqref{eq: replicated system} that allows us to first consider the average over the noise.

The key non-trivial point is that, in order to extrapolate as $n_0, \dots, n_T\to0$, we need to impose some symmetry on the saddle point. Otherwise, the discreteness of $n_t$ prevents us from extrapolating as $n_t\to0$. The simplest choice is the replica symmetric (RS) one:
\begin{align}
    Q^{(t)} &= \frac{\chi^{(t)}}{\beta^{(t)}}I_{n_t} + q^{(t)}\bm{1}_{n_t}\bm{1}_{n_t}\cdot,
    \label{eq:RS assumption1}
    \\
    \bm{R}^{(t)} &= R^{(t)}\bm{1}_{n_t},
    \label{eq:RS assumption2}
    \\
    \bm{m}^{(t)} &= m^{(t)}\bm{1}_{n_t},
    \label{eq:RS assumption3}
    \\
    \bm{B}^{(t)} &= B^{(t)}\bm{1}_{n_t},
    \label{eq:RS assumption4}
    \\
    \tilde{Q}^{(t)} &= \beta^{(t)}\hat{Q}^{(t)}I_{n_t} - (\beta^{(t)})^2 \hat{\chi}^{(t)} \bm{1}_{n_t}\bm{1}_{n_t}\cdot,
    \label{eq:RS assumption5}
    \\
    \tilde{\bm{R}}^{(t)} &= \beta^{(t)}\hat{R}^{(t)}\bm{1}_{n_t},
    \label{eq:RS assumption6}
    \\
    \tilde{\bm{m}}^{(t)} &= \beta^{(t)}\hat{m}^{(t)}\bm{1}_{n_t},
    \label{eq:RS assumption7}
\end{align}
that reflect the symmetry of the replicated system \eqref{eq: replicated system}. Under this ansatz, for $t\ge1$, $\Theta^{(t)}$ and $\hat{\Theta}^{(t)}$ represents $(q^{(t)}, \chi^{(t)}, R^{(t)}, m^{(t)}, B^{(t)})$ and $(\hat{Q}^{(t)}, \hat{\chi}^{(t)}, \hat{R}^{(t)}, \hat{m}^{(t)})$, respectively. Imposing this symmetry, we can obtain an analytical expression that can be formally extrapolated as $n_t\in\R$ from $n_t\in\sN$. Specifically, let $\mathcal{O}(n)\equiv\sum_{t=0}^{T}\mathcal{O}(n_t)$ be terms that vanish at the limit $n_0,n_1,\dots,n_T\to0$. Then, under the RS assumption, it can be shown that $\gS(\Theta, \hat{\Theta})=\gO(n)$. Hence $e^{N\gS(\Theta,\hat{\Theta})}$ yields a factor of unity at $n\to0$. Thus, the generating functional can be written as 
\begin{align}
    \phi_{n_0, \dots,n_T}^{(T)} &= \lim_{\beta\to\infty}\E_{\xi_w^{(t)}\sim_{\rm iid}\gN(0,1)}\left[
        \E_{\{w^{(t)}\}\sim p_{{\rm eff}, w}(\{w^{(t)}\}|\{\xi_w^{(t)}\})}\left[
            e^{\epsilon_w g_w(\{w^{(t)}\}_{t=0}^T)}
        \right]
    \right]
    \nonumber \\
    &\hspace{0truemm}\times e^{\epsilon_B g_B(\{B^{(t)}\}_{t=0}^T)} +\gO(n),
\end{align}
where
\begin{align}
    p_{{\rm eff}, w}(\{w^{(t)}\}_{t=0}^T|\{\xi_w^{(t)}\}) &= \gN\left(
        w^{(0)}
        \mid
        \frac{\hat{m}^{(0)}+\sqrt{\hat{\chi}^{(0)}}\xi_w^{(0)}}{\hat{Q}^{(0)}+\lambda^{(0)}},
        \frac{\hat{Q}^{(0)}+\lambda^{(0)}}{\beta^{(0)}}
    \right)
    \nonumber \\
    &\times\prod_{t=1}^T\gN\left(
        w^{(t)}
        \mid
        \frac{\hat{m}^{(t)} + \hat{R}^{(t)}w^{(t-1)} +\sqrt{\hat{\chi}^{(t)}}\xi_w^{(t)}}{\hat{Q}^{(t)}+\lambda^{(t)}},
        \frac{\hat{Q}^{(t)}+\lambda^{(t)}}{\beta^{(t)}}
    \right),
\end{align}
This implies that, at the limit $\beta\to\infty, n_0, \dots,n_T\to0$, it is governed by a one-dimensional Gaussian process:
\begin{align}
    \Xi_{\rm ST}(\epsilon_w, \epsilon_B) &= \E_{\xi_w^{(t)} \sim_{\rm iid}\gN(0,1)}\left[
        e^{\epsilon_w g_w(\{\wsf{t}\}_{t=0}^T)}
    \right]
    e^{\epsilon_B g_B(\{B^{(t)}\}_{t=0}^T)},
    \\
    \wsf{0} &= \frac{\hat{m}^{(0)}+\sqrt{\hat{\chi}^{(0)}}\xi_w^{(0)}}{\hat{Q}^{(0)}+\lambda^{(0)}},
    \\
    \wsf{t} &= \frac{\hat{m}^{(t)} + \hat{R}^{(t)}w^{(t-1)} +\sqrt{\hat{\chi}^{(t)}}\xi_w^{(t)}}{\hat{Q}^{(t)}+\lambda^{(t)}},
\end{align}
where, $\Theta, \hat{\Theta}$ are determined by the saddle point condition.

Similarly, it can be possible to consider another generating functional regarding $y_\nu^{(t)}$, $\tilde{u}_{\nu}=\bm{x}_{\nu}^{(t)}\cdot\bm{w}^{(t-1)}/\sqrt{N} + {B}^{(t-1)}$ and $u_{\nu}=\bm{x}_{\nu}^{(t)}\cdot\bm{w}^{(t)}/\sqrt{N} + {B}^{(t)}$:
\begin{equation}
    \Xi_{\rm ST}(\epsilon_u) = \lim_{N,\beta\to\infty}  \E_{\{\bm{\theta}^{(t)}\}_{t=0}^T\sim p_{\rm ST}, D}\left[
        e^{
            \epsilon_u g_u(\{(y_\nu^{(t)}, \tilde{u}_{\nu}^{(t)}, u_{\nu}^{(t)})\}_{t=1}^{T})
        }
    \right], 
    \label{eq: generating functional logits}
\end{equation}
where $\nu \in [M_U]$ and $t\ge1$. The calculation procedure is completely analogous to that of $\Xi_{\rm ST}(\epsilon_w, \epsilon_B)$. Analyzing this generating functional yields the statistical properties of the logits. See Appendix \ref{appsubeq: rs generating functional} for more detail.

The generating functional obtained by imposing this symmetry is called the {\it RS solution}. As already commented in the beginning of subsection \ref{subsec:replica}, in general, the expression for the replicated system \eqref{eq: replicated system} with integer $\{n_t\}_{t=0}^T$ alone cannot uniquely determine the expression for the replicated system at real $\{n_t\}_{t=0}^T$. However, for log-convex Boltzmann distributions, it has been empirically known that the replica symmetric choice of the saddle point, which should be valid for $n_t\in\sN$, yields the correct extrapolation \citep{gabrie2018, barbier_adaptive_2019, barbier2019optimal, mignacco2020role, gerbelot2020asymptotic, gerbelot2023asymptotic, montanari2024} in the sense that the same result by the replica method with the RS assumption have been obtained through a different mathematically rigorous approach. Since the Boltzmann distributions in our setup are log-convex functions once conditioned on the data and the parameter of the previous iteration step, we can expect the RS assumption to yield the correct result even in the current iterative optimization.

\subsection{RS solution}
\label{subsec:rs free energy}
In this subsection, we present the asymptotic properties of the regressors obtained by ST under the RS ansatz on the saddle point \eqref{eq:RS assumption1}-\eqref{eq:RS assumption7}.  It is described by a small finite set of scalar quantities determined as a solution of nonlinear equations, which we refer to as {\it self-consistent equations}. See Appendix \ref{app: replica method calculation} for the derivations.

\inred{
    The statements in this subsection are non-rigorous predictions obtained from the replica calculation under the RS ansatz. Hence, we refer to them as Predictions, rather than Theorems, to make their mathematical status explicit. They characterize the predicted large-system limit of the estimator and the logits. Their accuracy is assessed against finite-size numerical experiments in Section~\ref{subsec:cross-check}.
}

\subsubsection{RS saddle point}
\inred{
    First, we define the self-consistent equations that determine the scalar order parameters $\Theta^{(t)}$ and $\hat{\Theta}^{(t)}$. At a high level, these parameters provide a finite-dimensional effective description of the high-dimensional estimator at step $t$. The parameters in $\Theta^{(t)}$ describe macroscopic properties of the learned classifier, such as the norm of the weight vector, its alignment with the cluster direction. These quantities determine the effective distribution of the logits. The conjugate parameters in $\hat{\Theta}^{(t)}$ govern the associated scalar effective process for individual coordinates of the learned weight vector. The two sets of parameters are jointly determined by the self-consistent equations below.
}
Let us define $l_L(y,x)$ and $l_U(y,x;\tilde{\Gamma})$ as 
\begin{align}
    l_L(y,x) &= l(y, \sigma(x)),
    \\
    l_U(y, x;\tilde{\Gamma}) &= \1(|y|>\tilde{\Gamma})l_{\rm pl}(\sigma_{\rm pl}(y), \sigma(x)).
\end{align}
Then, the self-consistent equations are summarized as follows. 

\begin{definition}[Self-consistent equations]
\label{def: self-consistent equations}
The following set of quantities $\Theta=\{\Theta^{(t)}\}_{t=0}^T, \hat{\Theta}=\{\hat{\Theta}^{(t)}\}_{t=0}^T$ are determined as the solution of the following set of non-linear equations that are referred to as {\it self-consistent equations}:
\begin{align}
    \Theta^{(0)} &= (q^{(0)}, \chi^{(0)}, m^{(0)}, B^{(0)}),
    \\
    \Theta^{(t)} &= (q^{(t)}, \chi^{(t)}, R^{(t)}, m^{(t)}, B^{(t)}), \quad t \in [T],
    \\
    \hat{\Theta}^{(0)} &= (\hat{Q}^{(0)}, \hat{\chi}^{(0)}, \hat{m}^{(0)}),
    \\
    \hat{\Theta}^{(t)} &= (\hat{Q}^{(t)}, \hat{\chi}^{(t)}, \hat{R}^{(t)}, \hat{m}^{(t)}), \quad t \in [T].
\end{align}

Let $\usf{0}$ be the solution of the following one-dimensional randomized optimization problem:
\begin{align}
        \usf{0} &= \arg\min_{u^{(0)}\in\R}\left[
        \frac{(u^{(0)})^2}{2\Delta_L\chi^{(0)}} + l\left(
            y^{(0)},\sigma\left(
                h_u^{(0)} + u^{(0)}
            \right)
        \right)
    \right],
    \label{eq:rs-saddle-uhat0}
    \\
    h_u^{(0)} &= (2y^{(0)}-1)m^{(0)} + B^{(0)} + \sqrt{q^{(0)}}\xi_u^{(0)}, \quad \xi_u^{(0)}\sim\gN(0,1).
    \label{eq:rs-saddle-local-field0}
\end{align}
Then, the self-consistent equation for $\Theta^{(0)}$ and $\hat{\Theta}^{(0)}$ are given as follows:
\begin{align}
    q^{(0)} &= \frac{
        (\hat{m}^{(0)})^2 + \hat{\chi}^{(0)}
    }{(\hat{Q}^{(0)} + \lambda^{(0)})^2},
    \label{eq:rs-saddle-q0}
    \\
    \chi^{(0)} &=\frac{
        1
    }{
        \hat{Q}^{(0)} + \lambda^{(0)}
    },
    \label{eq:rs-saddle-chi0}
    \\
    m^{(0)} &=\frac{
        \hat{m}^{(0)}
    }{
        \hat{Q}^{(0)} + \lambda^{(0)}
    },
    \label{eq:rs-saddle-m0}
    \\
    \hat{Q}^{(0)} &= \alpha_L\Delta_L\E_{\xi_u^{(0)}\sim\gN(0,\Delta_L), y^{(0)}\sim p_{y,L}}\left[
        \frac{d }{d h_u^{(0)}}\partial_2 l_L(y^{(0)}, h_u^{(0)} + \usf{0})
    \right],
    \label{eq:rs-saddle-qhat0}
    \\
    \hat{\chi}^{(0)} &= \alpha_L\Delta_L\E_{\xi_u^{(0)}\sim\gN(0,\Delta_L), y^{(0)}\sim p_{y,L}}\left[
            \left(\partial_2 l_L(y^{(0)}, h_u^{(0)} + \usf{0})\right)^2
         \right],
     \label{eq:rs-saddle-chihat0}
    \\
    \hat{m}^{(0)} &= \alpha_L \E_{\xi_u^{(0)}\sim\gN(0,\Delta_L), y^{(0)}\sim p_{y,L}}\left[
            (2y-1)\partial_2 l_L(y^{(0)}, h_u^{(0)} + \usf{0})
         \right],
    \label{eq:rs-saddle-mhat0}
    \\
    0 &= \E_{\xi_u^{(0)}\sim\gN(0,\Delta_L), y^{(0)}\sim p_{y,L}}\left[
            \partial_2 l_L(y^{(0)}, h_u^{(0)} + \usf{0})
         \right],
    \label{eq:rs-saddle-b0}
\end{align}
Similarly, let $\usf{t}$ be the solution of the following randomized optimization problem:
\begin{align}
    \usf{t} &= \arg\min_{u^{(t)}\in\R}\left[
        \frac{(u^{(t)})^2}{2\Delta_U\chi^{(t)}} + \1(|\tilde{h}_u^{(t)}|>\Gamma \sqrt{q^{(t-1)}})l_{\rm pl}\left(
            \sigma_{\rm pl}\left(
                \tilde{h}_u^{(t)}
            \right),
            \sigma\left(
                h^{(t)}_u +  u^{(t)}
            \right)
        \right)
    \right]
    \label{eq:rs-saddle-uhat},
    \\
    \tilde{h}_u^{(t)} &= (2y^{(t)}-1)m^{(t-1)} + B^{(t-1)} + \sqrt{ q^{(t-1)}}\xi_{u,1}^{(t)},
    \label{eq:rs-saddle-local-field1}
    \\
    h_u^{(t)} &= (2y^{(t)}-1)m^{(t)} + B^{(t)} + \frac{R^{(t)}}{\sqrt{q^{(t-1)}}}\xi_{u,1}^{(t)} + \sqrt{q^{(t)} - \frac{(R^{(t)})^2}{q^{(t-1)}}}\xi_{u,2}^{(t)},
    \label{eq:rs-saddle-local-field2}
    \\
    &\xi_{u,1}^{(t)}, \xi_{u,2}^{(t)}\sim\gN(0,1).
\end{align}
Then, the self-consistent equations for $\Theta^{(t)}, \hat{\Theta}^{(t)}, t\in[T]$ are given as follows:
\begin{align}
    q^{(t)} &= \frac{
        (\hatm t)^2 + \hatchi t + (\hatr t)^2 q^{(t-1)}
        + 2 \hatm t \hatr t m^{(t-1)}
    }{
        (\hatq t + \lambda^{(t)})^2
    },
    \label{eq:rs-saddle-q}
    \\
    \chi^{(t)} &=\frac{
        1
    }{
        \hatq t + \lambda^{(t)}
    },
    \label{eq:rs-saddle-chi}
    \\
    m^{(t)} &=\frac{
        \hatm t + \hatr t m^{(t-1)}
    }{
        \hatq t + \lambda^{(t)}
    },
    \label{eq:rs-saddle-m}
    \\
    R^{(t)} &= \frac{
        \hatm t m^{(t-1)} + \hatr t q^{(t-1)}
    }{
        \hatq t + \lambda^{(t)}
    }
    ,
    \label{eq:rs-saddle-r}
    \\
\hat{Q}^{(t)} &= \alpha_U^{(t)}\Delta_U^{(t)}\E_{\xi_{u,1}^{(t)}, \xi_{u,2}^{(t)}\sim\gN(0,\Delta_U^{(t)}), y^{(t)}\sim p_{y}^{(t)}}\left[
        \frac{d }{d h_u^{(t)}}\partial_2 l_U(\tilde{h}_u^{(t)}, h_u^{(t)} + \usf{t};\Gamma\sqrt{q^{(t-1)}}) 
    \right],
    \label{eq:rs-saddle-hatq}
    \\
    \hat{\chi}^{(t)} &=  \alpha_U^{(t)}\Delta_U^{(t)} \E_{\xi_{u,1}^{(t)}, \xi_{u,2}^{(t)}\sim\gN(0,\Delta_U^{(t)}), y^{(t)}\sim p_{y}^{(t)}}\left[
            \left(
                \partial_2 l_U(\tilde{h}_u^{(t)}, h_u^{(t)} + \usf{t};\Gamma\sqrt{q^{(t-1)}})
            \right)^2
         \right],
         \label{eq:rs-saddle-hatchi}
    \\
    \hat{m}^{(t)} &= \alpha_U^{(t)} \E_{\xi_{u,1}^{(t)}, \xi_{u,2}^{(t)}\sim\gN(0,\Delta_U^{(t)}), y^{(t)}\sim p_{y}^{(t)}}\left[
            (2y^{(t)}-1)\partial_2 l_U(\tilde{h}_u^{(t)}, h_u^{(t)} + \usf{t};\Gamma\sqrt{q^{(t-1)}})
         \right],
         \label{eq:rs-saddle-hatm}
        \\
    \hat{R}^{(t)} &= -\alpha_U^{(t)}\Delta_U^{(t)}\E_{\xi_{u,1}^{(t)}, \xi_{u,2}^{(t)}\sim\gN(0,\Delta_U^{(t)}), y^{(t)}\sim p_{y}^{(t)}}\left[
        \frac{d }{d \tilde{h}_u^{(t)}}\partial_2 l_U(\tilde{h}_u^{(t)}, h_u^{(t)} + \usf{t};\Gamma\sqrt{q^{(t-1)}})
    \right]
    \label{eq:rs-saddle-hatr}
    \\
    0 &= \E_{\xi_{u,1}^{(t)}, \xi_{u,2}^{(t)}\sim\gN(0,\Delta_U^{(t)}), y^{(t)}\sim p_{y}^{(t)}}\left[
            \partial_2 l_U(\tilde{h}_u^{(t)}, h_u^{(t)} + \usf{t};\Gamma\sqrt{q^{(t-1)}})
         \right],
    \label{eq:rs-saddle-b}
\end{align}
\end{definition}

\subsubsection{Generating functional for the model parameter}
The solution of the above self-consistent equations gives the RS solution of the generating functional \eqref{eq: generating functional}. 
\begin{prediction}
\label{pred: rs generating functional w}
Let $\hat{\Theta}^{(t)}, t\ge 0$ be the solution of the self-consistent equations in Definition \ref{def: self-consistent equations}. Then, the generating functional \eqref{eq: generating functional} is given as follows:
\begin{align}
    \Xi_{\rm ST}(\epsilon_w, \epsilon_B) &= \E_{\xi_w^{(t)} \sim_{\rm iid}\gN(0,1)}\left[
        e^{\epsilon_w g_w(\{\wsf{t}\}_{t=0}^T)}
    \right]
    e^{\epsilon_B g_B(\{B^{(t)}\}_{t=0}^T)},
\end{align}
where $\{\wsf t\}_{t=0}^T$ follows the following Gaussian process:
\begin{equation}
    \wsf{0} = \frac{\hat{m}^{(0)}+\sqrt{\hat{\chi}^{(0)}}\xi_w^{(0)}}{\hat{Q}^{(0)}+\lambda^{(0)}},
    \quad
    \wsf{t} = \frac{\hat{m}^{(t)} + \hat{R}^{(t)}\wsf{t-1} +\sqrt{\hat{\chi}^{(t)}}\xi_w^{(t)}}{\hat{Q}^{(t)}+\lambda^{(t)}}.
    \label{eq: GP of w}
\end{equation}
\end{prediction}

In addition, using the formulae of the generating functional \eqref{eq: generating functional derivative w} and \eqref{eq: generating functional derivative B}, we can see that the averaged quantities regarding the weight and the bias are determined as follows.
\begin{prediction}
\label{pred: w effective average}
Let $\{\wsf t\}_{t=0}^T$ be the trajectory of the Gaussian process \eqref{eq: GP of w} in Prediction \ref{pred: rs generating functional w}, and $\{B^{(t)}\}_{t=0}^T$ be the solution of self-consistent equation in Definition \ref{def: self-consistent equations}. Then, the averaged quantities $\E_D[g_w(\{\hat{w}_i^{(t)}\})], i\in[N]$ and $\E_D[g_B(\{\hat{B}^{(t)}\})]$ are obtained as follows:
\begin{align}
    \E_D[g_w(\{\hat{w}_i^{(t)}\})] &= \E_{\xi_w^{(t)}\sim_{\rm iid}\gN(0,1)}\left[
        g_w(\{\wsf t\}_{t=0}^T)
    \right],
    \\
    \E_D[g_B(\{\hat{B}^{(t)}\})] &= g_B(\{B^{(t)}\}_{t=0}^T).
\end{align}
Hence, the next result also follows:
\begin{equation}
    \E_D[\frac{1}{N}\sum_{i=1}^N g_w(\{\hat{w}_i^{(t)}\})] = \E_{\xi_w^{(t)}\sim_{\rm iid}\gN(0,1)}\left[
        g_w(\{\wsf t\}_{t=0}^T)
    \right].
\end{equation}
\end{prediction}
This prediction indicates that each element of the weight vectors $\{\hat{w}_i^{(t)}\}_{t=0}^T, i\in[N]$ is statistically equivalent to the random variables $\{\wsf t\}_{t=0}^T$, which behaves as an effective surrogate for $\{\hat{w}_i^{(t)}\}_{t=0}^T$, i.e., $\{\hat{w}_i^{(t)}\}\deq \wsf{t}$. Here, it is understood that $\xi_w^{(t)}$ effectively plays the role of randomness coming from data $D$. In this effective description, the parameter $\hat{\Theta}^{(t)}$ has clear meanings; (i) $\hat{m}^{(t)}$ describes the correlation with the direction of $\bm{v}=(1,\dots, 1)$, (ii) $\hat{R}^{(t)}$ describes the amount of the information propagation from the previous step, (iii) $\hat{\chi}^{(t)}$ is the strength of the noise, and (iv) $\hat{Q}^{(t)}$ is the amount of confidence. Schematically, it can be represented as follows:
\begin{equation}
    \wsf{t} = \remark{\rm confidence}{\frac{1}{\hat{Q}^{(t)} + \lambda_U}}\left(
        \remark{\rm signal}{\hat{m}^{(t)}}
        + \remark{\rm noise}{\sqrt{\hat{\chi}^{(t)}}\xi_w^{(t)}}
        + \remark{\substack{\rm information \\ \rm propagation}}{\hat{R}^{(t)}\wsf{t-1}}
    \right).
    \label{eq: effective w intuitive}
\end{equation}
\inred{
    Hence, the statistical properties of the components of the high-dimensional weight vector are actually described by a Gaussian process that is governed by a finite number of parameters. These parameters are determined by the self-consistent equations in Definition \ref{def: self-consistent equations}. This finite-dimensional description enables us to evaluate the generalization error in the asymptotic limit efficiently, and to search for the optimal hyperparameters. Such a hyperparameter optimization is computationally difficult in finite-size simulations, because it requires a black-box optimization, such as the Nelder-Mead method, applied to results that are averaged over many realizations of the dataset.
}

\paragraph{\inred{Remark:}} Although the above formula is for averaged quantities, we expect that the values of macroscopic quantities of the form $\frac{1}{N}\sum_{i=1}^N g(\hat{w}_i^{(T)})$, such as the quantities in \eqref{eq:macro quantities}, and the biases will concentrate at LSL; in other words, their variances vanish in LSL.  Although we do not prove this concentration property, such concentration properties have been rigorously proven in analyses of convex optimization or Bayes-optimal inferences, such as the logistic regression \citep{mignacco2020role}, Bayes-optimal the generalized linear estimation \citep{barbier2019optimal}.  Since the optimization at each step of ST is convex, we expect the concentration property in our setting, too. 

Under the above assumption of concentration, the parameter $\Theta$ that is the solution of the self-consistent equations can be interpreted as follows. Let $\hat{\bm{\theta}}_{\epsilon_i}^{(t)}$ be the estimator obtained when a small perturbation is added to the original cost function:
\begin{equation}
    \hat{\bm{\theta}}^{(t)}_{\epsilon_i} = \argmin_{\bm{\theta}^{(t)}} \gL^{(t)}(\bm{\theta}^{(t)}|D_U^{(t)}, \hat{\bm{\theta}}^{(t-1)}) + \epsilon_i w_i^{(t)},
\end{equation}
\inred{where the additional term $\epsilon_i w_i^{(t)}$ is included to measure the stability of the solution $\hat{\bm{\theta}}^{(t)}$ at $\epsilon_i=0$.} Then $q^{(t)}, \chi^{(t)}, m^{(t)}$, and $R^{(t)}$ can be interpreted as follows:
\begin{align}
    q^{(t)} &= \lim_{N\to\infty} \frac{1}{N}\|\hat{\bm{w}}^{(t)}\|_2^2,
    \label{eq: meaning of q}
    \\
    \chi^{(t)} &= \lim_{N,\beta\to\infty}\frac{1}{N}\sum_{i=1}^N\beta^{(t)}\sV{\rm ar}_{p_{\rm ST}}\left[
        w_i^{(t)}
    \right] 
    \\
    &=\lim_{N\to\infty, \epsilon_i\to0} \frac{1}{N}\sum_{i=1}^N \frac{\partial \hat{w}_{\epsilon_i,i}^{(t)}}{\partial \epsilon_i},
    \label{eq: meaning of chi}
    \\
    m^{(t)} &= \lim_{N\to\infty} \frac{1}{N}\bm{v}\cdot \hat{\bm{w}}^{(t)},
    \label{eq: meaning of m}
    \\
    R^{(t)} &= \lim_{N\to\infty} \frac{1}{N}\hat{\bm{w}}^{(t-1)}\cdot \hat{\bm{w}}^{(t)},
    \label{eq: meaning of R}
\end{align}
The interpretation of $q^{(t)}, m^{(t)}$ and $R^{(t)}$ is clear from the right-hand-sides of these equations. The interpretation of $\chi^{(t)}$ may be slightly non-trivial, but it can be understood as representing how the learning result of the weight vector is stable against a small perturbation to the loss function. We shall refer to $\chi^{(t)}$ as {\it linear susceptibility} following the custom of statistical mechanics.

\subsubsection{Generalization error}
Using the solution of the self-consistent equations, the generalization error \eqref{eq: gen_err} are obtained as follows.
\begin{prediction}
\label{pred: generalization error}
Let $q^{(t)}, m^{(t)}, B^{(t)}, t=0,\dots,T$ be the solution of the self-consistent equations in Definition \ref{def: self-consistent equations}. Then, at LSL, the macroscopic quantities \eqref{eq:macro quantities} and the generalization error \eqref{eq: gen_err} is are obtained as follows:
\begin{align}
    \hat{B}^{(t)} &= B^{(t)},
    \label{eq: meaning of B}
    \\
    \bar{\epsilon}_{\rm g}^{(t)} &= \epsilon_{\rm g}^{(t)} \equiv \rho_U H\left(
        \frac{
            {m}^{(t)} + {B}^{(t)}
        }{
            \sqrt{\Delta_U {q}^{(t)}}
        }
    \right) + (1-\rho_U)H\left(
        \frac{
            {m}^{(t)} - {B}^{(t)}
        }{
            \sqrt{\Delta_U {q}^{(t)}}
        }
    \right)
    \label{eq: rs generalization error}.
\end{align}
\end{prediction}
Thus, by numerically solving the self-consistent equations \eqref{eq:rs-saddle-q0}-\eqref{eq:rs-saddle-uhat}, we can precisely evaluate the generalization error \eqref{eq: gen_err} in LSL with a limited number of variables. Notice that now the problem is finite-dimensional. In Appendix \ref{app: numerical recipe}, we sketch how to obtain numerical solutions of the self-consistent equations. Also, the cosine similarity between the direction of the classification plane $\hat{\bm{w}}^{(t)}$ and the direction of the cluster center $\bm{v}$ can be evaluated by using the above formula as 
\begin{equation}
    \frac{\hat{\bm{w}}^{(t)}\cdot \bm{v}}{\|\hat{\bm{w}}^{(t)}\|_2\|\bm{v}\|_2} \to \frac{m^{(t)}}{\sqrt{q^{(t)}}}.
    \label{eq: cosine similarity}
\end{equation}

\subsubsection{Generating functional for the logits}
As in the previous discussion of weight vectors, we can characterize the statistical properties regarding $y_\nu^{(t)}$, $\tilde{u}_{\nu}=\bm{x}_{\nu}^{(t)}\cdot\bm{w}^{(t-1)}/\sqrt{N} + {B}^{(t-1)}$ and $u_{\nu}=\bm{x}_{\nu}^{(t)}\cdot\bm{w}^{(t)}/\sqrt{N} + {B}^{(t)}$ by computing another generating functional \eqref{eq: generating functional logits}.

\begin{prediction}
\label{pred: u effective average}
Let $\usf{t}$ be the solution of the one-dimensional optimization problem in \eqref{eq:rs-saddle-uhat} otherwise. Also, $\tilde{h}_{u}^{(t)}$ and $h_u^{(t)}$ be the quantity defined in \eqref{eq:rs-saddle-local-field1} and \eqref{eq:rs-saddle-local-field2}, respectively. Then, the averaged quantities $\E[g_u(\{(y_\nu^{(t)}, \tilde{u}_{\nu}^{(t)}, u_{\nu}^{(t)})\}_{t=1}^{T})]$ can be evaluated as follows:
\begin{equation}
    \E_D\left[
         g_u(\{(y_\nu^{(t)}, \tilde{u}_{\nu}^{(t)}, u_{\nu}^{(t)})\}_{t=1}^{T})
    \right] = \E_{\{y^{(t)},\xi_{u,1}^{(t)}, \xi_{u,2}^{(t)}\}_{t=1}^T}\left[
        g_u\left(
            \{(
                y^{(t)},
                \tilde{h}_u^{(t)},
                \usf{t} + h_u^{(t)}
            )\}_{t=1}^T
        \right)
    \right].
\end{equation}
Hence, the next result also follows:
\begin{equation}
    \E_D[\frac{1}{M_U}\sum_{\nu=1}^{M_U} g_u(\{(y_\nu^{(t)}, \tilde{u}_{\nu}^{(t)}, u_{\nu}^{(t)})\}_{t=1}^{T})] = \E_{\{y^{(t)},\xi_{u,1}^{(t)}, \xi_{u,2}^{(t)}\}_{t=1}^T}\left[
        g_u\left(
            \{(
                y^{(t)},
                \tilde{h}_u^{(t)},
                \usf{t} + h_u^{(t)}
            )\}_{t=1}^T
        \right)
    \right].
 \end{equation}
Here $g_u$ is arbitrary as long as the above integral is convergent.
\end{prediction}

This prediction indicates that $\usf{t}, \tilde{h}_u^{(t)}$ and $h_u^{(t)}$ effectively describe the statistical properties of the logits, i.e., $\{\hat{\bm{w}}^{(t)}\cdot \bm{x}_\nu^{(t)}/\sqrt{N}+\hat{B}^{(t)}\}_{\nu=1}^{M_U}\deq \usf{t} + h_u^{(t)}$\inred{, similarly to the finite-dimensional description of the weight vector in Predictions \ref{pred: rs generating functional w} and \ref{pred: w effective average}}. The minimization problem \eqref{eq:rs-saddle-uhat}, which determines the value of $\usf{t}$, can be interpreted as an effective surrogate to determine the distribution of the logits. \inred{Schematically, it can be represented as follows:
\begin{align}
    \tilde{h}_u^{(t)} &= \remark{\rm bias}{B^{(t-1)}} 
    + \remark{\substack{\rm correlation~with \\ \rm the~cluster~center}}{(2y^{(t)}-1)m^{(t-1)}}
    + \remark{\substack{\rm fluctuation~in \\ \rm input~points}}{\sqrt{q^{(t-1)}}\xi_{u,1}^{(t)}},
    \label{eq: intuitive hu1}
    \\
    h_u^{(t)} &= \remark{\rm bias}{B^{(t)}} 
    + \remark{\substack{\rm correlation~with \\ \rm the~cluster~center}}{(2y^{(t)}-1)m^{(t)}}
    + \remark{\substack{\rm correlation~with \\ \rm the~prediction}}{\frac{R^{(t)}}{\sqrt{q^{(t-1)}}}\xi_{u,1}^{(t)}}
    + \remark{\substack{\rm fluctuation~in \\ \rm input~points}~+~\substack{\rm noise~induced~at~a~ \\ \rm parameter~update}}{\sqrt{q^{(t)}-\frac{(R^{(t)})^2}{q^{(t-1)}}}\xi_{u,2}^{(t)}}.
    \label{eq: intuitive hu2}
\end{align}}
\inred{Here}, $\Theta^{(t)}$, $\xi_{u,1}^{(t)}$ and $\xi_{u,2}^{(t)}$ have clear interpretations. First, $B^{(t-1)}$ and $B^{(t)}$ represent the bias terms at step $t-1$ and $t$. Second, $(2y^{(t)}-1)m^{(t-1)}$ and $(2y^{(t)}-1)m^{(t)}$ represent correlation with correlation with the cluster center at step $t$. Thirdly, the term $\sqrt{q^{(t-1)}}\xi_{u,1}^{(t)}$ represents the fluctuation of the prediction. This indicates that the Gaussian random variable $\xi_{u,1}^{(t)}$ comes from the statistical variation of the input points of the training data used at step $t$. Also $R^{(t)}/\sqrt{q^{(t-1)}}\xi_{u,1}^{(t)}$ in $h_{u}^{(t)}$ represents the correlation with the prediction. Finally, $\sqrt{q^{(t)} - (R^{(t)})^2/q^{(t-1)}}\xi_{u,2}^{(t)}$ contains two kinds of fluctuations; (i) the fluctuation of the inputs and (ii) the noise that arises when updating the parameters $\bm{\theta}$ using a finite amount of training data in the sense that the ratio to the input dimension $N$ is finite. Note that the equations \eqref{eq: meaning of q} and \eqref{eq: meaning of R} imply $q^{(t)}=R^{(t)}=q^{(t-1)}$ when $\hat{\bm{\theta}}^{(t)} = \hat{\bm{\theta}}^{(t-1)}$. Hence the factor $\sqrt{q^{(t)} - (R^{(t)})^2/q^{(t-1)}}\xi_{u,2}^{(t)}$ makes no contribution when there is no update. By using these quantities, the logit $\usf{t} + h_{u}^{(t)}$ is determined by the one dimensional optimization problem \eqref{eq:rs-saddle-uhat}.

The predictions \ref{pred: rs generating functional w}-\ref{pred: u effective average} are the first main results of this paper.

\subsection{Cross-checking with numerical experiments}
\label{subsec:cross-check}
To check the validity of the predictions presented in the previous section, we briefly compare the result of numerical solution of the self-consistent equations with numerical experiments of finite-size systems. The details of the numerical treatment of the self-consistent equations are described in Appendix \ref{app: numerical recipe}.

\begin{figure}[t!]
    \centering
    \begin{subfigure}[b]{0.495\textwidth}
        \centering
        \includegraphics[width=\textwidth]{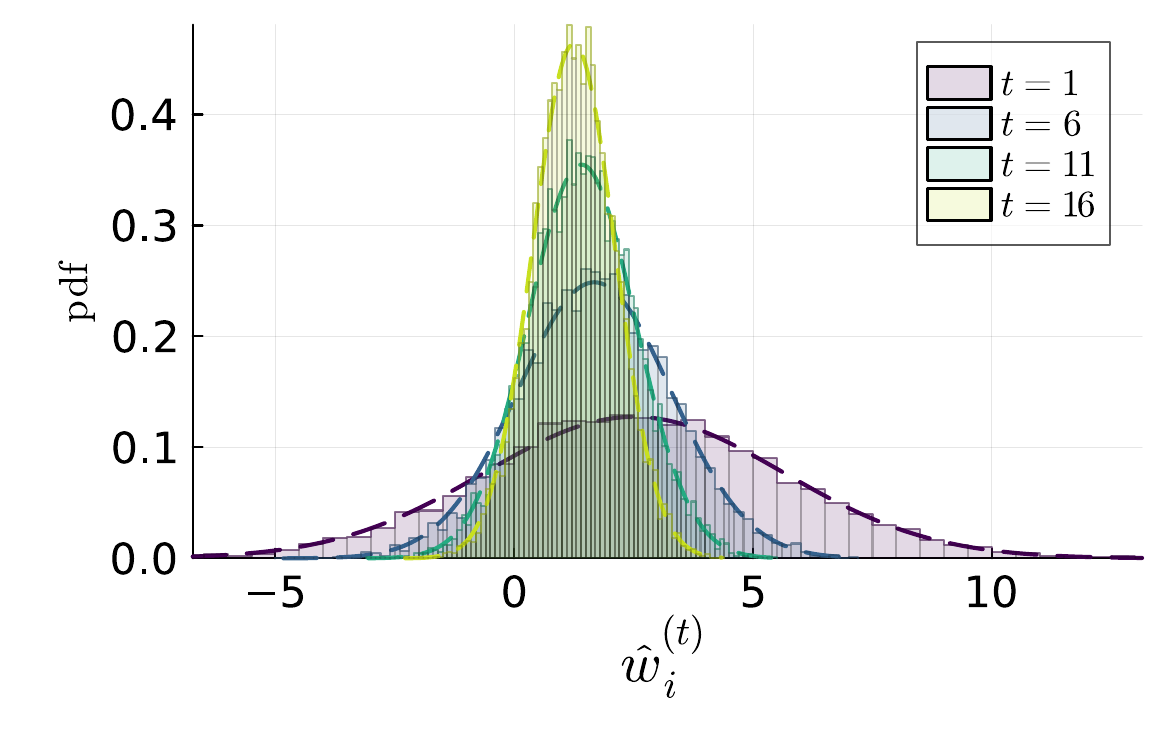}
        \caption{$\{\hat{w}_i^{(t)}\}: (N,\rho)=(2^{13}, 0.4)$}
    \end{subfigure}
    \begin{subfigure}[b]{0.495\textwidth}
        \centering
        \includegraphics[width=\textwidth]{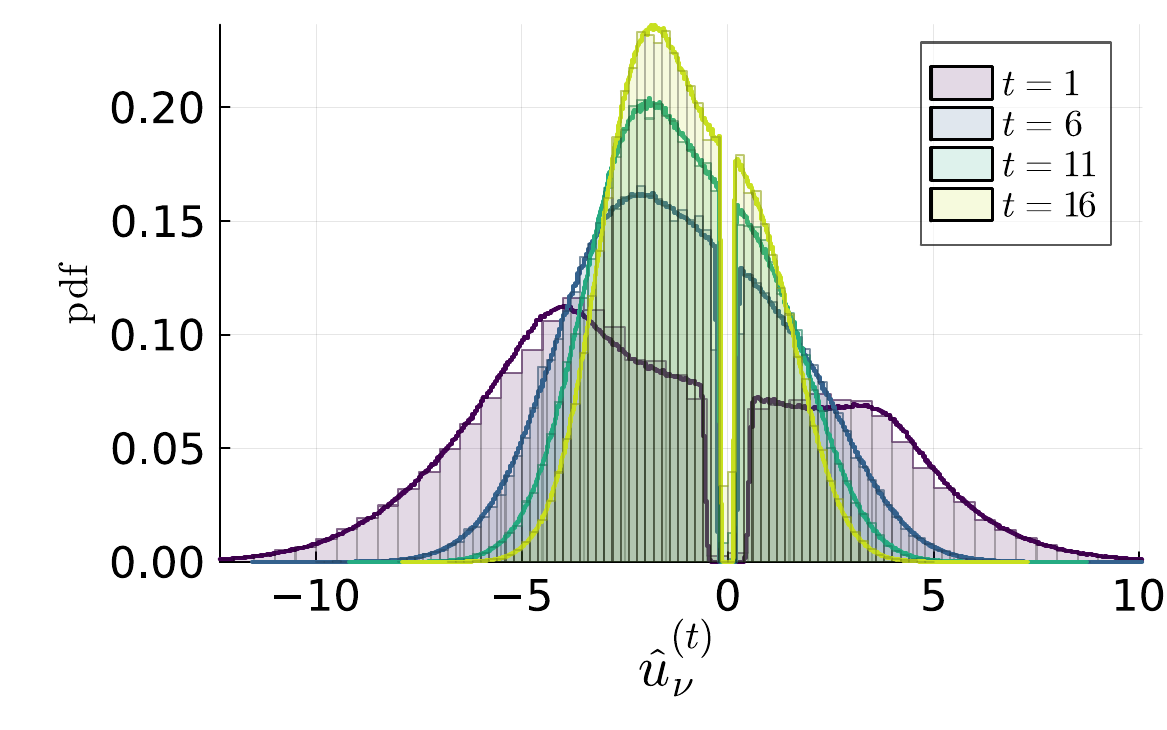}
        \caption{$\{\hat{u}_\nu^{(t)}\}: (N,\rho)=(2^{13}, 0.4)$}
    \end{subfigure}
    \caption{
        \inred{Representative comparison of the empirical distributions and the RS predictions for the learned weights and logits. The panels show the label-imbalanced case $\rho=0.4$ with $N=2^{13}$ and $T=16$. The histograms are obtained from finite-size numerical experiments, and the solid lines show the RS predictions.
        }
    }
    \label{fig: representative empirical dist with experiment}
\end{figure}

\begin{figure}[t!]
    \centering
    \begin{subfigure}[b]{0.495\textwidth}
        \centering
        \includegraphics[width=\textwidth]{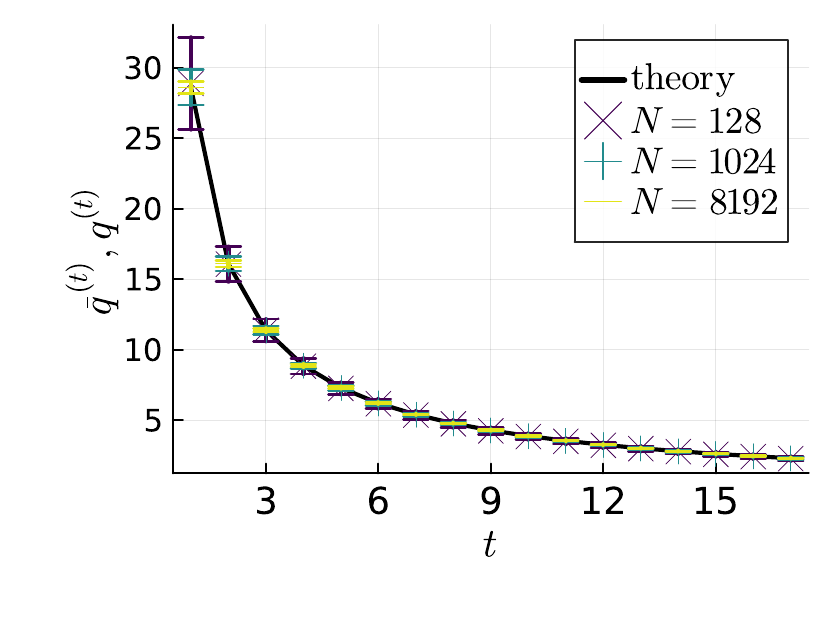}
        \caption{$q^{(t)}$}
    \end{subfigure}
    \begin{subfigure}[b]{0.495\textwidth}
        \centering
        \includegraphics[width=\textwidth]{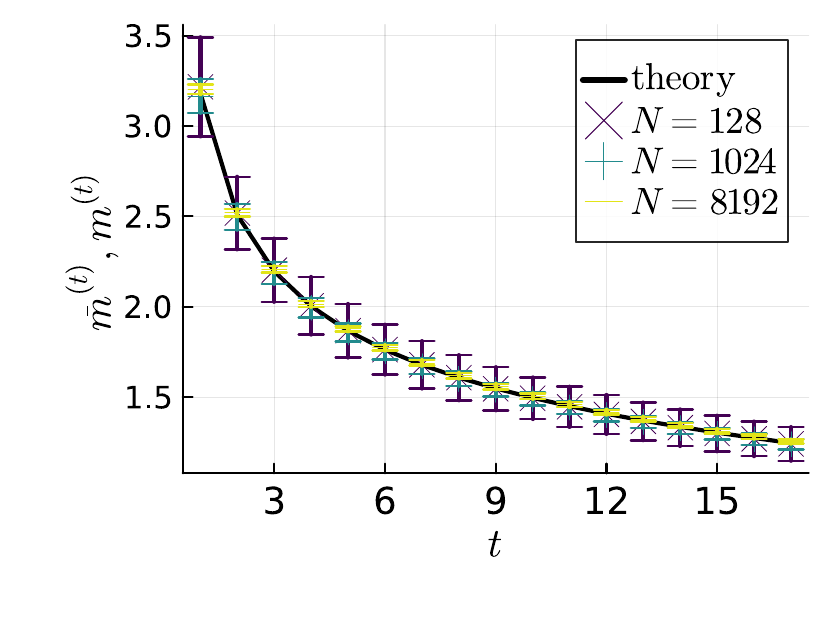}
        \caption{$m^{(t)}$}
    \end{subfigure}
    %
    \begin{subfigure}[b]{0.495\textwidth}
        \centering
        \includegraphics[width=\textwidth]{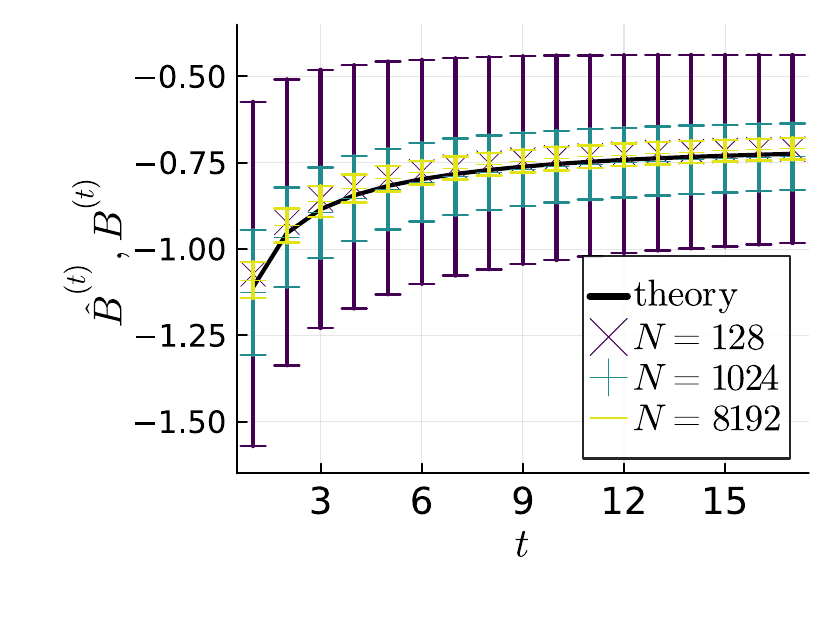}
        \caption{$B^{(t)}$}
    \end{subfigure}
    \begin{subfigure}[b]{0.495\textwidth}
        \centering
        \includegraphics[width=\textwidth]{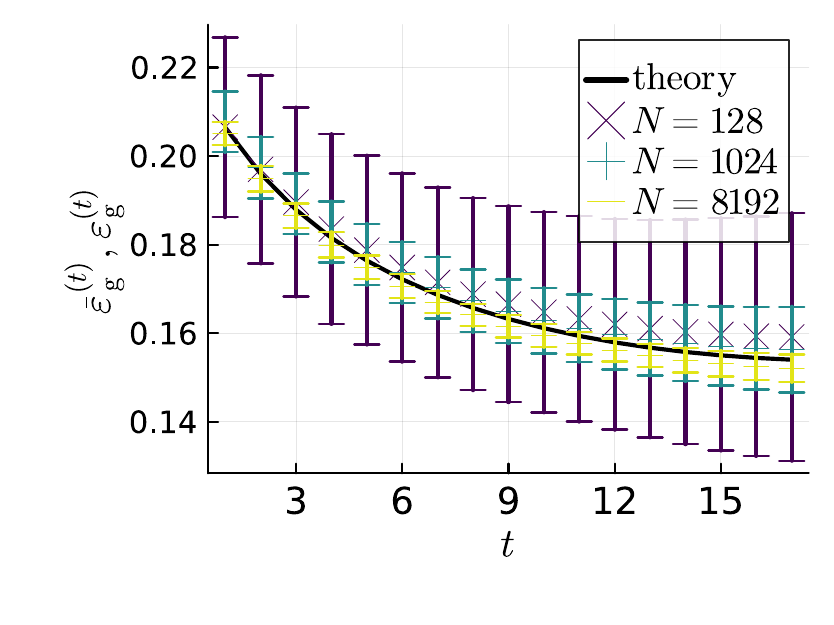}
        \caption{$\epsilon_{\rm g}^{(t)}$}
    \end{subfigure}
    \caption{
        \inred{
            Representative comparison of macroscopic quantities obtained from finite-size experiments and the RS predictions in the label-imbalanced case $\rho=0.4$. The markers with error bars show finite-size experiments, and the solid lines show the RS predictions.
        }
    }
    \label{fig: representative macro with experiment}
\end{figure}

For simplicity, we investigate the case without domain shift, i.e., $\rho_U = \rho_L = \rho$ and $\Delta_L = \Delta_U =\Delta$. Also, we focus on the case of $\lambda^{(0)}=\lambda_L, \lambda^{(t)}=\lambda_U={\rm Const.}$ for $t\ge1$ because tuning the regularization parameters in $\R^{T+1}$ is computationally demanding. The nonlinear function of the model, the pseudo-labeler, and the loss function are $\sigma(x)=\sigma_{\rm pl}(x)=1/(1+e^{-x})$ (sigmoid) and $l(p,q)=l_{\rm pl}(p,q)=-p\log q - (1-p)\log(1-q)$ (logistic loss), respectively. We optimize the regularization parameter $\lambda_L, \lambda_U$ and $\Gamma$ so that the generalization error $\epsilon_{\rm g}^{(T)}$ (the generalization error at the last step) is minimized by using the Nelder-Mead method in Optim.jl library \citep{mogensen2018optim}. Here, the total number of iterations is fixed as $T=16$, and the comparison at steps $t=1,6,11,16$ are shown. We remark that for small iteration cases with almost balanced clusters, such as $T=1$ and $\rho\simeq1/2$, the optimal regularization parameter often shows a diverging tendency as in the logistic regression \citep{Dobribal2018high, mignacco2020role} where infinitely large regularization yields the Bayes-optimal classifier. However, it is known that such a large regularization parameter induces a pathologically large finite-size effect as reported in \citep{mignacco2020role}. Therefore, we restrict the range of the regularization parameter as $\lambda_L, \lambda_U\in(0, 0.1)$. See Appendix \ref{app: pathological finite size effect} for more detail.

Figures~\ref{fig: representative empirical dist with experiment} and
\ref{fig: representative macro with experiment} show representative comparisons
between the RS predictions and finite-size numerical experiments.
We use the label-imbalanced setting $\rho=0.4$ as a representative
nontrivial case. When showing the histogram for logits, the contribution from these points are excluded from the figures, since the logits on the data points excluded by PLS are not detemined.  Additional comparisons for other values of $\rho$ and system sizes are provided in Appendix~\ref{app:
additional finite size checks}. 

Overall, the RS solution agrees well with the results of numerical experiments on finite size systems. Therefore, we expect that the RS solution exactly describes the statistical properties of the regressors obtained by ST. In the following section, we will use the RS solution to perform a more comprehensive analysis of the behavior of ST.

\section{Analysis of RS solution}
\label{section: analyzing RS solution}
In this section, using the RS solution obtained in the previous section, we study what specific properties of classifiers can be obtained.

First, in subsection \ref{subsec: numerical inspection}, to check the quantitative tendency, we numerically investigate how the generalization error depends on the label bias, the size of the unlabeled data used at each iteration, and the regularization parameter. Next, in subsection \ref{subsec: small regularization limit}, based on the finding of the numerical inspection, we analyze the behavior of ST at weak regularization limit $\lambda_U\ll 1$ in the under-parameterized setting. There, we find that the classification plane finds the best direction as long as the initial classifier is not completely uninformative, but the bias term may not be optimal in label imbalanced cases. 

\subsection{Numerical inspection}
\label{subsec: numerical inspection}

\begin{figure}[t!]
    \centering
    \begin{subfigure}[b]{\textwidth}
        \centering
        \includegraphics[width=0.95\linewidth]{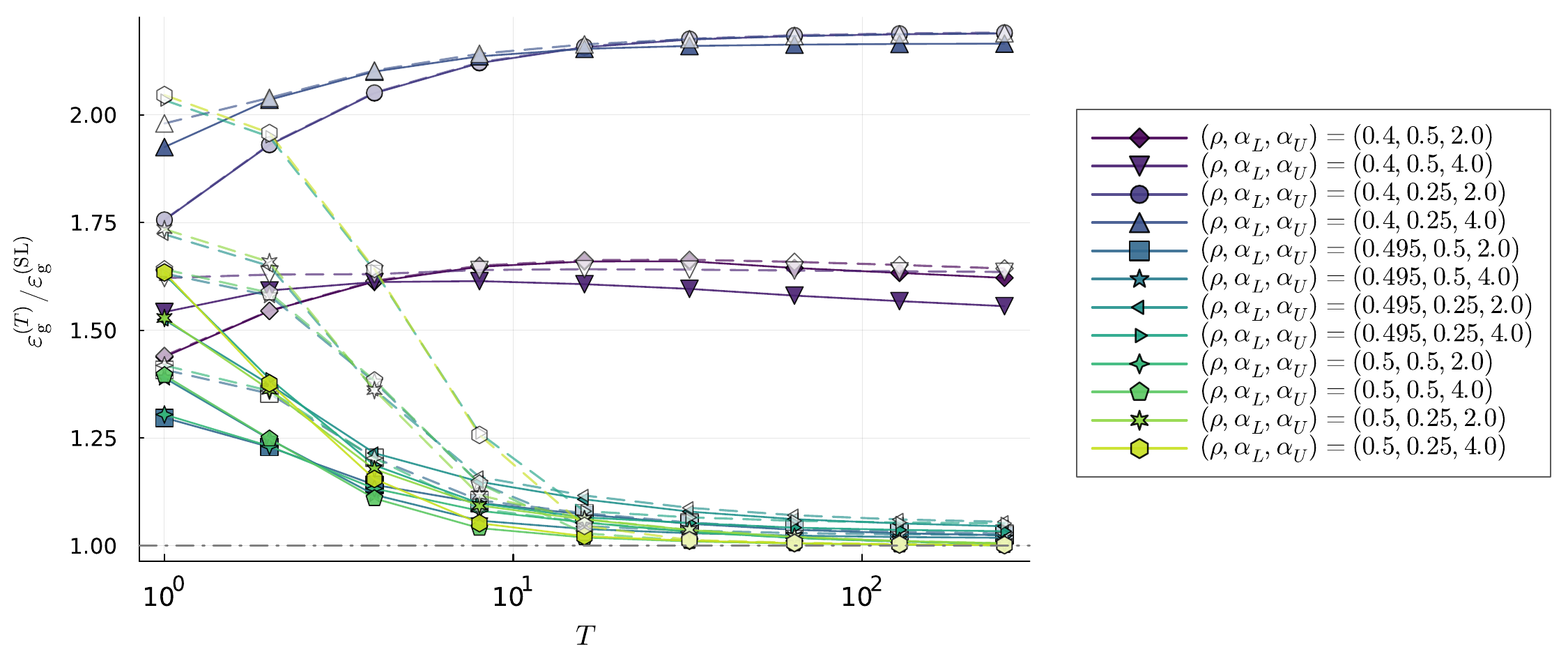}
        \caption{$\epsilon_{\rm g}^{(T)}/\epsilon_{\rm g}^{\rm (SL)}$ (relative generalization error)}
    \end{subfigure}
    \begin{subfigure}[b]{\textwidth}
        \centering
        \includegraphics[width=0.95\linewidth]{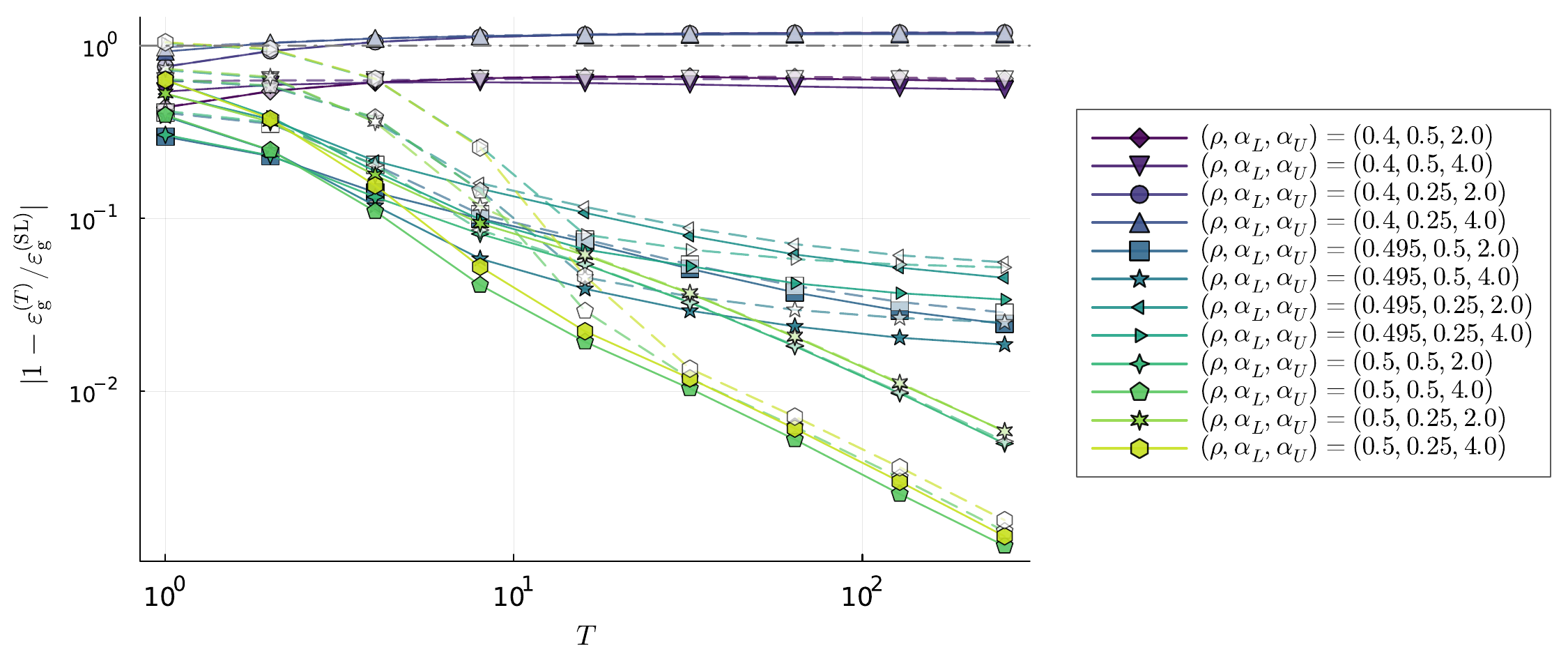}
        \caption{$|1 - \epsilon_{\rm g}^{(T)}/\epsilon_{\rm g}^{\rm (SL)}|$ (absolute value of deviation from unity)}
        \label{fig: relative generalization error absolute deviation}
    \end{subfigure}
    
    \caption{The ratio of the generalization error obtained at the end of ST \eqref{eq: rs generalization error} ($t=T$) to the SL with a labeled dataset of size $N(\alpha_L + \alpha_U \times T)$. When the ratio is unity, the ST with an unlabeled dataset has the same performance as the SL with labeled data of the same size. Different colors and lines represent different pairs of $(\rho, \Delta)$. The filled markers with solid lines represent the result with PLS, where $\Gamma$ is optimized so that $\epsilon_{\rm g}^{(T)}$ is minimized. The white markers with dashed lines represent the result without PLS ($\Gamma=0$). The upper panel shows the raw values. The lower panel shows their absolute values of the deviation from unity in the log scale. }
    \label{fig: relative generalization error}
\end{figure}

\paragraph{Setup:} To check quantitative tendency, we investigate the generalization error and macroscopic quantities in more detail. As in the previous subsection, we investigate the case $\rho_U = \rho_L = \rho$ and $\Delta_L = \Delta_U =\Delta$ (no domain shift). Also, we focus on the case of $\lambda^{(0)}=\lambda_L, \lambda^{(t)}=\lambda_U={\rm Const.}$ for $t\ge1$ because tuning the regularization parameters in $\R^{T+1}$ is computationally demanding. The nonlinear function of the model, the pseudo-labeler, and the loss function are $\sigma(x)=\sigma_{\rm pl}(x)=1/(1+e^{-x})$ (sigmoid) and $l(p,q)=l_{\rm pl}(p,q)=-p\log q - (1-p)\log(1-q)$ (logistic loss), respectively.  Also, the hyper parameters are optimized so that the generalization at the last step $T$ is minimized. In the following, we mainly focus on how the various quantities varies as the total number of iterations $T$ changes. Recall that $t\in[T]$ represents the iteration step of ST with the total number of iterations $T$, while $T$ represents the total number of iterations in ST.

\begin{figure}[t!]
    \centering
    \begin{subfigure}[b]{\textwidth}
        \centering
        \includegraphics[width=0.95\linewidth]{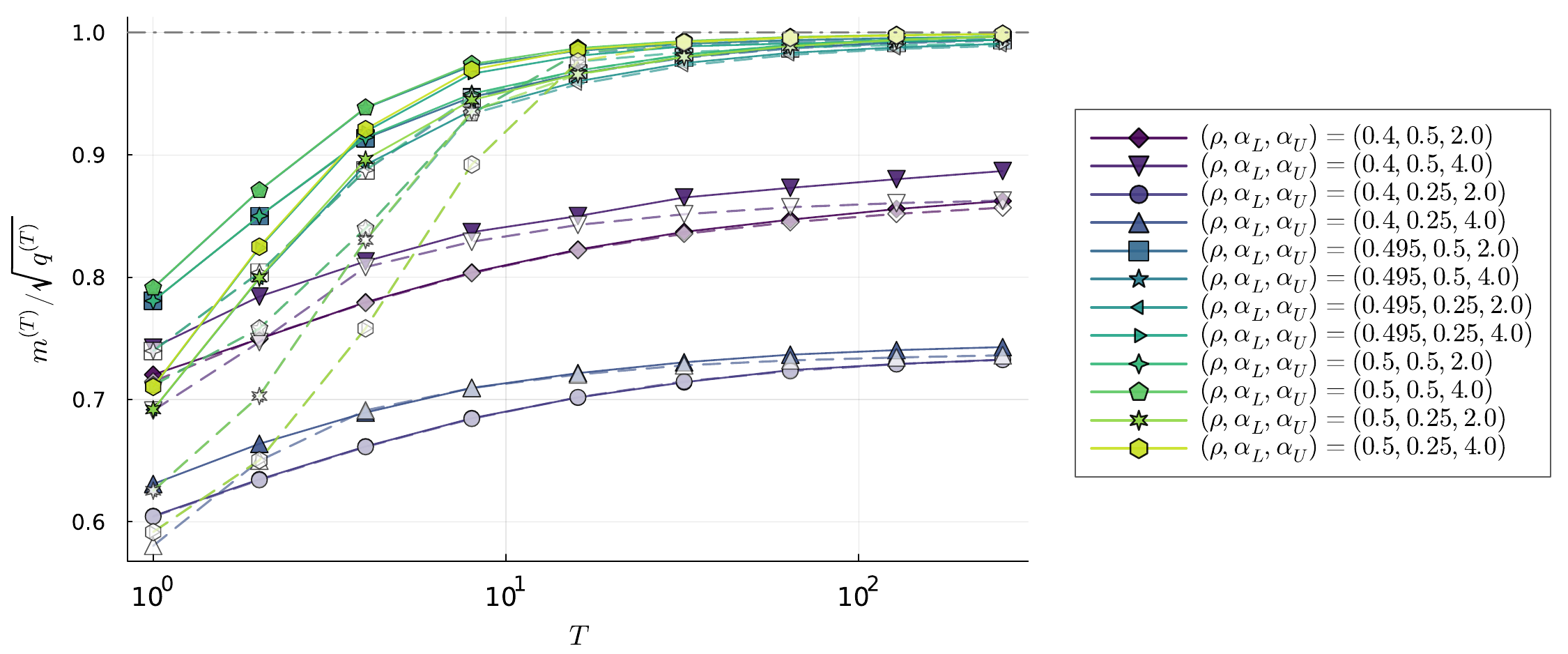}
        \caption{$m^{(t)}/\sqrt{q^{(T)}}$ (raw value)}
    \end{subfigure}
    \begin{subfigure}[b]{\textwidth}
        \centering
        \includegraphics[width=0.95\linewidth]{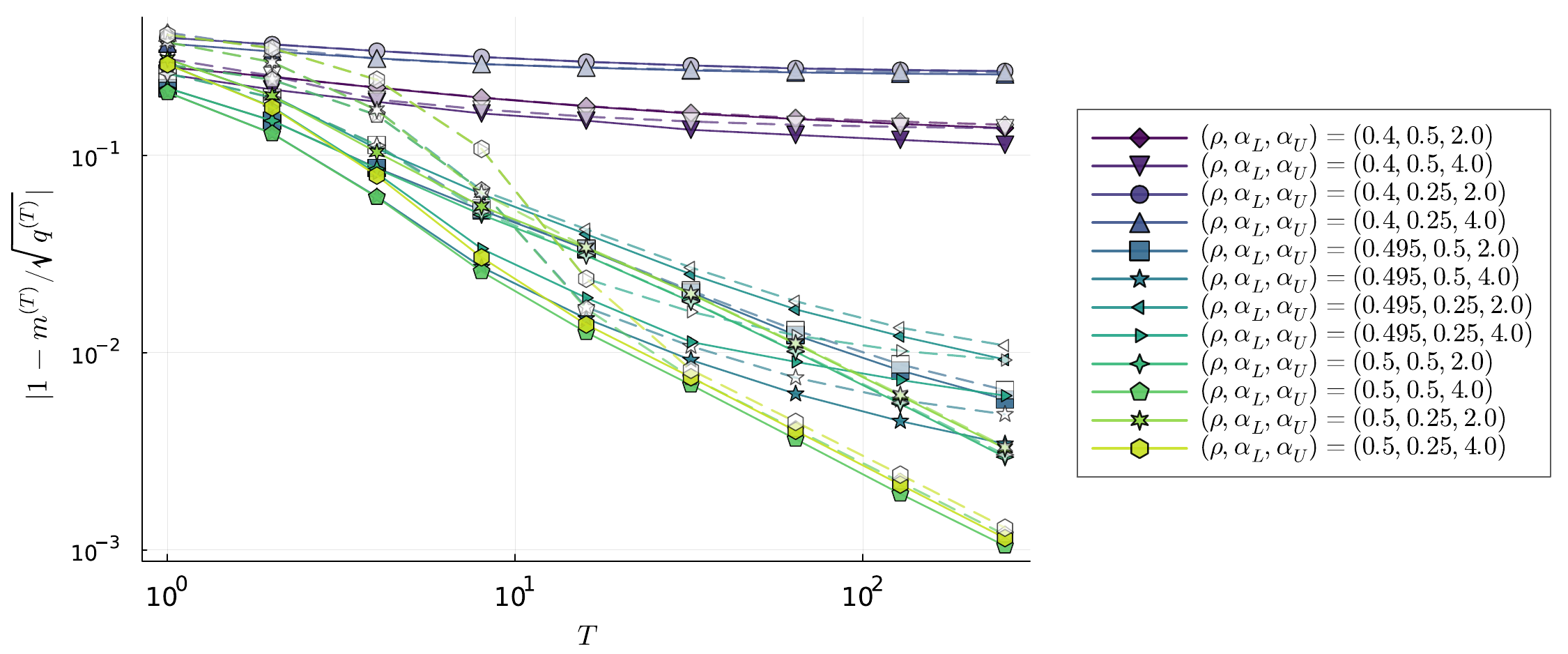}
        \caption{$|1 - m^{(t)}/\sqrt{q^{(T)}}|$ (absolute value of deviation from unity)}
        \label{fig: cosine similarity deviation from unity}
    \end{subfigure}
    \caption{The the cosine similarity \eqref{eq: cosine similarity} obtained at the end of ST \eqref{eq: rs generalization error} ($t=T$). When the value is unity, the classification plane is oriented in the optimal direction. Different colors and lines represent different pairs of $(\rho, \alpha_L, \alpha_U)$. The filled markers with solid lines represent the result with PLS, where $\Gamma$ is optimized so that $\epsilon_{\rm g}^{(T)}$ is minimized. The white markers with dashed lines represent the result without PLS ($\Gamma=0$). The upper panel shows the raw values. The lower panel shows their absolute values of the deviation from unity in the log scale. }
    \label{fig: cosine similarity}
\end{figure}

\begin{figure}[t!]
    \begin{subfigure}[b]{\textwidth}
        \includegraphics[width=0.95\linewidth]{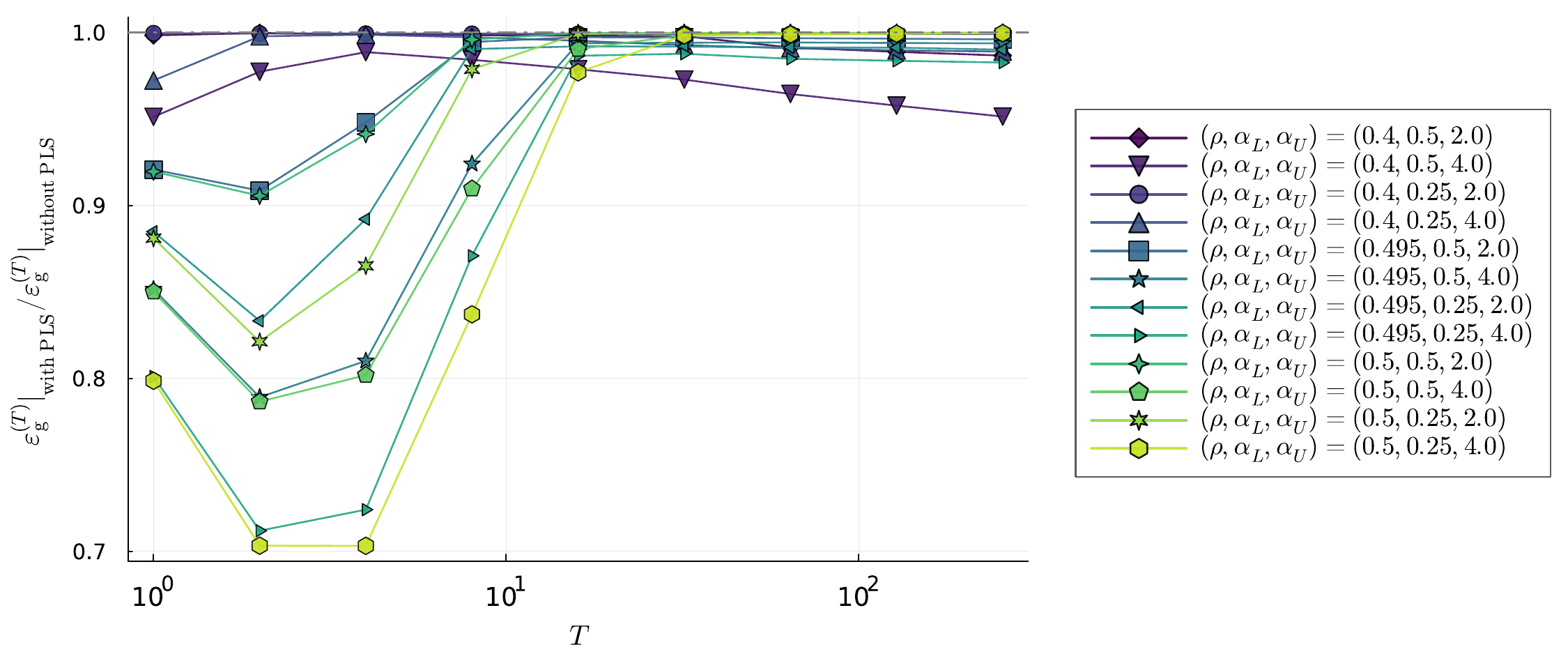}
        \caption{Including label imbalanced cases}
        \label{fig: vs PLS all}
    \end{subfigure}
    \begin{subfigure}[b]{\textwidth}
        \includegraphics[width=0.95\linewidth]{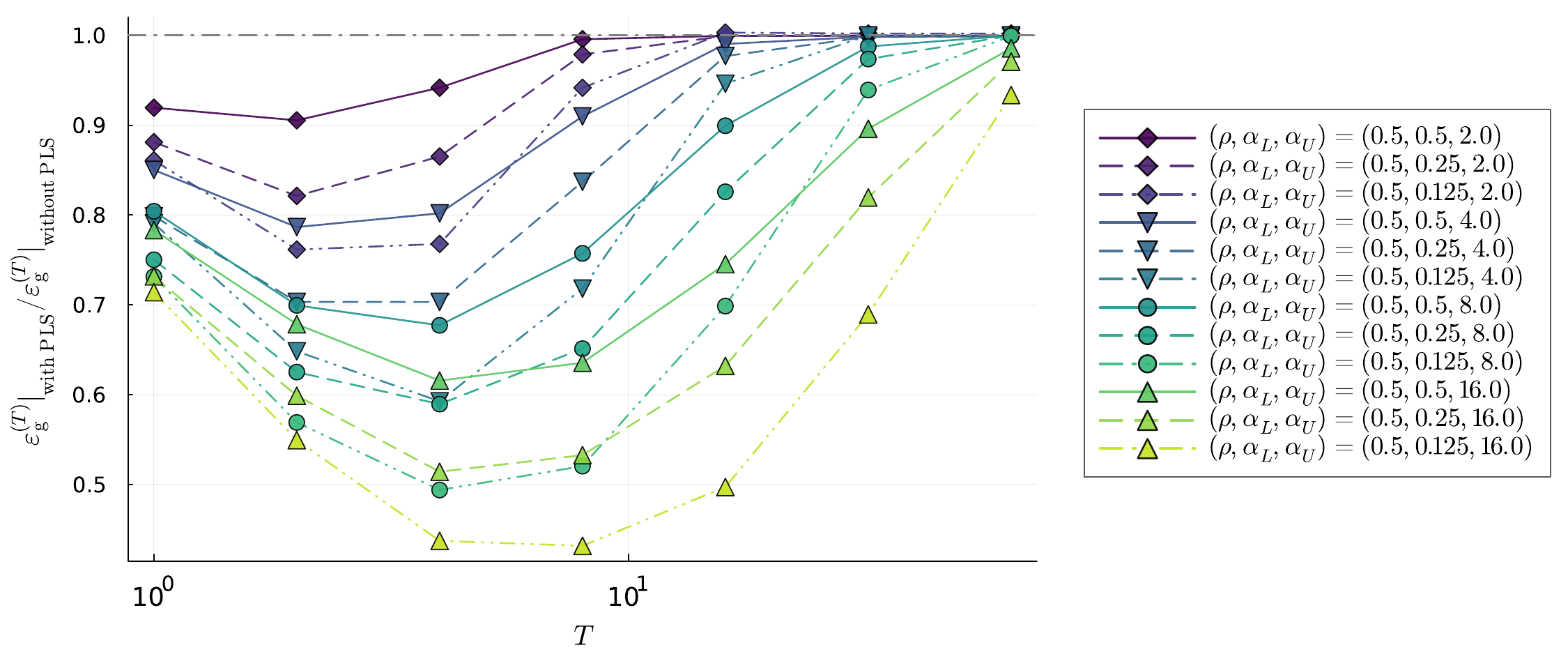}
        \caption{Only label balanced cases}
        \label{fig: vs PLS balanced}
    \end{subfigure}
    
    \caption{Comparison of generalization error with and without the implementation of PLS. Upper panel shows the result including label imbalanced cases. Lower panel shows the result of label balanced cases only. }
    \label{fig: vs PLS}
\end{figure}

\paragraph{Dependence on the label bias:} Figure \ref{fig: relative generalization error} shows the ratio of the generalization error obtained by ST \eqref{eq: rs generalization error} to the SL (ridge-regularized logistic regression) with a labeled dataset of size $N(\alpha_L + \alpha_U \times T)$. The generalization error of SL is obtained by the formula in \citep{mignacco2020role}. When the label imbalance is absent ($\rho=1/2$) and the total number of iterations $T$ is large, the performance of ST is almost compatible with the SL with the same data size but with the true label. This observation is compatible with previous studies \citep{oymak2020statistical, oymak2021theoretical, frei2022}. On the other hand, when there is a label imbalance, the performance of ST does not approach that of SL, even if we optimize the regularization parameter and the threshold of PLS. This point is more apparent in Figure \ref{fig: relative generalization error absolute deviation}. Although PLS improves the generalization error, especially when the total number of iterations are small (see Figure \ref{fig: vs PLS all}), ST's performance still significantly degrades as the label bias grows.  Similarly, Figure \ref{fig: cosine similarity} shows the cosine similarity \eqref{eq: cosine similarity} between the direction of the classification plane $\hat{\bm{w}}^{(t)}$ and the cluster center $\bm{v}$. The cosine similarity monotonically increases as the number of the total iteration grows. However, the growth is slow in label-biased cases.

\paragraph{Effect of PLS:} Figure \ref{fig: vs PLS} shows the comparison of the generalization error between the case with and without PLS. The upper panel show the result of both label balanced and imbalanced cases. When the total number of iterations $T$ is small, ST's performance is largely improved, which suggests that minimizing the cost function \eqref{eq: update of ST} has the intuitive effect of fitting to the pseudo-correct-{\it label}, \inred{as originally proposed in \citep{lee2013pseudo}. Hence,} just fitting the model to relatively reliable labels is appropriate. Recall that the pseudo-labels selected by PLS have relatively large input and closer to hard labels. This tendency is evident especially when labels are balanced ($\rho=1/2$). Detailed data for the label balanced cases is shown in Figure \ref{fig: vs PLS balanced}. As shown in this figure, PLS is effective especially when the initial classifier is poor ($\alpha_L$ is small) and the size of the unlabeled data used at each iteration is large ($\alpha_U$ is large).

\begin{figure}[t!]
    \centering
    \includegraphics[width=0.95\textwidth]{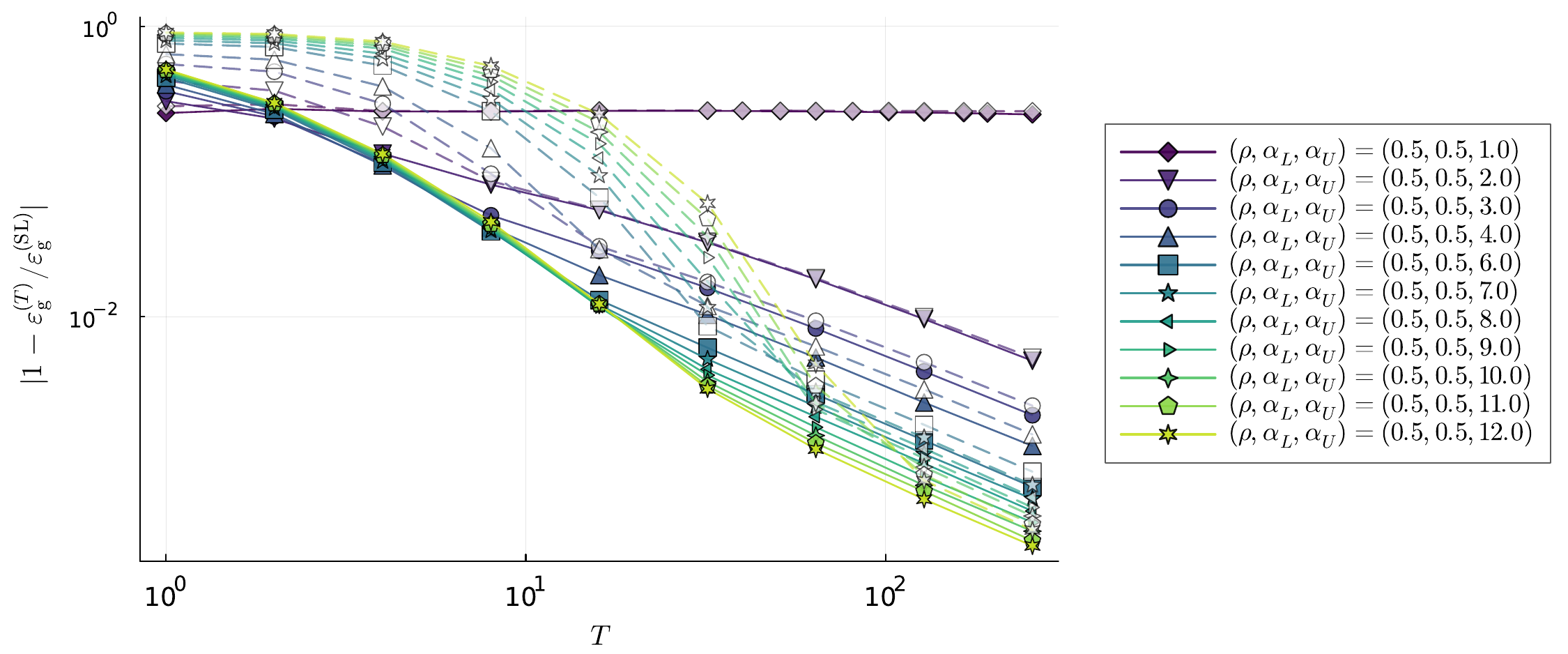}
    \caption{Relative generalization error $\epsilon_{\rm g}^{(T)}/\epsilon_{\rm g}^{\rm (ST)}$ when $(\rho, \Delta)=(1/2,0.75^2)$. Different colors represent different $\alpha_U$. }
    \label{fig: partition dependence balanced}
\end{figure}

\paragraph{Dependence on the size of unlabeled data:} 
Figure \ref{fig: partition dependence balanced} shows $\alpha_U$ dependence of relative generalization error when $(\rho, \Delta)=(1/2, 0.75^2)$.  As already observed, PLS drastically reduces the generalization error when the total number of iterations is small. However, as $T$ grows, the effect of PLS becomes minor and the decrease of the generalization error gets slower, indicating that the nature of self-training gradually changes at large $T$. Also, as the size of the unlabeled data used at each iteration $\alpha_U$ decreases, the approach to the supervised loss becomes slower, and the performance is catastrophically poor when $\alpha_U$ is smaller than some threshold.  \inred{
    This instability at small $\alpha_U$ can also be seen from the linear susceptibility. See Appendix \ref{app: linear susceptibility} for more detail.
}

\begin{figure}[t!]
    \centering
    \includegraphics[width=0.95\linewidth]{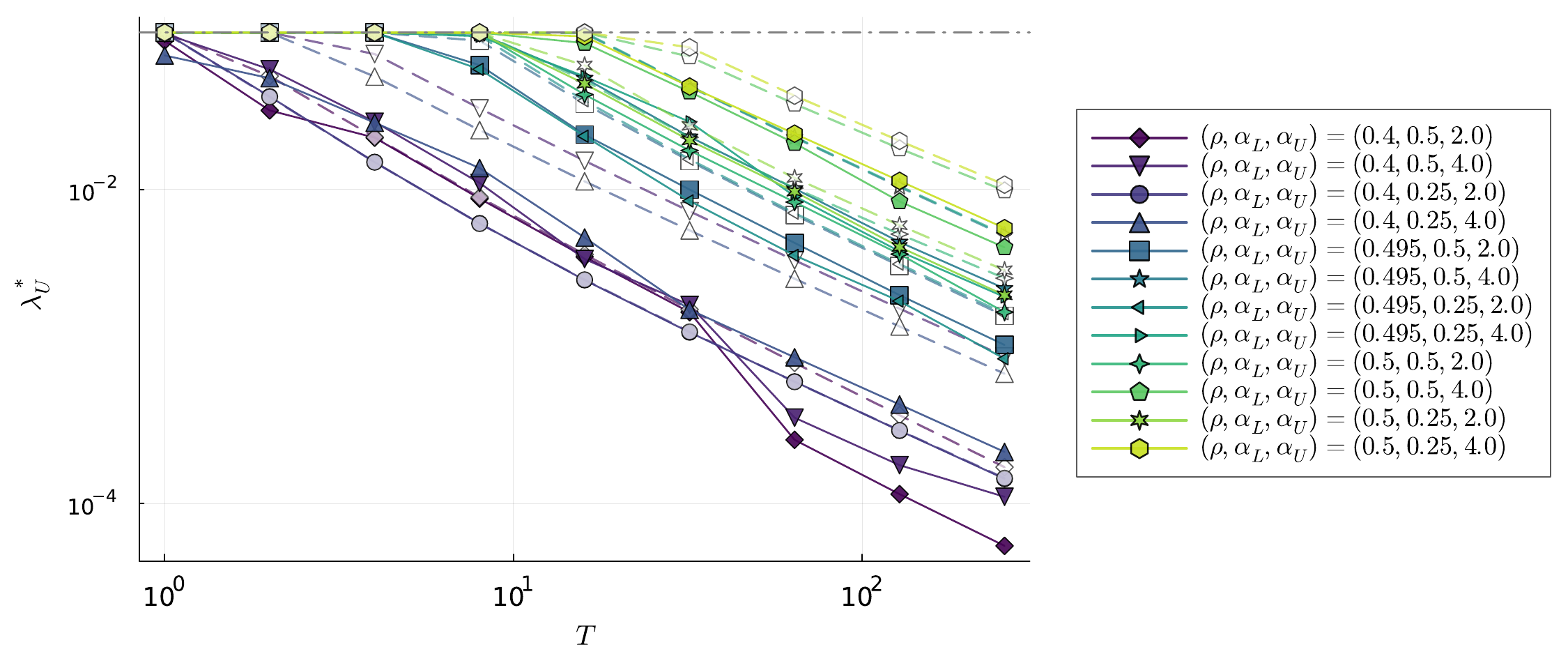}
    \caption{Optimal regularization parameter $\lambda_U^\ast$. Its dependence on the total number of iterations $T$ is shown.  Since $\lambda_U$ is upper bounded during the optimization, there is a plateau at $\lambda_U=0.1$.}
    \label{fig: optimal regularization}
\end{figure}

\paragraph{Optimal hyper parameters:} To obtain further insight on ST when $\alpha_U$ is moderately large so that long iteration of ST yields a better generalization, we show the optimal regularization parameter $\lambda_U^\ast$ in Figure \ref{fig: optimal regularization}. The figure indicates that the optimal regularization strength decreases in a power raw at $T\gg1$. Combined with the observations in Figure \ref{fig: partition dependence balanced}, it is suggested that ST can find a good classifier by accumulating small parameter updates when a large number of iterations can be used. Recall that, in our setting, ST returns the same parameters obtained in the previous step, i.e., $\hat{\bm{\theta}}^{(t)} = \hat{\bm{\theta}}^{(t-1)}$ when the training at each step of ST is in an underparametrized setup and the regularization is removed ($\lambda_U=0$). We also remark that this tendency does not depend on the label bias $\rho$.

\paragraph{\inred{Summary of numerical observations:}} 
\inred{The main numerical observations are summarized as follows.}
\begin{enumerate}
    \item \textbf{Small number of iterations.} When the total number of iterations $T$ is small, \inred{using pseudo-labels with high confidence, which also means that such pseudo-labels are close to hard labels, is effective. These observations imply that the optimal ST, whose hyperparameters are optimized, fits the model to relatively reliable pseudo-labels. In this sense, } the pseudo-label plays the intuitive role of pseudo-{\it labels}\inred{, as originally termed in \citep{lee2013pseudo}}.
    \item \textbf{Large number of iterations.} When $T$ is large and $\alpha_U$ is moderately large, the optimal regularization parameter \(\lambda_U^\ast\) becomes small. Thus, the difference in the model parameters between successive time steps is small due to the pseudo-label loss, and ST improves the classifier by accumulating small parameter updates. In this regime, the pseudo-label loss is better interpreted as maintaining continuity with the previous iterate, rather than simply fitting noisy labels.
    \item \textbf{Effect of label imbalance.} The large-$T$ behavior is qualitatively similar for different values of the label bias $\rho$, but the final performance relative to supervised learning depends strongly on the imbalance. ST is almost comparable to supervised learning when the label bias is small, whereas it remains suboptimal when the label imbalance is large.
\end{enumerate}

To deepen our understanding of the above observation, we will consider a perturbative analysis in the next subsection.

\subsection{Small regularization limit: perturbative analysis}
\label{subsec: small regularization limit}
According to the observations made in the previous subsection, to obtain a good classifier with long iterations of ST, the best strategy is to accumulate small parameter updates by using small regularizations and underparametrized setups at each iteration. To gain a deeper understanding of the behavior of ST when $T\gg1$ and $\lambda_U\ll1$, we consider a perturbative analysis. \inred{This is a formal perturbative  analysis of the self-consistent equations predicted in Section~\ref{subsec:rs free energy}. Accordingly, Predictions \ref{pred: smoothing}, \ref{pred: differential equation}, and \ref{pred: squared loss} are conditional on Predictions \ref{pred: rs generating functional w}--\ref{pred: u effective average} and on the assumptions stated below.  In particular, the existence and uniform validity of the expansion in $\lambda_U$ and the continuum limit are not established in a rigorous sense. Proposition \ref{prop: best direction} is a direct concequence of Prediction \ref{pred: differential equation}.
}

Specifically, the variables $\Theta^{(t)}$ and $\hat{\Theta}^{(t)}$ in Definition \ref{def: self-consistent equations} are expanded in terms of $\lambda_U$ like $q^{(t)} = q_0^{(t)} + q_1^{(t)}\lambda_U + \dots$. These expansion coefficients are then determined at each order of $\lambda_U$ in a self-consistent manner. Under appropriate assumptions, the variables at step $t$ are equal to those of the previous steps at $\lambda_U=0$: $(\Theta^{(t)}, \hat{\Theta}^{(t)})|_{\lambda_U=0} = (\Theta^{(t-1)}, \hat{\Theta}^{(t-1)})$. Hence, one can expect that the expansion coefficients at the first order determine the evolution of these parameters at $\lambda_U\to0$ like $(q^{(t)} - q^{(t-1)})/\lambda_U \to q_1^{(t)}$. In other words, by taking $\lambda_U$ as the time unit and $\tilde{t}=\lambda_U\times t$ as the continuous time, it can determine the continuous-time dynamics at $\lambda_U\to0$ as $\frac{d q_{\tilde{t}}}{d\tilde{t}} = q_{\tilde{t}}$, where $q_{\tilde{t}}$ is the value of $q^{(t)}$ in this continuous time scale. In this context, we also refer to the limit $\lambda_U\to0$ as the {\it continuum limit}.

We use the following two assumptions for the perturbative analysis. The first assumption is about the behavior of the loss function.
\begin{assumption}
\label{assumption: loss}
    Let $\tilde{l}_{\tilde{\Gamma}}(y, x) = \1(|y|>\tilde{\Gamma})l_{\rm pl}(\sigma_{\rm pl}(y), \sigma(x))$. Also, let $\hat{u}$ be the solution of the following equation with a positive constant $\tilde{c}>0$:
    \begin{equation}
        \hat{u}(y,x) = \argmin_{u} \frac{u^2}{2\tilde{c}} + \tilde{l}_{\tilde{\Gamma}}(y, x+u).
    \end{equation}
    Then the following holds for $\hat{u}(y,x)$ and $\tilde{l}$.
    \begin{enumerate}
    		\item $\hat{u}(y, x)$ is unique. 
		\item The derivative of the loss function is zero when $y=x$:
		\begin{equation}
        \left.
            \frac{\partial}{\partial x}\tilde{l}_{\tilde{\Gamma}}(y,x)
        \right|_{y=x} = 0,
        \label{eq: at extremum}
    \end{equation}
    \end{enumerate}

\end{assumption}
This is a natural condition for moderately regular loss function such as the squared loss $l_{\rm pl}(\sigma_{\rm pl}(y),\sigma(x))=(y-x)^2/2$ or the cross-entropy loss with the soft-sigmoid pseudo-label $l_{\rm pl}(\sigma_{\rm pl}(y), \sigma(x))=-\sigma(y)\log\sigma(x) - (1-\sigma(y))\log(1-\sigma(x))$.

The other assumption is the following:
\begin{assumption}
\label{assumption: stability}
    At $\lambda_U=0$, ST returns the same parameter with the previous step:
    \begin{equation}
        \hat{\bm{\theta}}^{(t)}|_{\lambda_U=0} = \hat{\bm{\theta}}^{(t-1)}.
        \label{eq: no update}
    \end{equation}
    Also the solution of the self-consistent equation can be series expanded in $\lambda_U$:
    \begin{align}
        \Theta^{(t)} &= \Theta_0^{(t)} + \Theta_1^{(t)}\lambda_U + \dots,
        \label{eq: expansion1}
        \\
        \hat{\Theta}^{(t)} &= \hat{\Theta}_0^{(t)} + \hat{\Theta}_1^{(t)}\lambda_U + \dots,
        \label{eq: expansion2}
    \end{align}
    which should be interpreted as component-wise expansion. For example, $q^{(t)}=q_0^{(t)}+q_1^{(t)}\lambda_U + \dots, m^{(t)}=m_0^{(t)} + m_1^{(t)}\lambda_U + \dots$, and so on.
\end{assumption}
The equation \eqref{eq: no update} is expected to hold naturally when the following conditions are used; (i) moderately large $\alpha_U$, i.e., underparametrized setup, (ii) monotonically increasing nonlinear functions $\sigma_{\rm pl}=\sigma$, and (iii) moderately regular loss functions, such as the squared loss and the cross-entropy loss. The latter equations \eqref{eq: expansion1}-\eqref{eq: expansion2} are the technical assumption. 

Under the Assumptions \ref{assumption: loss} and \ref{assumption: stability}, we first obtain the following result by analyzing the saddle point conditions except the condition \eqref{eq:rs-saddle-b} that determines the update of $\hat{B}^{(t)}$:

\begin{prediction}
\label{pred: smoothing}
	The lowest order of the expansion of $\hat{\chi}^{(t)}$ is at the second order: 
    \begin{equation}
        \hat{\chi}^{(t)} = \gO(\lambda_U^2).
    \end{equation}
    Also, the lowest order of the expansion of $q^{(t)} - (R^{(t)})^2/q^{(t-1)}$ is at the second order:
    \begin{equation}
		q^{(t)} - \frac{(R^{(t)})^2}{q^{(t-1)}} = \gO(\lambda_U^2).
    \end{equation}
\end{prediction}
See Appendix \ref{app: derivation of smoothing} for the derivation.
The consequence of this result is that the contributions from the noise terms $\sqrt{\hat{\chi}^{(t)}}\xi_w^{(t)}$ in \eqref{eq: effective w intuitive} and $\sqrt{q^{(t)} - \frac{(R^{(t)})^2}{q^{(t-1)}}}\xi_{u,2}^{(t)}$ in \eqref{eq: intuitive hu2} are higher order in $\lambda_U^{(t)}$ because these terms appear only in the form of a square (second order in $\lambda_U^{(t)}$) or the raw average (equals to zero). 
\inred{
    This implies that the contributions of the noise terms are negligible at the leading order in the weight update of ST in this small regularization setup. Hence, ST can extract information from the data in an almost noiseless way at least when updating the direction of the classification plane. This intuition is further supported by the following predictions.
}

For the square of the cosine similarity \eqref{eq: cosine similarity}, we obtain the following.
\begin{prediction}
    \label{pred: differential equation}
    Let $M^{(t)}$ be the square of the cosine similarity \eqref{eq: cosine similarity}: $M^{(t)} = (m^{(t)})^2/q^{(t)}$. Then, at the first order of expansion, it obeys the following equation:
    \begin{align}
        M^{(t)} &= M^{(t-1)} + C^{(t)}M^{(t-1)}(1-M^{(t-1)})\lambda_U + \gO(\lambda_U^2),
        \\
        C^{(t)} &= \frac{2}{\hat{Q}_0^{(t)}}\frac{\hat{m}_1^{(t)}}{m^{(t-1)}},
    \end{align}
    Also,
    \begin{equation}
        \hat{m}_0^{(t)} = 0,
    \end{equation}
    which shows that the zeros-order of the series expansion of $\hat{m}^{(t)}$ is zero. Moreover, $\hat{Q}_0^{(t)}>0$.
    
    Equivalently, it can be rephrased as follows. Let $(M_{\tilde{t}}, C_{\tilde{t}})$ be the value of $(M^{(t)}, C^{(t)})$ in the continuous time scale. Then, $M_{\tilde{t}}$ follows the following differential equation at the continuum limit:
    \begin{equation}
        \frac{d M_{\tilde{t}}}{d\tilde{t}} = C_{\tilde{t}} M_{\tilde{t}}(1-M_{\tilde{t}}), \quad C_{\tilde{t}} >0,
    \end{equation}
\end{prediction}
See Appendix \ref{app: derivation of differential equation} for the derivation of this prediction. Remarkably, the saddle point condition \eqref{eq:rs-saddle-b} that determines the update of $\hat{B}^{(t)}$ is not used to derive Prediction \ref{pred: differential equation}, which indicates that formally same expression for $M_1^{(t)}$ is obtained even if $B^{(t)}$ is kept constant. 

Since $\hat{m}_1^{(t)}$ is the leading order of $\hat{m}^{(t)}$ that represents the signal component of $\wsf{t}$ accumulated at step $t$ as depicted in \eqref{eq: effective w intuitive}, $C^{(t)}$ is positive if the training yields a positive accumulation of the signal component to $\wsf{t}$ when $m^{(t-1)}=\bm{v}\cdot \hat{\bm{w}}^{(t-1)}/N>0$, i.e., the classification plane at step $t-1$ is positively correlated with the cluster center, which seems to naturally hold when using a legitimate loss function. We remark that, in general, due to the presence of the noise term $\sqrt{\hat{\chi}^{(t)}}\xi_w^{(t)}$, an increase in cosine similarity is not guaranteed even if  $\hat{m}^{(t)}>0$. 

In the following, we assume that $\hat{m}^{(t)}$ is positive at the leading order in $\lambda_U$.
\begin{assumption}
\label{assumption: mhat is positive}
    The quantity $\hat{m}^{(t)}$ is positive at the leading order of $\lambda_U$, i.e., 
    \begin{equation}
        \hat{m}_1^{(t)} > 0.
        \label{eq: assumption of positivity of mhat}
    \end{equation}
\end{assumption}
Under this assumption, $C^{(t)}$ or $C_{\tilde{t}}$ is positive.

When the time-dependent variable $C_{\tilde{t}}$ is positive, the fixed point is only $M_{\tilde{t}}=0$ or $M_{\tilde{t}}=1$. Furthermore, as long as the initial classifier is informative, i.e., $M_0>0$, which is naturally expected by the supervised learning at the initial stage, the classification plane is oriented to the optimal direction at the long time limit. Hence, we can expect that the classification plane will find the best direction by accumulating small updates of ST. This is summarized as follows.
\begin{conditionalproposition}
    \label{prop: best direction}
    Under Assumption \ref{assumption: mhat is positive}, at the long time limit $\tilde{t}=\lambda_U\times t\to\infty$, the classification plane is oriented to the best direction if the initial classifier is informative $M^{(0)}>0$:
    \begin{equation}
        \inred{
            \lim_{\tilde{t}\to\infty} M_{\tilde{t}} = 1.
        }
    \end{equation}
\end{conditionalproposition}

The \inred{Prediction} \ref{pred: smoothing} and \inred{Proposition} \ref{prop: best direction} are the second main results of this paper. We remark that the above result is valid even if $\rho^{(t)}$ or $\Delta^{(t)}$ are time-dependent, although such time-dependence affects the value of $C_{\tilde{t}}$ and $C^{(t)}$. See the derivations in Appendix \ref{app: derivation of smoothing}-\ref{app: derivation of differential equation}.

\inred{
    Proposition \ref{prop: best direction} is a consequence of the noiseless accumulation of information shown in Predictions \ref{pred: smoothing} and \ref{pred: differential equation}. It corresponds to the message in the introduction that, when the number of iterations is large, ST improves the model by accumulating small parameter updates with a small regularization parameter. This supports the numerical observation that says weak regularization is better in large $T$ regime.
}

\inred{
    We note, however, that Proposition \ref{prop: best direction} concerns the direction of the classification plane only. As we discuss below, the generalization error is not necessarily optimal in the label-imbalanced case, even when the direction is aligned, because it also depends on the balance between the norm of the weight vector and the magnitude of the bias.
}

\begin{figure}[t!]
    \centering
    \begin{subfigure}[b]{0.322\textwidth}
        \centering
        \includegraphics[width=\textwidth]{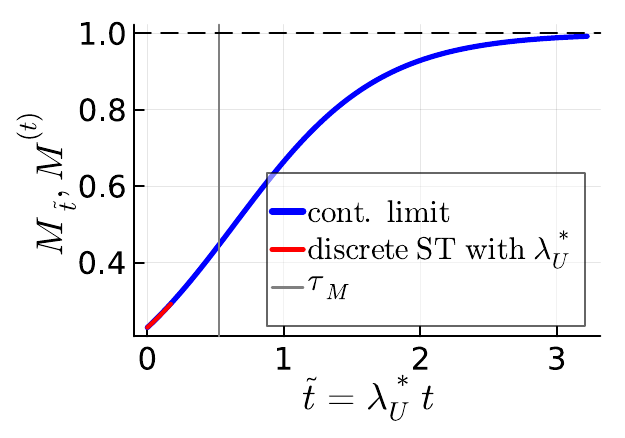}
        \caption{$(T, \rho) = (64, 0.2)$}
    \end{subfigure}
    %
    %
    \begin{subfigure}[b]{0.322\textwidth}
        \centering
        \includegraphics[width=\textwidth]{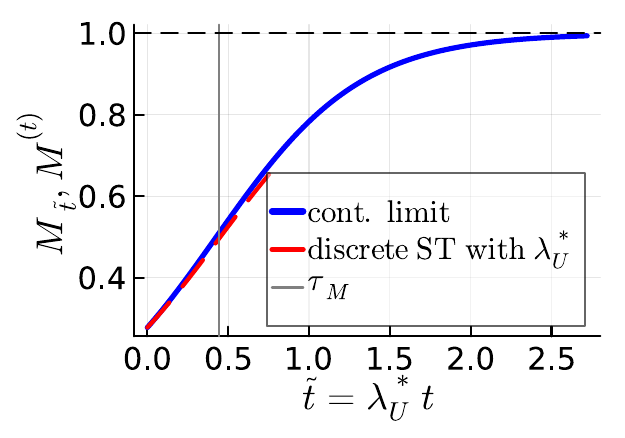}
        \caption{$(T, \rho) = (64, 0.4)$}
    \end{subfigure}
    %
    %
    \begin{subfigure}[b]{0.322\textwidth}
        \centering
        \includegraphics[width=\textwidth]{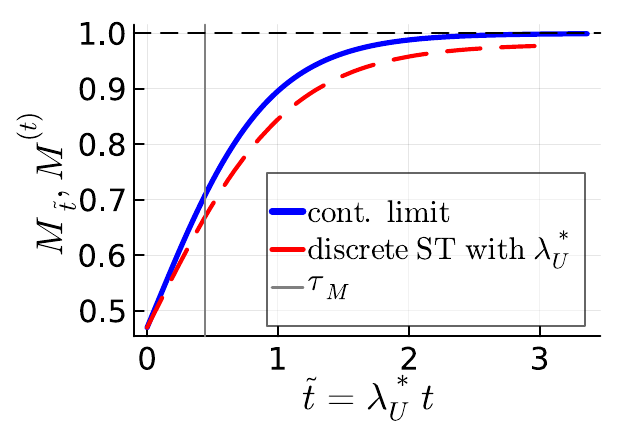}
        \caption{$(T, \rho) = (64, 0.5)$}
    \end{subfigure}
    \begin{subfigure}[b]{0.322\textwidth}
        \centering
        \includegraphics[width=\textwidth]{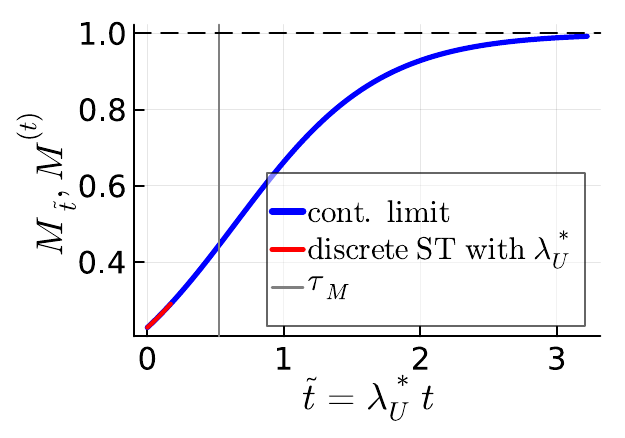}
        \caption{$(T, \rho) = (256, 0.2)$}
    \end{subfigure}
    %
    %
    \begin{subfigure}[b]{0.322\textwidth}
        \centering
        \includegraphics[width=\textwidth]{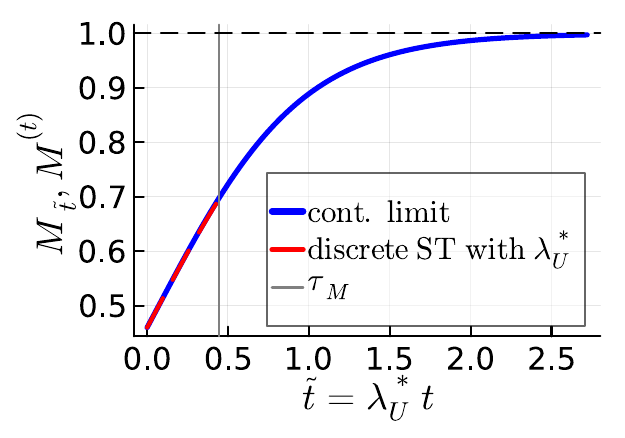}
        \caption{$(T, \rho) = (256, 0.4)$}
    \end{subfigure}
    %
    %
    \begin{subfigure}[b]{0.322\textwidth}
        \centering
        \includegraphics[width=\textwidth]{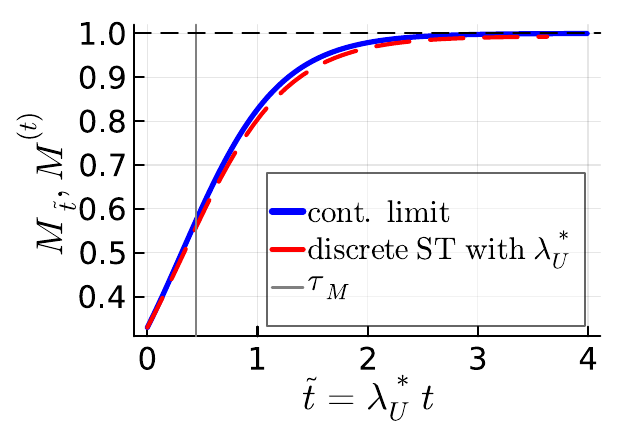}
        \caption{$(T, \rho) = (256, 0.5)$}
    \end{subfigure}
    \begin{subfigure}[b]{0.322\textwidth}
        \centering
        \includegraphics[width=\textwidth]{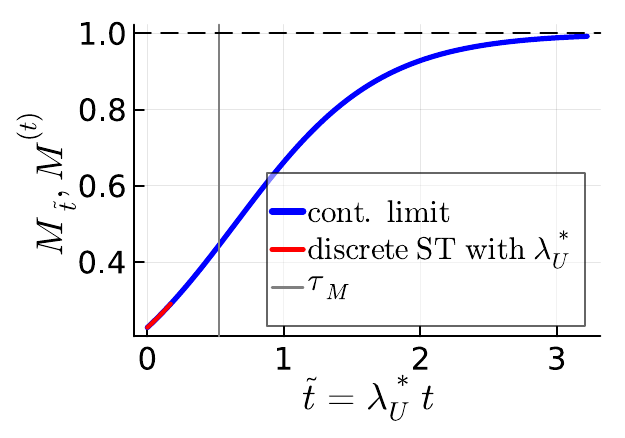}
        \caption{$(T, \rho) = (1024, 0.2)$}
    \end{subfigure}
    %
    %
    \begin{subfigure}[b]{0.322\textwidth}
        \centering
        \includegraphics[width=\textwidth]{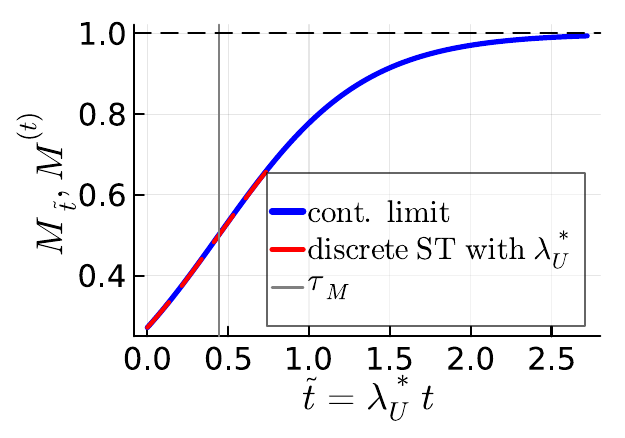}
        \caption{$(T, \rho) = (1024, 0.4)$}
    \end{subfigure}
    %
    %
    \begin{subfigure}[b]{0.322\textwidth}
        \centering
        \includegraphics[width=\textwidth]{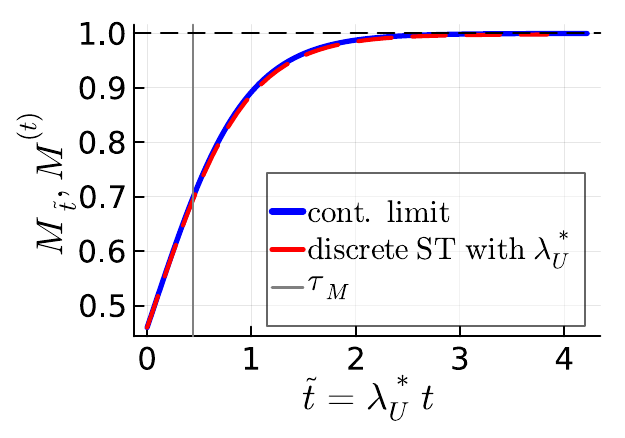}
        \caption{$(T, \rho) = (1024, 0.5)$}
    \end{subfigure}
    \caption{
        Comparison of the continuous-time dynamics \eqref{eq: continuous squared M} and the actual evolution of the discrete iteration of ST. $M^{(t)}$ and $M_{\tilde{t}}$ are the square of the cosine similarity \eqref{eq: cosine similarity} in discrete and continuous time scale, respectively.   The blue solid curve represents the continuous dynamics \eqref{eq: continuous squared M}. The red dashed line represents the result of the actual discrete update of ST. $\tau_M$ is the time scale required for the classification plane to sufficiently align in the same direction as $\bm{v}$. $(\alpha_L, \alpha_U, \Delta)=(0.5, 2.0, 0.75^2)$.
    }
    \label{fig: vs continuous squared loss}
\end{figure}

Notably, we can obtain a closed form solution when considering the squared loss without PLS: $\tilde{l}_{\tilde{\Gamma}}(y,x)=\frac{1}{2}(y-x)^2$, which can be a legitimate loss function in this Gaussian mixture setting \citep{mignacco2020role, oymak2020statistical, oymak2021theoretical}. In this case, we can obtain the following result:
\begin{prediction}
\label{pred: squared loss}
    Let $V_U=4\rho_U(1-\rho_U)$ be the variance of the true label. When $\tilde{l}_{\tilde{\Gamma}}(y,x) = \frac{1}{2}(y-x)^2$, the following hold at the continuum limit $\lambda_U\to0$ under Assumption \ref{assumption: stability}:
    \begin{align}
        M_{\tilde{t}} &= \frac{
            1
        }{
            1 + (1/M_0-1)e^{-\tilde{t}/\tau_M}
        }, 
        \label{eq: continuous squared M}
        \\
        m_{\tilde{t}} &= m_0 e^{-\frac{\tilde{t}}{\tau_m}}
        \\
        B_{\tilde{t}} &= B_0 + (2\rho_U-1) m_0(1-e^{-\frac{\tilde{t}}{\tau_m}}),
        \\
        \tau_M &= \frac{1}{2}\frac{\Delta}{V_U}(\alpha_U-1)(\Delta + V_U),
        \\
        \tau_m &= (\alpha_U - 1)(\Delta_U + V_U),
    \end{align}
    where $M_0, m_0, B_0$ are the initial conditions.
\end{prediction}
See Appendix \ref{app: derivation of differential equation squared loss} for the derivation of this prediction. As expected, the cosine similarity monotonically increases. In particular, the deviation of $M_{\tilde{t}}$ from unity is decreasing in an exponential manner. \inred{
    Moreover, this result is valid as long as $\alpha_U>1$. Otherwise the time scales $\tau_m, \tau_M$ are negative, causing the divergence of $M_t, m_t$. This indicates that the theory is not self-consistent. Recall that the value $\alpha_U=1$ is exactly the point that separates the under-/over-parameterized setup in this squared loss case. This perturbative analysis supports the numerical observation that a batch size $\alpha_U$ that is too small may not improve performance, implying that other special care is needed in overparameterized cases.
}

\inred{\subsubsection{Implications for the label-imbalanced cases}}
\inred{From Predictions \ref{pred: smoothing}, \ref{pred: differential equation}, \ref{pred: squared loss} and Proposition \ref{prop: best direction}}, we can expect that the classification plane is pointing in the best direction as the number of iterations grows with moderately small regularization parameter $\lambda_U$. However, as we saw in Figure \ref{fig: relative generalization error} and \ref{fig: cosine similarity}, it is not the case in label-biased cases even after hundreds of iterations, with practical values of $\lambda_U^\ast$.

We argue that this is due to a potential imbalance between the magnitude of the bias $\hat{B}^{(t)}$ and the norm of the weight vector $\sqrt{q^{(t)}} = \sqrt{\|\hat{\bm{w}}^{(t)}\|_2^2/N}$. As the generalization error \eqref{eq: rs generalization error} depends on the ratio between these quantities $\hat{B}^{(t)}/\sqrt{q^{(t)}}$, the generalization error may be at a random level if the ratio is significantly large. Indeed, this ratio can become large when the labels are imbalanced for the following reasons. Since the pseudo-label just ensures the continuity between the successive ST steps in our setup under Assumption \ref{assumption: stability}, the norm of the weight vector gradually decreases as the iteration grows due to the shrinkage effect of the regularizer $\lambda_U\|\bm{w}^{(t)}\|_2^2$. On the other hand, the magnitude of the bias term does not necessarily decrease because it is not directly regularized.

This point is evident when $\tilde{l}_{\Gamma}(y,x)=(y-x)^2/2$. From the closed-form solution in Prediction \ref{pred: squared loss}, we see that $B_{\tilde{t}}\to B_{\infty}\neq 0$ while $M_{\tilde{t}}\to1$ and $m_{\tilde{t}}\to0$, at $\tilde{t}\to\infty$. This indicates that the norm of the weight vanishses. Hence $|B_{\tilde{t}}/\sqrt{q_{\tilde{t}}}|\to\infty$. To avoid this pathological behavior, the best regularization parameter $\lambda_U^\ast$ tends to be very small so that $\tilde{t}$ does not diverge. In figure \ref{fig: vs continuous squared loss}, we compare the continuous-time dynamics \eqref{eq: continuous squared M} and the actual evolution of the discrete iteration of ST. As $T$ increases, $\lambda_U^\ast$ decreases to justify the description by continuous-time dynamics \eqref{eq: continuous squared M}. However, the elapsed time $\lambda_U^\ast \times t$ of the actual optimal regularization parameter $\lambda_U^\ast$ is not sufficiently larger than $\tau_M$ when $\rho\neq 1/2$, although $M_{\tilde{t}}\simeq 1$ at $\tilde{t}\gg1$ as expected. Note that the above continuous time argument is valid as long as the regularization parameter $\lambda_U$ is small so that the second-order term of the perturbative expansion can be ignored, but it says nothing about the regularization parameter $\lambda_U^\ast$ that is actually obtained by minimizing the generalization error in the last step of ST. 

Therefore, we may need some heuristics to compensate for the shrinkage of the norm of the weight vector due to the regularization, at least in our online setup. 

\section{Heuristics for label imbalanced cases}
\label{sec: heuristics}

\begin{figure}[t!]
    \centering
    \begin{subfigure}[b]{\textwidth}
        \centering
        \includegraphics[width=\linewidth]{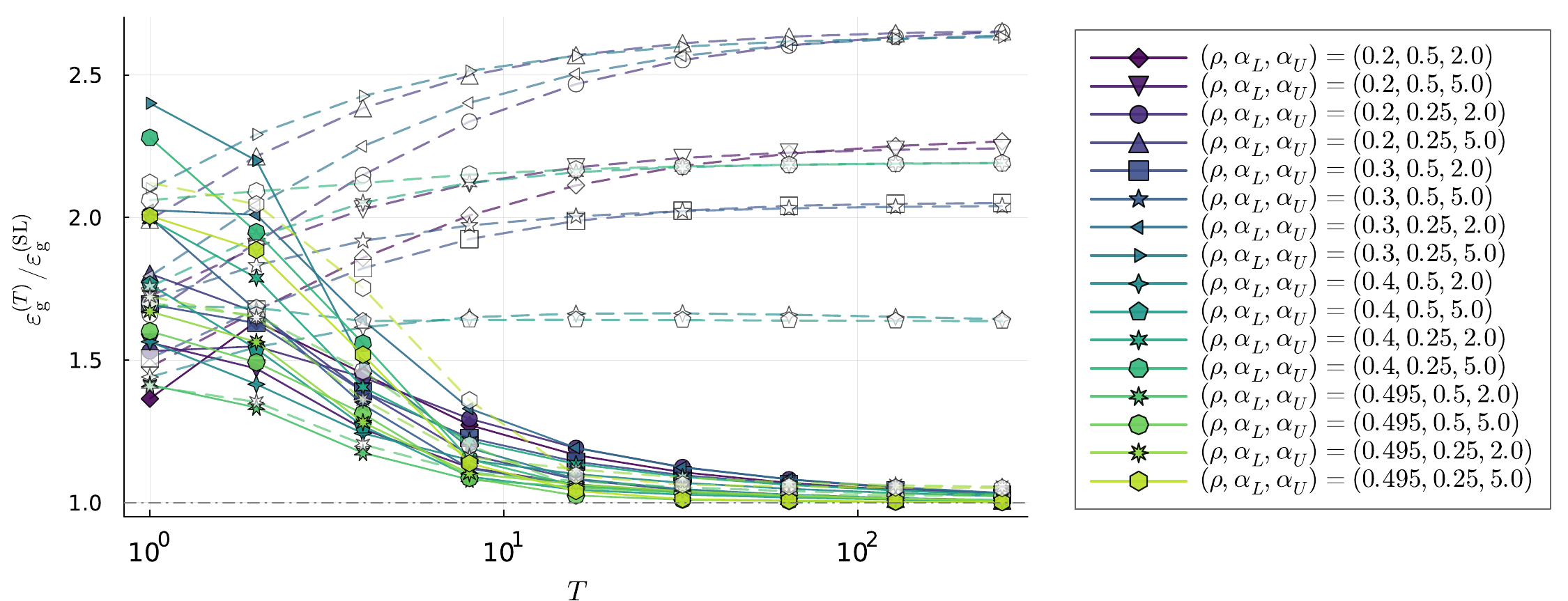}
        \caption{$\epsilon_{\rm g}^{(T)}/\epsilon_{\rm g}^{\rm (SL)}$ (relative generalization error)}
    \end{subfigure}
    \begin{subfigure}[b]{\textwidth}
        \centering
        \includegraphics[width=\linewidth]{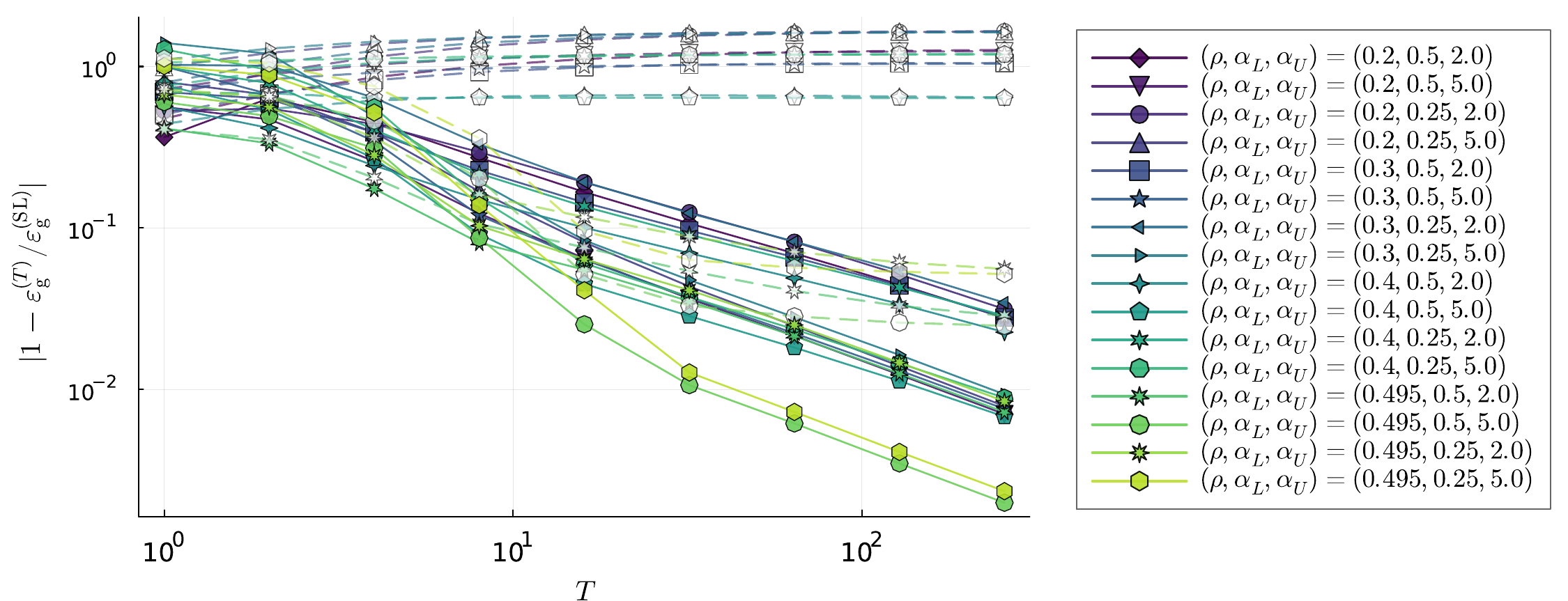}
        \caption{$|1 - \epsilon_{\rm g}^{(T)}/\epsilon_{\rm g}^{\rm (SL)}|$ (absolute value of deviation from unity)}
        \label{fig: absolute deviation}
    \end{subfigure}
    
    \caption{
        The ratio of the generalization error obtained at the end of ST ($t=T$) to the SL with a labeled dataset of size $N(\alpha_L + \alpha_U \times T)$. The cases with the Heuristics \ref{heuristics: annealing} and \ref{heuristics: bias fixing} are included. The white markers with dash lines represent the ``naive'' ST with $l(y,x)=-y\log x -(1-y)\log(1-x), \sigma_{\rm pl}(x)=\sigma(x) = 1/(1+e^{-x})$. The filled markers with solid lines represent the result with the heuristics. Here, $\Gamma$ is set to be zero (without PLS). The upper panel shows the raw values. The lower panel shows their absolute values of the deviation from unity in the log scale.
    }
    \label{fig: pseudo-label annealing}
\end{figure}

In the previous section, we saw that ST can find a classification plane with the optimal direction with a large number of iterations and a small regularization. However, the imbalance of the norm of the weight and the magnitude of the bias can be problematic in a label imbalanced case. To overcome the problems in label imbalanced cases, we introduce the following two heuristics.
\begin{heuristics}[Pseudo-label annealing]
    \label{heuristics: annealing}
    In the ST steps, we use the model's nonlinear function as the pseudo-labeler, but gradually amplify its input:
    \begin{align}
        \sigma_{\rm pl}(x) &= \sigma(\gamma^{(t)} x),
        \\
        \gamma^{(t)} &= 1 + a t \quad a>0.
        \label{eq: annealing parameter}
    \end{align}
\end{heuristics}
We expect that the classification plane will gradually align in the optimal direction at small $t$ because $\gamma^{(t)} \simeq 1$ at such time, which is close to the condition assumed in the Prediction \ref{pred: differential equation}. Later, by gradually making the pseudo-labels closer to hard labels, the norm shrinkage of the weight vector will be compensated. However, even with this heuristic, it may still be difficult to learn the bias and the weight at the same time. Therefore, we propose to fix the bias term to that of the initial classifier:
\begin{heuristics}[Bias-fixing]
    \label{heuristics: bias fixing}
    Fix the bias term at that obtained at the supervised learning at $t=0$:
    \begin{equation}
        \hat{B}^{(t)} = \hat{B}^{(0)}, \quad t=1,2,\dots.
    \end{equation}
\end{heuristics}
We expect the bias term of the initial classifier to be moderately good unless the labeled data is severely limited.

\subsection{Demonstration}
\label{subsec: heuristic demonstrations}
To check the effectiveness of the above heuristics, we numerically solve the self-consistent equations and investigate the generalization error. To include Heuristics \ref{heuristics: bias fixing}, we remove the equation \eqref{eq:rs-saddle-b} from the self-consistent equation, which corresponds to the saddle point condition with respect to the bias. Here, the loss function is the logistic loss $l(y,x)=-y\log x -(1-y)\log(1-x)$ and the nonlinear function of the model is the sigmoid function $\sigma(x)=1/(1+e^{-x})$. Also, $\Gamma$ is set to be zero (without PLS) since PLS did not make significant difference when $T\gg1$ and $\rho_U=1/2$ (see Figure \ref{fig: relative generalization error}). The annealing parameter $a$ is optimized to minimize the generalization error in the last step of ST.

Figure \ref{fig: pseudo-label annealing} shows the relative generalization error as in Figure \ref{fig: relative generalization error}.  It is clearly shown that the performance is significantly improved by using the heuristics. Indeed, it is shown that the generalization error obtained by ST with the heuristics is almost compatible with the supervised learning with true labels, even in the highly label imbalanced case $\rho=0.2$, demonstrating the effectiveness of the proposed heuristics. Without the heuristics, the $|1 - \epsilon_{\rm g}^{(T)}/\epsilon_{\rm g}^{\rm (ST)}|$ converges to a non-zero value even when almost label balanced case $\rho=0.495$ as shown in Figure \ref{fig: absolute deviation}.

\begin{figure}[t!]
    \centering
    \includegraphics[width=\linewidth]{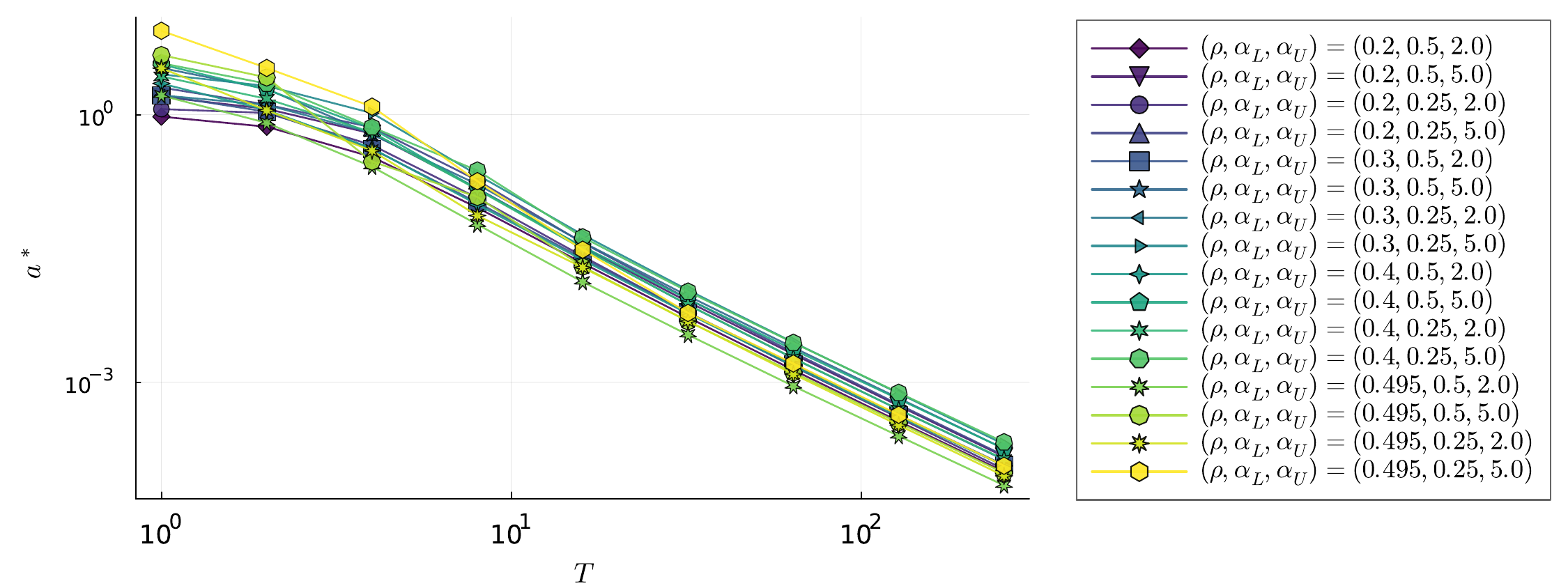}
    \caption{
        Optimal values of the annealing parameter $a$ in \eqref{eq: annealing parameter}. The settings are the same as in Figure \ref{fig: pseudo-label annealing}.
    }
    \label{fig: optimal values of a}
\end{figure}

Figure \ref{fig: optimal values of a} shows the optimal values of the annealing parameter $a^\ast$ in \eqref{eq: annealing parameter}. As the total number of iterations $T$ increases, this parameter gradually decreases, indicating that the pseudo-label is slowly approaching hard labels when the total number of iterations $T$ is large. In such long iterations, the cost function \eqref{eq: update of ST} involving the pseudo-labels can be interpreted as playing the role of maintaining continuity with the previous step, rather than the intuitive role of fitting to a pseudo-{\it label} as already observed in subsection \ref{subsec: numerical inspection}. On the other hand, when the total number of iterations $T$ is small, its value is rather large, indicating ST fits the model to the almost hard labels. In such cases, the cost function \eqref{eq: update of ST} can be interpreted as playing the intuitive role of fitting the pseudo-{\it label}. These observations suggest that there is a crossover regarding the role of the pseudo-label as the total number of iterations changes.

\inred{
    Finally, we remark that Heuristic \ref{heuristics: annealing} and Heuristic \ref{heuristics: bias fixing} are motivated by the understanding of ST obtained in Section \ref{subsec: small regularization limit}, which indicates the usefulness of our theoretical analysis.
}

\section{Summary and conclusion}
\label{section:summary and conclusion}
In this work, we developed a theoretical framework for analyzing the iterative ST, when a new batch of unlabeled data is fed into the algorithm at each training step. The main idea was to analyze the generating functional \eqref{eq: generating functional} by using the replica method. We applied the developed theoretical framework to the iterative ST that uses the ridge-regularized convex loss at each training step. In particular, we considered the data generation model in which each data point is generated from a binary Gaussian mixture with a general variance of each cluster $\Delta$, the label bias $\rho$, and the magnitude of the PLS $\Gamma$. Prediction \ref{pred: rs generating functional w}-\ref{pred: u effective average} itself, which captures the sharp asymptotics of ST by the effective single body problem, is valid even when the cluster size and the label bias are step-dependent.

Based on the derived formula, we studied the behavior of ST quantitatively. First, we explored the behavior of ST by numerically solving the self-consistent equations defined in Definition \ref{def: self-consistent equations}. Our findings include that (i) ST can find a model whose performance is nearly compatible with supervised learning with true labels when the label bias is small, (ii) when large number of iterations $T$ can be used, the best achievable model results from accumulating small parameter updates over long iterations by using a small regularization parameter and moderately large batches of unlabeled data, and (iii) when available number of iterations $T$ is small, PLS drastically reduce the generalization error. Second, to gain a deeper understanding of the behavior of ST when $1\ll T$ and $\lambda_U\ll1$, we performed a perturbative analysis by expanding the solution of the self-consistent equation in terms of $\lambda_U$. As a result, under some assumptions on the loss functions, it was shown that ST can potentially find a classification plane with the optimal direction regardless of the label bias by considering a perturbative analysis in $\lambda_U$. Intuitively speaking, this is due to the fact that the noise terms in effective description of the weight \eqref{eq: effective w intuitive} and logits \eqref{eq: intuitive hu2} becomes higher order in $\lambda_U$, hence noiseless accumulation of the information is possible.  However, the imbalance between the norm of the weight and the magnitude of the bias might be problematic in label imbalanced cases. Notably, when using the squared loss without PLS, we could derive a closed-form solution for the evolution of cosine similarity, which clearly confirms the above result. Finally, to overcome the issues in label imbalanced cases, we proposed two heuristics: namely the pseudo-label annealing (Heuristics \ref{heuristics: annealing}) and the bias-fixing (Heuristics \ref{heuristics: bias fixing}). These heuristics drastically improved the performance of ST in label imbalanced cases. It was demonstrated that ST's performance is nearly compatible with supervised learning using true labels, even in cases of severe label imbalance. In label balanced case $\rho=1/2$, we do not need to include the bias term, i.e., $\hat{B}^{(t)}=0$ is optimal and the imbalance between the bias and the weight's norm is irrelevant. Thus, our findings are compatible with the previous studies \citep{oymak2020statistical, oymak2021theoretical, frei2022} in the label balanced case. However, our results suggest that ST is effective even in label imbalanced cases.

\inred{
    Overall, the numerical results suggest a crossover in the role of pseudo-labels depending on the number of ST iterations. When $T$ is small, PLS and nearly hard pseudo-labels are effective, indicating that ST mainly benefits from fitting relatively reliable high-confidence pseudo-labels. This picture is consistent with the name of pseudo-labels. When $T$ is large and $\alpha_U$ is moderately large, the optimal regularization parameter $\lambda_U^\ast$ becomes small, so consecutive models remain close to each other and the pseudo-label loss acts mainly as a continuity constraint between successive iterates. In this regime, ST improves the classifier by accumulating many small updates, rather than by directly fitting noisy pseudo-labels.
}

Promising feature directions include extending the present analysis to more advanced models such as the random feature model \citep{gerace2020generalisation}, kernel method \citep{canatar2021spectral, dietrich1999svm}, and multi-layer neural networks \citep{schwarze1992generalization, yoshino2020complex}. Also, quantitatively comparing with other semi-supervised learning methods such as the entropy regularization \citep{Grandvalet2006} is a natural future research direction. Although the literature \citep{chen2020self} shows that ST is the same with the entropy regularization if the pseudo-labels are updated after each SGD step, the connection is vague when the pseudo-label is fixed until the student model completely minimizes the loss as in our setup.  Another natural but challenging direction is to extend our analysis to the case in which the same \inred{labeled/}unlabeled data are used in each iteration, which requires careful handling of the correlations of the estimators at different iteration steps as was done in the analysis of alternating minimization \citep{okajima2024asymptotic}. \inred{In such case, as reported in the analysis of knowledge distillation, one may need another heuristics such as early stopping \citep{NEURIPS2025_18ddfb19}. The mixture of hard/soft-label is also an issue, requiring thorough analysis.}

\acks{
    This study was supported by JSPS KAKENHI Grant No. 21K21310 and 23K16960, JST ACT-X Grant Number JPMJAX24CG, and Grant-in-Aid for Transformative Research Areas (A), ``Foundation of Machine Learning Physics'' (22H05117).
}

\appendix
\section{Replica method calculation}
\label{app: replica method calculation}
In this appendix, we outline the replica calculation leading to the predictions \ref{pred: rs generating functional w}-\ref{pred: u effective average}.  Also, let $\lambda_U^{(t)}, \alpha_U^{(t)}, \rho_U^{(t)}$ and $\Delta_U^{(t)}$ be the values of the regularization parameter, the relative size of the unlabeled data, and the label imbalance and the size of the cluster at step $t\ge1$. Although we mainly focus on the case of $\lambda_U^{(t)}=\lambda_U, \alpha_U^{(t)}=\alpha_U, \rho_U^{(t)}=\rho_U$ and $\Delta_U^{(t)}=\Delta_U$ in the main text, the following derivation is valid even in this step-dependent cases. 

\subsection{Replica method for generating functional of ST}
\label{subapp: replica method}
We begin with recalling the basic strategy for evaluating the generating functional \eqref{eq: generating functional} using the replica method. The key observation is that the evaluation of the generating functional \eqref{eq: generating functional} requires averaging the inverse of the normalization constants $(Z^{(0)}(D_L))^{-1}$ and $\prod_{t=1}^{T-1}(Z^{(t)}(D_U^{(t)},\bm{\theta}^{(t-1)}))^{-1}$ in the Boltzmann distributions:
\begin{align}
    &\Xi_{\rm ST}(\epsilon_w, \epsilon_B) = \lim_{N,\beta\to\infty}\E_D\left[\int 
        e^{
            \epsilon_w g_w(\{w_i^{(t)}\}_{t=0}^{T}) 
            + \epsilon_Bg_B(\{B^{(t)}\}_{t=0}^{T})} 
        \nonumber
        \right.
        \\
        &\left.
        \times 
        \frac{
            1
        }{
            Z^{(0)}(D_L)
        }e^{
            -\beta^{(0)}\gL^{(0)}(\bm{\theta}^{(0)};D_L)
        }
        \prod_{t=1}^T\frac{
            1
        }{
            Z^{(t)}(D_U^{(t)}, \bm{\theta}^{(t-1)})
        }e^{
            -\beta^{(t)}\gL^{(t)}(\bm{\theta}^{(t)};D_U^{(t)}, \bm{\theta}^{(t-1)})
        }
    d\bm{\theta}^{(0)}\dots d\bm{\theta}^{(T)}\right].
\end{align}
Since the dependence of the normalization constants on $D$ is non-trivial, taking average over $D$ is technically difficult. 

To resolve this difficulty, we use the replica method as follows. This method rewrites $\Xi_{\rm ST}$ using the identity $(Z^{(t)})^{-1} = \lim_{n_t\to0}(Z^{(t)})^{n_t-1}$ as 
\begin{align}
    \Xi_{\rm ST} &= \lim_{n_0,\dots,n_T\to0} \phi^{(T)}_{n_0,\dots,n_T},
    \label{appeq: replica method}
    \\
    \phi^{(T)}_{n_0,\dots,n_T} &= \lim_{N,\beta\to\infty}\E_D\left[\int 
        e^{\epsilon_w g_w(\{w_i^{(t)}\}_{t=0}^{T}) 
        + \epsilon_Bg_B(\{B^{(t)}\}_{t=0}^{T})} 
        \nonumber 
        \right.
        \\
        &\left.
        \times \left(Z^{(0)}(D_L)\right)^{n_0-1}
        e^{
            -\beta^{(0)}\gL^{(0)}(\bm{\theta}^{(0)};D_L)
        }
        \right.
        \nonumber
        \\
        &\left.
        \times \prod_{t=1}^T
        \left(Z^{(t)}(D_U^{(t)}, \bm{\theta}^{(t-1)})\right)^{n_t-1}
        e^{
            -\beta^{(t)}\gL^{(t)}(\bm{\theta}^{(t)};D_U^{(t)}, \bm{\theta}^{(t-1)})
        }
    d\bm{\theta}^{(0)}\dots d\bm{\theta}^{(T)}\right].
\end{align}
Although the equation \eqref{appeq: replica method} itself is just an identity, the evaluation of ${\phi}_{n_0,\dots,n_T}^{(T)}$ for $n_t\in\R$ is difficult. Still, for positive integers $n_0, \dots,n_T=1,2,\dots$, it has an appealing expression:
\begin{align}
    \phi^{(T)}_{n_0,\dots,n_T} &= \lim_{N,\beta\to\infty}\E_D\left[\int 
        e^{\epsilon_w g_w(\{w_{1,i}^{(t)}\}_{t=0}^{T}) 
        + \epsilon_Bg_B(\{B_1^{(t)}\}_{t=0}^{T})} 
        \prod_{a_0=1}^{n_0}
        e^{
            -\beta^{(0)}\gL^{(0)}(\bm{\theta}_{a_t}^{(0)}; D_L)
        }
        \right.
        \nonumber
        \\
        &\left.
        \times \prod_{t=1}^T
        \prod_{a_t=1}^{n_t}
        e^{
            -\beta^{(t)}\gL^{(t)}(\bm{\theta}_{a_t}^{(t)};D_U^{(t)}, \bm{\theta}_1^{(t-1)})
        }
    d^{n_0}\bm{\theta}^{(0)}\dots d^{n_T}\bm{\theta}^{(T)}\right],
\end{align}
where $d^{n_{t}}\bm{\theta}^{(t)}, t=0,1,\dots, T$ are the shorthand notations for $d\bm{\theta}_1^{(t)}\dots d\bm{\theta}_{n_t}^{(t)}$, and $\{a_t\}_{t=0}^T$ are indices to distinguish the additional variables introduced by the replica method. Hereafter we will refer to this type of index as the {\it replica index}. Note that the index $1$ that appears in $\epsilon_w g_w(\{w_{1, i}^{(t)}\}_{t=0}^{T})  + \epsilon_Bg_B(\{B_1^{(t)}\}_{t=0}^{T})$ is also the replica index.  The augmented system \eqref{appeq: replicated system}, which we refer to as the {\it replicated system}, is much easier to handle than the original generating functional because all of the factors to be evaluated are now explicit. Indeed, we can consider the average over $D$ {\it before} completing the integral with respect to $\{\bm{\theta}_{\a t}^{(t)}\}$:
\begin{align}
    \phi^{(T)}_{n_0,\dots,n_T} &= \lim_{N,\beta\to\infty}\int
        e^{\epsilon_w g_w(\{w_{1,i}^{(t)}\}_{t=0}^{T}) 
        + \epsilon_Bg_B(\{B_1^{(t)}\}_{t=0}^{T})} 
        \E_D\left[
            \prod_{a_0=1}^{n_0}
            e^{
                -\beta^{(0)}\gL^{(0)}(\bm{\theta}_{a_t}^{(0)}; D_L)
            }
            \right.
            \nonumber
            \\
            &\left.
            \times \prod_{t=1}^T
            \prod_{a_t=1}^{n_t}
            e^{
                -\beta^{(t)}\gL^{(t)}(\bm{\theta}_{a_t}^{(t)};D_U^{(t)}, \bm{\theta}_1^{(t-1)})
            }
        \right]
    d^{n_0}\bm{\theta}^{(0)}\dots d^{n_T}\bm{\theta}^{(T)}.
    \label{appeq: replicated system}
\end{align}

In the following, utilizing this formula, we obtain an analytical expression for $n_t\in \sN$. Subsequently, under appropriate symmetry assumption of the resultant expression, we extrapolate it to the limit as $n_t\to0$. This analytical continuation $n_t\to0$ from $n_t\in\sN$ is the procedure that has not yet been developed into a mathematically rigorous formulation. This aspect is what renders the replica method a non-rigorous technique. 	In principle, the evaluation of the replicated system for $n_t\in\sN$ is a mathematically well-defined problem, and this parts could be made rigorous by an appropriate probabilistic analysis.  See \citep{montanari2024} for an example in the analysis of the spiked tensor model.  However, the expression for the replicated system \eqref{appeq: replicated system} with integer $\{n_t\}_{t=0}^T$ alone may not uniquely determine the expression for the replicated system at real $\{n_t\}_{t=0}^T$. Hence the analytical continuation of $\phi_{n_0,\dots, n_{T}}^{(T)}$ from $n_t\in\sN$ to $n_t\in\R$ is a mathematically ill-defined problem. Currently, what we can do is just {\it guess} the appropriate form from $n_t\in\sN$ and extrapolate is to $n_t\in\R$.  Nevertheless, for linear models, a widely applicable extrapolation procedure is known, which has succeeded in obtaining exact results for a variety of non-trivial problems as we already commented in subsection \ref{subsec: related works}. In the following, we outline the treatment of the replicated system \eqref{appeq: replicated system}.

\subsection{Handling of the replicated system}
\label{subapp: handling of the replicated system}
We next consider how to evaluate the replicated system \eqref{appeq: replicated system} using the RS assumption and saddle point method.

Inserting the explicit form of the loss functions \eqref{eq: loss labeled}-\eqref{eq: update of ST} and the feature vectors \eqref{eq:data genereation labeled}-\eqref{eq:data genereation unlabeled} and using the independence of each data point, we obtain the following:
\begin{align}
    &\phi^{(T)}_{n_0,\dots,n_T} = \lim_{N,\beta\to\infty}\int
        e^{\epsilon_w g_w(\{w_{1,i}^{(t)}\}_{t=0}^{T}) 
        + \epsilon_Bg_B(\{B_1^{(t)}\}_{t=0}^{T})} 
        \nonumber\\
        &\times\prod_{\mu=1}^{M_L}\E_{\bm{x}_\mu^{(0)}, y_\mu^{(0)}}\left[
            \prod_{\a 0=1}^{n_0}e^{
            -\beta^{(0)} l\left(
                y_\mu^{(0)}, 
                \sigma(
                    f_{\a 0, \mu}^{(0)}
                )
            \right)
        }
        \right]
        \nonumber \\
        &
        \times\prod_{t=1}^T \prod_{\nu=1}^{M_U}\E_{\bm{x}_\nu^{(t)}, y_\nu^{(t)}}\left[
            \prod_{\a t=1}^{n_t}e^{
                -\beta^{(t)}\1(
                \left|
                    \tilde{f}_{\a t, \nu}^{(t)}
                \right| > \Gamma \sqrt{\|\bm{w}_{\a t}^{(t)}\|_2^2/N}
            )l_{\rm pl}\left(
                    \sigma_{\rm pl}(
                        \tilde{f}_{\nu}^{(t)},
                    ), \sigma(
                        f_{\a t, \nu}^{(t)}
                    )
                \right)
            }
        \right]
        \nonumber \\
        &\times
        \prod_{t=0}^T \prod_{j=1}^N \prod_{\a t=1}^{n_t}e^{-\frac{\beta^{(t)}\lambda^{(t)}}{2}(w_{\a t,i}^{(t)})^2}
    d^{n_0}\bm{\theta}^{(0)}\dots d^{n_T}\bm{\theta}^{(T)},
    \label{appeq: replicated system inserted}
\end{align}
with the notations
\begin{align}
    f_{\a0,\mu}^{(0)} &= B_{\a 0}^{(0)} + (2y_\mu^{(0)} -1)\scalen \rw{\a0}{0}\cdot \bm{v} + \scalensqrt  \rw{\a0}{0}\cdot \bm{z}_\mu^{(0)},
    \\
    \tilde{f}_{\nu}^{(t)} &= B_1^{(t-1)} + (2y_\nu^{(t)}-1)\scalen \rw{1}{t-1}\cdot\bm{v} +  \scalensqrt \rw{1}{t-1}\cdot\bm{z}_\nu^{(t)},
    \\
    f_{\a t,\nu}^{(t)} &= B_{\a t}^{(t)} + (2y_\nu^{(t)}-1)\scalen \rw{1}{t}\cdot\bm{v} + \scalensqrt \rw{1}{t}\cdot\bm{z}_\nu^{(t)}.
\end{align}

\subsubsection{Introduction of order parameters and the saddle point method}
The important observation is that the Gaussian noise terms in \eqref{appeq: replicated system inserted} appear only through the following vectors:
\begin{align}
    \bm{u}_\mu^{(0)} &= \left(
        \scalensqrt \rw{1}{0}\cdot \bm{z}_\mu^{(0)}, \scalensqrt \rw{2}{0} \cdot \bm{z}_\mu^{(0)}, \dots, \scalensqrt \rw{n_0}{0}\cdot \bm{z}_\mu^{(0)}
    \right)^\top
    \in\R^{n_0}
    ,
    \\
    \bm{u}_\nu^{(t)} &= \left(
        \scalensqrt \rw{1}{t-1}\cdot \bm{z}_\nu^{(t)}, \scalensqrt \rw{1}{t}\cdot \bm{z}_\nu^{(t)}, \scalensqrt \rw{2}{t}\cdot \bm{z}_\nu^{(t)}, \dots,  \scalensqrt \rw{n_t}{t}\cdot \bm{z}_\nu^{(t)}
    \right)^\top
    \in\R^{n_t+1}
    ,
    \nonumber \\
    t &= 1,2,\dots,T, 
\end{align}
which follows independently the multivariate Gaussians as  
\begin{align}
    \bm{u}_\mu^{(0)}&\sim_{\rm iid} \mathcal{N}(0, \Sigma^{(0)}), \quad
    \Sigma^{(0)} = \Delta_L\times\left[\begin{array}{ccc}
        Q_{11}^{(0)} & \cdots & Q_{1n_0}^{(0)}  \\
        \vdots &  \ddots & \vdots \\
        Q_{n_01}^{(0)} & \cdots & Q_{n_0n_0}^{(0)}
    \end{array}
    \right],
    \\
    Q_{\a{0}\b0}^{(0)}&\equiv \scalen \bm{w}_{\a{0}}^{(0)}\cdot \bm{w}_{\b{0}}^{(0)}, \quad \a{0},\b{0} = 1,2,\dots, n_0,
\end{align}
and
\begin{align}
    \bm{u}_\nu^{(t)}&\sim_{\rm iid} \mathcal{N}(0, \Sigma^{(t)}),
    \quad 
    \Sigma^{(t)} = \Delta_U^{(t)}\times\left[
        \begin{array}{c|ccc}
             Q_{11}^{(t-1)} & R_{1}^{(t)} & \cdots & R_{n_t}^{(t)} \\
             \hline
             R_{1}^{(t)} & Q_{11}^{(t)} & \cdots & Q_{1n_t}^{(t)} \\
             \vdots & \vdots & \ddots & \vdots \\
             R_{n_t}^{(t)} & Q_{n_t 1}^{(t)} & \cdots & Q_{n_tn_t}^{(t)}
        \end{array}
    \right]
    \\
    Q_{\a t \b t}^{(t)} &= \scalen\bm{w}^{(t)}_{\a t}\cdot \bm{w}_{\b t}^{(t)},
    \quad 
    R_{\a t}^{(t)} = \scalen \bm{w}^{(t-1)}_1\cdot \bm{w}_{\a t}^{(t)},
    \quad 
    \a{t},\b{t}=1,2,\dots,n_{t},
 \end{align}
 for a fixed set of $\{\bm{w}_{\a{t}}^{(t)}\}_{\a t=1}^{n_t}, t=0,\dots,T$. Also, the center of the cluster $\bm{v}$ appears only in the form of the inner product $m_{\a t}^{(t)}=\frac{1}{N}\bm{v}\cdot \bm{w}_{\a t}^{(t)}$. Thus, the replicated system \eqref{eq: replicated system} depends on $\{\bm{w}_{\a{t}}^{(t)}\}_{\a t=1}^{n_t}, t=0,\dots,T$ only through their inner products, such as 
\begin{align}
    &\scalen \bm{w}_{\a t}^{(t)}\cdot \bm{w}_{\b t}^{(t)}, \quad \a t,\b t=1,2,\dots,n_t, \; t=0,1,\dots T,
    \\
    &\scalen \bm{w}_{\a t}^{(t)}\cdot \bm{v}, \quad \a t = 1,\dots, n_t,\; t=0,1,\dots, T,
    \\
    &\scalen \bm{w}_{1}^{(t-1)}\cdot \bm{w}_{\a t}^{(t)}, \quad \a t =1,2,\dots,n_t,\; t=1,\dots,T,
\end{align}
which capture the geometric relations between the estimators and the centroid of clusters $\bm{v}$. We refer to them as {\it order parameters}. Furthermore, since each data point is independent, the integral over $(\bm{u}_\mu^{(0)}, y_\mu^{(0)})$ and $(\bm{u}_{\nu}^{(t)}, y_\nu^{(t)})$ can be taken independently. As a result, the generating functional does not depend on the index $\mu$ and $\nu$. 

The above observation indicates that the factors regarding the loss function in \eqref{appeq: replicated system inserted} can be evaluated as Gaussian integrals once the order parameters are fixed. To implement this idea, we insert the trivial identities of the delta functions
\begin{align}
         1 &= \prod_{t=0}^{T}\prod_{1\le\a{t}\le\b{t}\le n_{t}}N\int \delta_{\rm d}\left(
        NQ_{\a t\b t}^{(t)} - \bm{w}_{\a t}^{(t)}\cdot \bm{w}_{\b t}
    \right)dQ_{\a{t}\b{t}}^{(t)},
     \\
     1 &= \prod_{t=0}^{T} \prod_{\a t=1}^{n_{t}}N\int
        \delta_{\rm d}\left(
            Nm_{\a t}^{(t)} - \bm{w}_{\a t}^{(t)}\cdot\bm{v}
        \right)
     dm_{\a t}^{(t)},
     \\
     1 &= \prod_{t=1}^T \prod_{\a t=1}^{n_ t}N\int
        \delta_{\rm d}\left(
            NR_{\a t}^{(t)} - \bm{w}_{1}^{(t-1)}\cdot\bm{w}_{\a t}^{(t)}
        \right)
     dR_{\a t}^{(t)},
\end{align}
into \eqref{appeq: replicated system inserted}. Then the replicated system can be written as follows:
\begin{align}
    \phi^{(T)}_{n_0,\dots,n_T} &= \lim_{N,\beta\to\infty}\int
        e^{\epsilon_w g_w(\{w_{1,i}^{(t)}\}_{t=0}^{T}) 
        + \epsilon_Bg_B(\{B_1^{(t)}\}_{t=0}^{T})} 
        e^{N\alpha_L \varphi_u^{(0)} +\sum_{t=1}^T N\alpha_U \varphi_u^{(t)}}
        \nonumber\\
        &\times
        \prod_{t=0}^{T}\prod_{1\le\a{t}\le\b{t}\le n_{t}}\delta_{\rm d}\left(
            NQ_{\a t\b t}^{(t)} - \bm{w}_{\a t}^{(t)}\cdot \bm{w}_{\b t}
        \right)
        \prod_{\a t=1}^{n_{t}}
        \delta_{\rm d}\left(
            Nm_{\a t}^{(t)} - \bm{w}_{\a t}^{( t)}\cdot\bm{v}
        \right)
        \nonumber\\
        &\times
        \prod_{t=1}^T\prod_{\a t=1}^{n_ t}
        \delta_{\rm d}\left(
            NR_{\a t}^{(t)} - \bm{w}_{1}^{(t-1)}\cdot\bm{w}_{\a t}^{(t)}
        \right)
        \prod_{t=0}^T \prod_{j=1}^N \prod_{\a t=1}^{n_t}e^{-\frac{\beta^{(t)}\lambda^{(t)}}{2}(w_{\a t,i}^{(t)})^2}
    d^{n_0}\bm{\theta}^{(0)}\dots d^{n_T}\bm{\theta}^{(T)}d\Theta,
    \\
    e^{\varphi_u^{(0)}} &= \E_{\bm{u}^{(0)}  \sim\gN(\bm{0},\Sigma^{(0)}), 
    ~y^{(0)}\sim p_{y}^{(0)}}\left[
        \prod_{\a 0=1}^{n_0} e^{
            -\beta^{(0)}l(
                y^{(0)}, 
                \sigma(f_{\a 0}^{(0)})
            )
        }
    \right],
    \\
    e^{\varphi_u^{(t)}} &= \E_{\bm{u}^{(t)} \sim\gN(\bm{0},\Sigma^{(t)}), 
    ~y^{(t)}\sim p_{y}^{(t)}}\left[
        \prod_{\a t=1}^{n_t} e^{
            -\beta^{(t)}\1(|\tilde{f}^{(t)}|>\Gamma \sqrt{Q_{11}^{(t-1)}})l_{\rm pl}(
                \sigma_{\rm pl}(
                    \tilde{f}^{(t)}
                ), 
                \sigma(
                    f_{\a t}^{(t)}
                )
            )
        }
    \right]
    \\
    f_{\a 0}^{(0)} &= B_{\a 0}^{(0)} + (2y^{(0)}-1)m_{\a 0}^{(0)} + u_{\a 0}^{(0)} 
    ,
    \\
    \tilde{f}^{(t)} &= B_{1}^{(t-1)} + (2y^{(t)}-1)m_{1}^{(t-1)} + u_1^{(t)},
    \\
    f_{\a t}^{(t)} &= B_{\a t}^{(t)} + (2y^{(t)}-1)m_{\a t}^{(t)} + u_{\a t+1}^{(t)}
    ,
\end{align}
where $\Theta$ is the collection of the following variables
\begin{align}
    Q^{(t)}&=[Q_{\a t \b t}^{(t)}]_{\substack{1\le \a t \le n_t \\ 1\le \b t\le n_t}},
    \;
    \bm{m}^{(t)}=[m_{\a t}^{(t)}]_{1\le \a{t} \le n_t},
    \;
    \bm{R}^{(t)} = [R_{\a t}^{(t)}]_{1\le \a t \le n_t},
    \;
    \bm{B}^{(t)}=[b_{\a t}^{(t)}]_{1\le \a t \le n_t}.
    \label{appeq: order parameters}
\end{align}

\subsubsection{Decoupling the integrals and the saddle point method}
To complete the remaining integrals over $\{\bm{\theta}_{\a t}^{(t)}\}$, we use the Fourier representations of the delta functions:
\begin{align}
    \delta_{\rm d}\left(
        NQ_{\a t\b t}^{(t)} - \bm{w}_{\a t}^{(t)}\cdot \bm{w}_{\b t}
    \right) &= \frac{1}{4\pi}\int e^{-
        (NQ_{\a t\b t}^{(t)} - \bm{w}_{\a t}^{(t)}\cdot \bm{w}_{\b t})\frac{\tilde{Q}_{\a t \b t}^{(t)}}{2}
    }d\tilde{Q}_{\a t \b t}^{(t)},
    \\
    \delta_{\rm d}\left(
            Nm_{\a t}^{(t)} - \bm{w}_{\a t}^{(t)}\cdot\bm{v}
        \right) &= \frac{1}{2\pi}\int
        e^{
            (Nm_{\a t}^{(t)} - \bm{w}_{\a t}^{(t)}\cdot\bm{v})\tilde{m}_{\a t}^{(t)}
        }
    d\tilde{m}_{\a t}^{(t)},
    \\
    \delta_{\rm d}\left(
        NR_{\a t}^{(t)} - \bm{w}_{1}^{(t-1)}\cdot\bm{w}_{\a t}^{(t)}
    \right) & = \frac{1}{2\pi}\int
        e^{
            (NR_{\a t}^{(t)} - \bm{w}_{1}^{(t-1)}\cdot\bm{w}_{\a t}^{(t)})\tilde{R}_{\a t}^{(t)}
        }
    d\tilde{R}_{\a t}^{(t)}.
\end{align}
After using these expressions, the integrals over $\{w_{a_t, j}^{(t)}\}$ can be independently performed for each $j$. As a result, the result no longer depends on the index $i$, hence, we can safely omit it. This is a natural consequence of the data distribution having rotation-invariant symmetry. 

Specifically, we find that $\phi_{n_0,\dots,n_T}^{(T)}$ can be written as
\begin{equation}
    \phi_{n_0,\dots,n_T}^{(T)} = \lim_{N, \beta\to\infty}\int e^{N\gS(\Theta,\hat{\Theta})}\E_{\{\bm{w}^{(t)}\}\sim \tilde{p}_{{\rm eff}, w}}\left[e^{\epsilon_w g_w(\{w_1^{(t)}\}_{t=0}^T)}\right]
    e^{\epsilon_B g_B(\{B_1^{(t)}\}_{t=0}^T)}d\Theta d\hat{\Theta},
    \label{appeq: generating functional general}
\end{equation}
where $\hat{\Theta}$ is the collection of variables $
    \tilde{Q}^{(t)}=[\tilde{Q}_{\a t \b t}^{(t)}]_{\substack{1\le \a t \le n_t \\ 1\le \b t\le n_t}},
    \;
    \tilde{\bm{m}}^{(t)}=[\tilde{m}_{\a t}^{(t)}]_{1\le \a{t} \le n_t},
    \;
    \tilde{\bm{R}}^{(t)} = [\tilde{R}_{\a t}^{(t)}]_{1\le \a t \le n_t},
$ and 
\begin{align}
    \tilde{p}_{{\rm eff},w}(\{\bm{w}^{(t)}\}_{t=0}^T) &= \farc{1}{\tilde{Z}_{{\rm eff}, w}} e^{
        -\frac{1}{2}(\bm{w}^{(0)})^\top (\tilde{Q}^{(0)} + \beta^{(t)}I_{n_t})\bm{w}^{(0)} + (\tilde{\bm{m}}^{(t)})\cdot \bm{w}^{(0)} 
    }
    \nonumber \\
    &\hspace{20truemm}\times\prod_{t=1}^T e^{
        -\frac{1}{2}(\bm{w}^{(t)})^\top (\tilde{Q}^{(t)} + \beta^{(t)}I_{n_t})\bm{w}^{(t)} + (\tilde{\bm{m}}^{(t)} + w_1^{(t-1)}\bm{R}^{(t)})\cdot \bm{w}^{(t)} 
    },
    \\
    \gS(\Theta, \hat{\Theta}) &= (1-1/N)\alpha_L \varphi_u^{(0)} + \sum_{t=1}^T(1-1/N)\alpha_U^{(t)} \varphi_u^{(t)} + (1-1/N)\varphi_w,
    \label{appeq: gibbs general}
    \\
    \varphi_w &= \log \tilde{Z}_{{\rm eff},w},
\end{align}
where the factors $1/N$ that appears in the coefficients of $\varphi_u^{(t)}$ and $\varphi_w$ are due to the presence of $g_w$ and $g_B$. Also $\bm{w}^{(t)} = (w_1^{(t)}, \dots w_{n_t}^{(t)})\cdot \in \R^{n_t}$. Thus, $\Theta$ and $\hat{\Theta}$ are evaluated by the extremum condition $\mathop{\rm extr}_{\Theta, \hat{\Theta}}\gS(\Theta, \hat{\Theta})$, in which the above factor $1/N$ is negligible. Recall that the index $1$ in \eqref{eq: generating functional general} is the replica index, and the result no longer depends on the index $i\in[N]$. It should also be noted that, essentially, only Gaussian integrals and the saddle-point method are required for the above calculations once the order parameters are determined. This computational simplicity is a major advantage of the replica method that introduces the replicated system \eqref{appeq: replicated system}, which enables us to consider the average over the noise before completing the integral (or the optimization at $\beta\to\infty$) over $\{\bm{\theta}_{\a t}\}_{t=0}^T$. 

To obtain the saddle point condition regarding $\hat{\Theta}$, it is useful to use the following identity for the Gaussian integral with mean $\bar{\bm{\mu}}$ and covariance $S$ and twice-differentiable function $\gF$:
\begin{equation}
    \E_{\bm{z}\sim\gN(\bar{\bm{\mu}}, S)}[\gF(\bm{z})] = \left.\exp\left(
        \frac{1}{2}\sum_{i,j}S_{ij}\frac{\partial^2}{\partial h_{i}\partial h_j}
    \right)\gF(\bm{h})\right|_{\bm{h}=\bar{\bm{\mu}}},
    \label{appeq: identiy gaussian}
\end{equation}
which is commonly used in the mean-field theory of glassy systems \citep{parisi2020theory}. The right-hand side of \eqref{appeq: identiy gaussian} is defined by a formal Taylor expansion of the exponential function:
\begin{equation}
    \exp\left(
        \frac{1}{2}\sum_{i,j}S_{ij}\frac{\partial^2}{\partial h_{i}\partial h_j}
    \right)\gF(\bm{h})
    \equiv \left(
        1 + \frac{1}{2}\sum_{i,j}S_{ij}\frac{\partial^2}{\partial h_{i}\partial h_j} + \dots
    \right)\gF(\bm{h}).
\end{equation}
Let $\alpha^{(0)}$ and $\Delta^{(0)}$ be $\alpha_L$ and $\Delta_L$, respectively.
Let $\alpha^{(t)}$ be $\alpha_L$ if $t=0$ and $\alpha_U$ otherwise.  Also, define $\tilde{p}_{{\rm eff}, u|y}^{(t)}, \tilde{p}_{{\rm eff}, uy}^{(t)}$ and $l_{\a t}^{(t)}$ as follows:
\begin{align}
    \tilde{p}_{{\rm eff}, u|y}^{(t)}(\bm{u}^{(t)}|y^{(t)}) &\propto 
        \prod_{\a t=1}^{n_t} e^{
            -\beta^{(t)}l_{\a t}^{(t)}
        }\times\gN(\bm{u}^{(t)}; \bm{0}, \Sigma^{(t)})
        ,
    \\
	\tilde{p}_{{\rm eff}, uy}^{(t)}(\bm{u}^{(t)}, y^{(t)}) & = \tilde{p}_{{\rm eff}, u|y}(\bm{u}^{(t)}|y^{(t)})p_y^{(t)}(y^{(t)}),
    \\
    l_{\a 0}^{(0)} &= l_{\a 0}^{(0)}(y^{(0)}, f_{\a 0}^{(0)}) = l\left(y^{(0)}, \sigma(f_{\a 0}^{(0)})\right),
    \\
    l_{\a t}^{(t)} &= l_{\a t}^{(t)}(\tilde{f}^{(t)}, f_{\a t}^{(t)}) = \1\left(|\tilde{f}^{(t)}|>\Gamma\sqrt{Q_{11}^{(t-1)}}\right)l_{\rm pl}\left(\sigma_{\rm pl}(\tilde{f}^{(t)}), \sigma(f_{\a t}^{(t)})\right).
\end{align}

Then, by using the identity \eqref{appeq: identiy gaussian} for expressing the Gaussian average in $\varphi_u^{(t)}$, the saddle point condition over $(\Theta ,\hat{\Theta})$ can be written as follows:
\begin{align}
    Q_{\a t \b t}^{(t)} &= \E_{\{\bm{w}^{(t)}\}\sim \tilde{p}_{{\rm eff}, w}}\left[w_{\a t}^{(t)}w_{\b t}^{(t)}\right],
    \\
    R_{\a t}^{(t)} &= \E_{\{\bm{w}^{(t)}\}\sim \tilde{p}_{{\rm eff}, w}}\left[w_{\a t}^{(t)}w_1^{(t-1)}\right],
    \\
    m_{\a t}^{(t)} &= \E_{\{\bm{w}^{(t)}\}\sim \tilde{p}_{{\rm eff}, w}}\left[w_{\a t}^{(t)}\right],
    \\
    \tilde{Q}_{\a t \b t}^{(t)} &= -\alpha^{(t)}\Delta^{(t)}\E_{
    		\bm{u}^{(t)}, y^{(t)}\sim p_{{\rm eff}, uy}^{(t)}
	}\left[
        (\beta^{(t)})^2 \partial_2 l_{\a t}^{(t)}\partial_2 l_{\b t}^{(t)}
        -\beta^{(t)} \partial_2^2 l_{\a t}^{(t)} \delta_{\a t, \b t}
    \right],
    \\
    \tilde{R}_{\a t \b t}^{(t)} &= \alpha^{(t)}\Delta^{(t)}\beta^{(t)}\E_{
    		\bm{u}^{(t)}, y^{(t)}\sim p_{{\rm eff}, uy}^{(t)}
	}\left[
        \partial_1\partial_2 l_{\a t}^{(t)}
    \right],
    \\
    \tilde{m}_{\a t}^{(t)} &= -\alpha^{(t)} \beta^{(t)}\E_{
    		\bm{u}^{(t)}, y^{(t)}\sim p_{{\rm eff}, uy}^{(t)}
	}\left[
        (2y^{(t)}-1)\partial_2 l_{\a t}^{(t)}
    \right],
    \\
    0 &= \E_{\bm{u}^{(t)}\sim p_{{\rm eff}, u}^{(t)}y\sim p_y^{(t)}}\left[
        \partial_2 l_{\a t}^{(t)}
    \right].
\end{align}
Recall that, for a bivariate function $\gF(y, x)$, let $\partial_i\gF$ be the partial derivative of $\gF$ with respect to the $i$-th argument as defined in subsection \ref{subsec: notation}. For example, $\partial_1 \gF(Y, X) = \left.\frac{\partial \gF}{\partial y}\right|_{y=Y, x=X}$. Higher-order derivatives are defined similarly.  In fact, he saddle point condition regarding $Q_{11}^{(t)}, m_1^{(t)}$ and $B_1^{(t)}, t\in[T-1]$ yields the derivative regarding $l_{\a{t+1}}^{(t+1)}$. However, such terms are an effect from step $t+1$ to $t$, which is causally inconsistent. Hence we omit them. In fact, a rigorous calculation reveals that these terms vanish in the limit as $n_{t+1}\to0$ under the RS assumption that will be introduced below.

\subsubsection{RS saddle point condition}
Since equation \eqref{appeq: generating functional general} depends on the discrete nature of $n_t$, the extrapolation of $\lim_{n\to0}\phi_{n_0,\dots,n_T}^{(T)}$ is still difficult. The key issue here is to identify the correct form of the saddle point. The simplest form is the following RS saddle point:
\begin{align}
    Q^{(t)} &= \frac{\chi^{(t)}}{\beta^{(t)}}I_{n_t} + q^{(t)}\bm{1}_{n_t}\bm{1}_{n_t}\cdot,
    \label{appeq:RS assumption1}
    \\
    \bm{R}^{(t)} &= R^{(t)}\bm{1}_{n_t},
    \label{appeq:RS assumption2}
    \\
    \bm{m}^{(t)} &= m^{(t)}\bm{1}_{n_t},
    \label{appeq:RS assumption3}
    \\
    \bm{B}^{(t)} &= B^{(t)}\bm{1}_{n_t},
    \label{appeq:RS assumption4}
    \\
    \tilde{Q}^{(t)} &= \beta^{(t)}\hat{Q}^{(t)}I_{n_t} - (\beta^{(t)})^2 \hat{\chi}^{(t)} \bm{1}_{n_t}\bm{1}_{n_t}\cdot,
    \label{appeq:RS assumption5}
    \\
    \tilde{\bm{R}}^{(t)} &= \beta^{(t)}\hat{R}^{(t)}\bm{1}_{n_t},
    \label{appeq:RS assumption6}
    \\
    \tilde{\bm{m}}^{(t)} &= \beta^{(t)}\hat{m}^{(t)}\bm{1}_{n_t},
    \label{appeq:RS assumption7}
\end{align}
which is the simplest form of the saddle point reflecting the symmetry of the variational function. This choice is motivated by the fact that $\mathcal{S}^{(T)}(\Theta, \hat{\Theta})$ is invariant under the permutation of the indices $\a t\in\{1,2,\dots,n_t\}$ for any $t=0,1,\dots$. In a mathematically rigorous sense, the expression for the replicated system \eqref{eq: replicated system} with integer $\{n_t\}_{t=0}^T$ alone cannot uniquely determine the expression for the replicated system at real $\{n_t\}_{t=0}^T$. However, for log-conevx Boltzmann distributions, it is empirically known that the replica symmetric choice of the saddle point yields the correct extrapolation \citep{barbier_adaptive_2019, barbier2019optimal, mignacco2020role, gerbelot2020asymptotic, gerbelot2023asymptotic, montanari2024} in the sense that the same result by the replica method with the RS assumption have been obtained through a different mathematically rigorous approach. Since the Boltzmann distributions are log-convex functions once conditioned on the data and the parameter of the previous iteration step, we can expect the RS assumption to yield the correct result even in the current iterative optimization.

Under the RS assumption, after simple algebra, $\tilde{p}_{{\rm eff}, w}$ and $\tilde{p}_{{\rm eff}, u|y}$ can be rewritten as follows:
\begin{align}
    \tilde{p}_{{\rm eff}, w}(\{\bm{w}^{(t)}\}) &= \E_{\xi_w^{(t)}\sim_{\rm iid}\gN(0,1)}\left[
        \prod_{\a 0=1}^{n_0}\gN\left(
            w^{(0)}_{\a 0}
            \mid
            \frac{\hat{m}^{(0)}+\sqrt{\hat{\chi}^{(0)}}\xi_w^{(0)}}{\hat{Q}^{(0)}+\lambda^{(0)}},
            \frac{\hat{Q}^{(0)}+\lambda^{(0)}}{\beta^{(0)}}
        \right)
        \right.
        \nonumber \\
        &\left.
        \times\prod_{t=1}^T\prod_{\a t=1}^{n_t}\gN\left(
            w^{(t)}_{\a t}
            \mid
            \frac{\hat{m}^{(t)} + \hat{R}^{(t)}w_1^{(t-1)} +\sqrt{\hat{\chi}^{(t)}}\xi_w^{(t)}}{\hat{Q}^{(t)}+\lambda^{(t)}},
            \frac{\hat{Q}^{(t)}+\lambda^{(t)}}{\beta^{(t)}}
        \right)
    \right],
    \\
    \tilde{p}_{{\rm eff}, u|y}^{(0)}(\bm{u}^{(0)}|y^{(0)}) &= \E_{\xi_u^{(0)}\sim\gN(0,\Delta_L)}\left[
        \prod_{\a 0=1}^{n_0}e^{
            -\beta^{(0)}l\left(
                    y^{(0)}, \sigma(h_u^{(0)} + u_{\a 0}^{(0)})
            \right)
        }\gN(u_{\a 0}^{(0)}\mid 0, \frac{\chi^{(0)}\Delta_L}{\beta^{(0)}})
    \right],
    \\
    \tilde{p}_{{\rm eff}, u|y}^{(t)}(\bm{u}^{(t)}|y^{(t)}) &= \E_{\xi_{u,1}^{(t)}, \xi_{u,2}^{(t)}\sim_{\rm iid}\gN(0,\Delta_U^{(t)})}\left[
        \prod_{\a t=1}^{n_t}e^{
            -\beta^{(t)}l_{\rm pl}\left(
                    \sigma_{\rm pl}(
                        \tilde{h}_u^{(t)} + \tilde{u}^{(t)}
                    ), \sigma(
                        h_u^{(t)} + u_{\a t}^{(t)}
                    )
            \right)
        }
    \right.
    \nonumber \\
    &\hspace{30truemm}\left.
        \times\gN(u_{\a t}^{(t)}\mid 0, \frac{\chi^{(t)}\Delta_U^{(t)}}{\beta^{(t)}}) 
        \gN(\tilde{u}^{(t)}\mid 0, \frac{\chi^{(t-1)}\Delta_U^{(t)}}{\beta^{(t-1)}})
    \right],
    \label{appeq: effective density u integer n}
    \\
    h_{u}^{(0)} &= B^{(0)} + (2y^{(t)}-1)m^{(0)} + \sqrt{q^{(0)}}\xi_u^{(0)},
    \\
    \tilde{h}_u^{(t)} &= B^{(t-1)} + (2y^{(t)}-1)m^{(t-1)} + \sqrt{q^{(t-1)}}\xi_{u,1}^{(t)},
    \\
    h_u^{(t)} &= B^{(t)} + (2y^{(t)}-1)m^{(t)} + \frac{R^{(t)}}{\sqrt{q^{(t-1)}}}\xi_{u,1}^{(t)} + \sqrt{q^{(t)} - \frac{(R^{(t)})^2}{q^{(t-1)}}}\xi_{u,2}^{(t)}.
\end{align}
Importantly, the factors inside the average over $\xi_w^{(t)}, \xi_{u,1}^{(t)}, \xi_{u,2}^{(t)}$ are factorized over the replica indexes. This indicates that the integral over $w_{\a t}^{(t)}, u_{\a t}^{(t)}$ are taken independently before the taking the averaging over $\xi_w^{(t)}, \xi_{u,1}^{(t)}, \xi_{u,2}^{(t)}$, which makes the computation considerably simple. Also, due to this symmetry, the resultant expressions can often be formally extrapolated from $n_t\in \sN$ to $n_t\in\R$.  To see this, first, let us consider $\varphi_u^{(t)}$. By conducting straightforward integrals, we obtain the following 
\begin{align}
	e^{\varphi_u^{(0)}} &= \E_{y^{(0)}, \xi_{u}^{(0)}}\left[
		\left(\int e^{
            -\beta^{(0)}l\left(
                    y^{(0)}, \sigma(h_u^{(0)} + u^{(0)})
            \right)
        }\gN(u^{(0)}\mid 0, \frac{\chi^{(0)}\Delta_L}{\beta^{(0)}})
        du^{(0)}
        \right)^{n_0}
	\right],
	\\
	e^{\varphi_u^{(t)}} &=\E_{y^{(t)}, \xi_{u,1}^{(t)}, \xi_{u,2}^{(t)}, \tilde{u}^{(t)}}\left[
		\left(
			\int 
				e^{
                      -\beta^{(t)}l_{\rm pl}\left(
                              \sigma_{\rm pl}(
                                  \tilde{h}_u^{(t)} + \tilde{u}^{(t)}
                              ), \sigma(
                                  h_u^{(t)} + u^{(t)}
                              )
                      \right)
                  }
                \gN(u^{(t)}\mid 0, \frac{\chi^{(t)}\Delta_U^{(t)}}{\beta^{(t)}}) 
			du^{(t)}
		\right)^{n_t}
	\right].
\end{align}
Next, let us consider $\varphi_w$. For this, let us define $p_{{\rm eff}, w}^{(t)}$ as follows.
\begin{align}
	p_{{\rm eff}, w}^{(0)}(w^{(0)}) &= \frac{1}{Z_{{\rm eff}, w}^{(0)}}e^{-\beta^{(0)}(
		\frac{\hat{Q}^{(0)} + \lambda^{(0)}}{2}(w^{(0)})^2 - (\hat{m}^{(0)} + \sqrt{\hat{\chi}^{(0)}}\xi_{w}^{(t)})w^{(0)}
	)},
	\\
	p_{{\rm eff}, w}^{(t)}(w^{(t)}) &= \frac{1}{Z_{{\rm eff}, w}^{(t)}(w^{(t-1)})}e^{-\beta^{(t)}(
		\frac{\hat{Q}^{(t)}+\lambda^{(t)}}{2}(w^{(t)})^2 - (\hat{m}^{(t)} + \hat{R}^{(t)}w^{(t-1)}+\sqrt{\hat{\chi}^{(t)}}\xi_w^{(t)})w^{(t)}
	)},
	\\
	 w^{(t-1)} \sim p_{{\rm eff}, w}^{(t-1)}.
\end{align}
Then, by performing the integrals successively on $\{w_{\a T}^{(T)}\}, \{w_{\a{T-1}}^{(T-1)}\}, \dots$, we obtain the following:
\begin{align}
	\Phi^{(T)}(w^{(T-1)}) &= (Z_{{\rm eff}^{(T)}}(w^{(T-1)}))^{n_T}, \quad w^{(T-1)}\sim p_{{\rm eff}, w}^{(T-1)},
	\\
	\Phi^{(t)}(w^{(t-1)}) &= \E_{w^{(t)}\sim p_{{\rm eff}, w}^{(t)}}[\Phi^{(t+1)}(w^{(t)})] (Z_{{\rm eff}, w}^{(t)}(w^{(t-1)}))^{n_{t}}, 
	\nonumber 
	\\
	&\hspace{40truemm}w^{(t-2)}\sim p_{{\rm eff}, w}^{(t-2)}, \quad t=T-1,\dots, 2, 1,
	\\
	e^{\varphi_w} &= \E_{w^{(0)}\sim p_{{\rm eff}, w}^{(0)}}[\Phi^{(1)}(w^{(0)})] (Z_{{\rm eff}, w}^{(0)})^{n_0}.
\end{align}
By formally considering $(\dots)^{n_t}$ as a exponential function of $n_t$, we can extrapolate the above result to $n_t\in\R$. Then, it is clear that $\varphi_u^{(t)}$ and $\varphi_w$ are expanded as $1+\gO(n)$, where $\mathcal{O}(n)\equiv\sum_{t=0}^{T}\mathcal{O}(n_t)$ be terms that vanish at the limit $n_0,n_1,\dots,n_T\to0$. From this, it can be straightforwardly shown that $\gS$ evaluated as the saddle point is also expanded as $\gS = \gO(n)$. Therefore, $\phi_{n_0,\dots,n_T}^{(T)}$ can be evaluated as follows:
\begin{align}
    \phi_{n_0, \dots,n_T}^{(T)} &= \lim_{\beta\to\infty}\E_{\xi_w^{(t)}\sim_{\rm iid}\gN(0,1)}\left[
        \E_{\{w^{(t)}\}\sim p_{{\rm eff}, w}(\{w^{(t)}\}|\{\xi_w^{(t)}\})}\left[
            e^{\epsilon_w g_w(\{w^{(t)}\}_{t=0}^T)}
        \right]
    \right]e^{\epsilon_B g_B(\{B^{(t)}\}_{t=0}^T)} +\gO(n),
\end{align}
where
\begin{align}
    p_{{\rm eff}, w}(\{w^{(t)}\}_{t=0}^T|\{\xi_w^{(t)}\}) &= \gN\left(
        w^{(0)}
        \mid
        \frac{\hat{m}^{(0)}+\sqrt{\hat{\chi}^{(0)}}\xi_w^{(0)}}{\hat{Q}^{(0)}+\lambda^{(0)}},
        \frac{\hat{Q}^{(0)}+\lambda^{(0)}}{\beta^{(0)}}
    \right)
    \nonumber \\
    &\times\prod_{t=1}^T\gN\left(
        w^{(t)}
        \mid
        \frac{\hat{m}^{(t)} + \hat{R}^{(t)}w^{(t-1)} +\sqrt{\hat{\chi}^{(t)}}\xi_w^{(t)}}{\hat{Q}^{(t)}+\lambda^{(t)}},
        \frac{\hat{Q}^{(t)}+\lambda^{(t)}}{\beta^{(t)}}
    \right).
\end{align}
The parameters such as $q^{(t)}, \chi^{(t)}, \dots $ are determined by the saddle point conditions. Also, at the limit $\beta\to\infty, n_0, \dots,n_T\to0$, the measure defined by $p_{{\rm eff}, w}$ concentrates, and it is governed by a one-dimensional Gaussian process:
\begin{align}
    \Xi_{\rm ST}(\epsilon_w, \epsilon_B) &= \E_{\xi_w^{(t)} \sim_{\rm iid}\gN(0,1)}\left[
        e^{\epsilon_w g_w(\{\wsf{t}\}_{t=0}^T)}
    \right]
    e^{\epsilon_B g_B(\{B^{(t)}\}_{t=0}^T)},
    \label{appeq: generating functional rs 1}
    \\
    \wsf{0} &= \frac{\hat{m}^{(0)}+\sqrt{\hat{\chi}^{(0)}}\xi_w^{(0)}}{\hat{Q}^{(0)}+\lambda^{(0)}},
    \label{appeq: effective w 1}
    \\
    \wsf{t} &= \frac{\hat{m}^{(t)} + \hat{R}^{(t)}\wsf{t-1}+\sqrt{\hat{\chi}^{(t)}}\xi_w^{(t)}}{\hat{Q}^{(t)}+\lambda^{(t)}},
    \label{appeq: effective w 2}
\end{align}
where, $\Theta, \hat{\Theta}$ are determined by the saddle point condition.

Moreover, similar integrals over $w_{\a t}^{(t)}, \tilde{u}^{(t)}$ and $u_{\a t}^{(t)}$ considered above yields the expressions of the saddle point condition under RS assumption as follows: 
\begin{align}
    q^{(t)} &= \E_{\xi_w^{(t)}\sim_{\rm iid}\gN(0,1)}\left[
        \E_{\{w^{(t)}\}\sim p_{{\rm eff}, w}(\{w^{(t)}\}|\{\xi_w^{(t)}\})}\left[
            w^{(t)}
        \right]^2
    \right] + \gO(n),
    \\
    \chi^{(t)} &= \E_{\xi_w^{(t)}\sim_{\rm iid}\gN(0,1)}\left[
        \beta^{(t)}\left(
            \sV{\rm ar}_{\{w^{(t)}\}\sim p_{{\rm eff}, w}(\{w^{(t)}\}|\{\xi_w^{(t)}\})}\left[
                w^{(t)}
            \right]
        \right)
    \right] + \gO(n),
    \label{appeq: chi t}
    \\
    &= \E_{\xi_w^{(t)}\sim_{\rm iid}\gN(0,1)}\left[
            \frac{\partial }{\partial (\sqrt{\hat{\chi}^{(t)}\xi_w^{(t)}})}\E_{\{w^{(t)}\}\sim p_{{\rm eff}, w}(\{w^{(t)}\}|\{\xi_w^{(t)}\})}\left[
                w^{(t)}
        \right]
    \right] + \gO(n),
    \\
    R^{(t)} &= \E_{\xi_w^{(t)}\sim_{\rm iid}\gN(0,1)}\left[
        \E_{\{w^{(t)}\}\sim p_{{\rm eff}, w}(\{w^{(t)}\}|\{\xi_w^{(t)}\})}\left[
            w^{(t)}w^{(t-1)}
        \right]
    \right] + \gO(n),
    \\
    m^{(t)} &= \E_{\xi_w^{(t)}\sim_{\rm iid}\gN(0,1)}\left[
        \E_{\{w^{(t)}\}\sim p_{{\rm eff}, w}(\{w^{(t)}\}|\{\xi_w^{(t)}\})}\left[
            w^{(t)}
        \right]
    \right] + \gO(n).
\end{align}
For $\hat{\Theta}$, we need to write separately for $t=0$ and $t\ge1$. For $t=0$,
\begin{align}
    \hat{Q}^{(0)} &= \alpha_L\Delta_L \E_{\xi\sim\gN(0,\Delta_L), y^{(0)}\sim p_{y,L}}\left[
            \frac{d }{d h_u^{(0)}}\E_{u^{(0)}\sim p_{{\rm eff}, u|y}^{(0)}}\left[
                \partial_2 l^{(0)}
            \right]
         \right]+ \gO(n),
    \\
    \hat{\chi}^{(0)} &= \alpha_L\Delta_L \E_{\xi\sim\gN(0,\Delta_L), y^{(0)}\sim p_{y,L}}\left[
            \E_{u^{(0)}\sim p_{{\rm eff}, u|y}^{(0)}}\left[
                \partial_2 l^{(0)}
            \right]^2
         \right] + \gO(n),
    \\
    \hat{m}^{(0)} &= -\alpha_L \E_{\xi\sim\gN(0,\Delta_L), y^{(0)}\sim p_{y,L}}\left[
            (2y^{(0)}-1)\E_{u^{(0)}\sim p_{{\rm eff}, u|y}^{(0)}}\left[
                \partial_2 l^{(0)}
            \right]
         \right] + \gO(n)
    \\
    0 &= \E_{\xi\sim\gN(0,\Delta_L), y^{(0)}\sim p_{y,L}}\left[
            \E_{u^{(0)}\sim p_{{\rm eff}, u|y}^{(0)}}\left[
                \partial_2 l^{(0)}
            \right]
         \right] + \gO(n),
\end{align}
where 
\begin{align}
    l^{(0)} &=l^{(0)}(y, h_u^{(0)} + u^{(0)}) = l(y, \sigma(h_u^{(0)} + u^{(0)})),
    \\
    p_{{\rm eff}, u|y}^{(0)}(u|y^{(0)}) &= \gN(u^{(0)}\mid 0, \frac{\chi^{(0)}\Delta_L}{\beta^{(0)}})e^{-\beta^{(0)}l^{(0)}},
    \\
    h_u^{(0)} &= B^{(0)} + (2y^{(0)}-1) m^{(t)} + \sqrt{\hat{\chi}^{(t)}}\xi_u^{(0)}.
\end{align}
For $t\ge1$,
\begin{align}
    \\
    \hat{Q}^{(t)} &=  \alpha_U^{(t)}\Delta_U ^{(t)}\E_{\xi_{u,1}^{(t)},\xi_{u,2}^{(t)}\sim\gN(0,\Delta_U)^{(t)}, y^{(t)}\sim p_{y}^{(t)}}\left[
            \frac{d }{d h_u^{(t)}}\E_{\tilde{u}^{(t)}, u^{(t)} \sim p_{{\rm eff}, u|y}^{(t)}}\left[
                \partial_2 l^{(t)}
            \right]
         \right]  + \gO(n).
    \\
    \hat{\chi}^{(t)} &= \alpha_U^{(t)}\Delta_U^{(t)} \E_{\xi_{u,1}^{(t)},\xi_{u,2}^{(t)}\sim\gN(0,\Delta_U^{(t)}), y^{(t)}\sim p_{y}^{(t)}}\left[
            \E_{\tilde{u}^{(t)}, u^{(t)} \sim p_{{\rm eff}, u|y}^{(t)}}\left[
                \partial_2 l^{(t)}
            \right]^2
         \right]  + \gO(n).
    \\
    \hat{R}^{(t)} &= -\alpha_U^{(t)}\Delta_U^{(t)} \E_{\xi_{u,1}^{(t)},\xi_{u,2}^{(t)}\sim\gN(0,\Delta_U^{(t)}), y^{(t)}\sim p_{y}^{(t)}}\left[
            \frac{d }{d \tilde{h}_u^{(t)}}\E_{\tilde{u}^{(t)}, u^{(t)} \sim p_{{\rm eff}, u|y}^{(t)}}\left[
                \partial_2 l^{(t)}
            \right]
         \right]  + \gO(n).
    \\
    \hat{m}^{(t)} &= -\alpha_U^{(t)} \E_{\xi_{u,1}^{(t)},\xi_{u,2}^{(t)}\sim\gN(0,\Delta_U^{(t)}), y^{(t)}\sim p_{y}^{(t)}}\left[
            (2y^{(t)}-1)\E_{\tilde{u}^{(t)}, u^{(t)} \sim p_{{\rm eff}, u|y}^{(t)}}\left[
                \partial_2 l^{(t)}
            \right]
         \right]  + \gO(n).
    \\
    0 &= \E_{\xi_{u,1}^{(t)},\xi_{u,2}^{(t)}\sim\gN(0,\Delta_U^{(t)}), y^{(t)}\sim p_{y}^{(t)}}\left[
            \E_{\tilde{u}^{(t)}, u^{(t)} \sim p_{{\rm eff}, u|y}^{(t)}}\left[
                \partial_2 l^{(t)}
            \right]
         \right]  + \gO(n).
\end{align}
where 
\begin{align}
    l^{(t)} &= l^{(t)}(\tilde{h}_{u}^{(t)}+\tilde{u}^{(t)}, h_u^{(t)}+u^{(t)}) 
    \nonumber \\
    &= \1\left(|\tilde{h}_{u}^{(t)}+\tilde{u}^{(t)}|>\Gamma\sqrt{q^{(t-1)}}\right)l_{\rm pl}\left(\sigma_{\rm pl}(\tilde{h}_{u}^{(t)}+\tilde{u}^{(t)}), \sigma(h_u^{(t)}+u^{(t)})\right),
    \\
    p_{{\rm eff}, u|y}^{(t)}(u|y^{(t)}) &= \gN(u^{(0)}\mid 0, \frac{\chi^{(0)}\Delta_L}{\beta^{(0)}})
    \gN(0,\frac{\chi^{(t-1)}\Delta_U^{(t)}}{\beta^{(t-1)}})e^{-\beta^{(t)}l^{(t)}},
    \\
    \tilde{h}_u^{(t)} &= B^{(t-1)} + (2y^{(t)}-1)m^{(t-1)} \sqrt{q^{(t-1)}}\xi_{u,1}^{(t)},
    \\
    h_{u}^{(t)} &= B^{(t)} + (2y^{(t)}-1)m^{(t)} + \frac{R^{(t)}}{\sqrt{q^{(t-1)}}}\xi_{u,1}^{(t)} + \sqrt{q^{(t)}-\frac{(R^{(t)})^2}{q^{(t-1)}}}\xi_{u,2}^{(t)}.
\end{align}
At the limit $\beta\to\infty$, in addition to $p_{{\rm eff}, w}$, $p_{{\rm eff},u}^{(t)}$ also concentrates, and the average over $u^{(t)}, \tilde{u}^{(t)}$ are governed by one-dimensional effective problem with Gaussian disorder. Let us define $\usf{t}$ as follows:
\begin{align}
    \usf{0} &= \arg\min_{u^{(0)}\in\R}\left[
        \frac{(u^{(0)})^2}{2\Delta_L\chi^{(0)}} + l\left(
            y^{(0)},\sigma\left(
                h_u^{(0)} + u^{(0)}
            \right)
        \right)
    \right],
    \label{appeq: effective problem u 0}
    \\
    \usf{t} &= \arg\min_{u^{(t)}\in\R}\left[
        \frac{(u^{(t)})^2}{2\Delta_U\chi^{(t)}} + \1(|\tilde{h}^{(t)}|>\Gamma \sqrt{q^{(t-1)}})l_{\rm pl}\left(
            \sigma_{\rm pl}\left(
                \tilde{h}_u^{(t)}
            \right),
            \sigma\left(
                h_u^{(t)} +  u^{(t)}
            \right)
        \right)
    \right].
    \label{appeq: effective problem u t}
\end{align}
Also, let us define $l_L(y,x)$ and $l_U(y,x;\tilde{\Gamma})$ as 
\begin{align}
    l_L(y,x) &= l(y, \sigma(x)),
    \\
    l_U(y, x;\tilde{\Gamma}) &= \1(|y|>\tilde{\Gamma})l_{\rm pl}(\sigma_{\rm pl}(y), \sigma(x)).
\end{align}
Then, the RS saddle point conditions are written at the limit $\beta\to\infty, n\to0$ are summarized as follows:
\begin{align}
    q^{(0)} &= \E_{\xi_w^{(0)}\sim_{\rm iid}\gN(0,1)}\left[
        (\wsf{0})^2
    \right],
    \\
    \chi^{(0)} &= \E_{\xi_w^{(0)}\sim_{\rm iid}\gN(0,1)}\left[
        \frac{\partial }{\partial (\sqrt{\hat{\chi}^{(0)}\xi_w^{(0)}})} \wsf{0}
    \right],
    \\
    m^{(0)} &= \E_{\xi_w^{(0)}\sim_{\rm iid}\gN(0,1)}\left[
        \wsf{0}
    \right],
    \\
    \hat{Q}^{(0)} &= \alpha_L\Delta_L\E_{\xi_u^{(0)}\sim\gN(0,\Delta_L), y^{(0)}\sim p_{y,L}}\left[
        \frac{d }{d h_u^{(0)}}\partial_2 l_L(y^{(0)}, h_u^{(0)} + \usf{0})
    \right],
    \\
    \hat{\chi}^{(0)} &= \alpha_L\Delta_L\E_{\xi_u^{(0)}\sim\gN(0,\Delta_L), y^{(0)}\sim p_{y,L}}\left[
            \left(\partial_2 l_L(y^{(0)}, h_u^{(0)} + \usf{0})\right)^2
         \right],
    \\
    \hat{m}^{(0)} &= \alpha_L \E_{\xi_u^{(0)}\sim\gN(0,\Delta_L), y^{(0)}\sim p_{y,L}}\left[
            (2y-1)\partial_2 l_L(y^{(0)}, h_u^{(0)} + \usf{0})
         \right],
    \\
    0 &= \E_{\xi_u^{(0)}\sim\gN(0,\Delta_L), y^{(0)}\sim p_{y,L}}\left[
            \partial_2 l_L(y^{(0)}, h_u^{(0)} + \usf{0})
         \right],
\end{align}
and 
\begin{align}
    q^{(t)} &= \E_{\xi_w^{(t)}\sim_{\rm iid}\gN(0,1)}\left[
        (\wsf{t})^2
    \right],
    \\
    \chi^{(t)} &= \E_{\xi_w^{(t)}\sim_{\rm iid}\gN(0,1)}\left[
        \frac{\partial }{\partial (\sqrt{\hat{\chi}^{(t)}\xi_w^{(t)}})} \wsf{t}
    \right],
    \\
    R^{(t)} &= \E_{\xi_w^{(t)}\sim_{\rm iid}\gN(0,1)}\left[
        \wsf{t}\wsf{t-1}
    \right],
    \\
    m^{(t)} &= \E_{\xi_w^{(t)}\sim_{\rm iid}\gN(0,1)}\left[
        \wsf{t}
    \right],
    \\
    \hat{Q}^{(t)} &= \alpha_U^{(t)}\Delta_U^{(t)}\E_{\xi_{u,1}^{(t)}, \xi_{u,2}^{(t)}\sim\gN(0,\Delta_U^{(t)}), y^{(t)}\sim p_{y}^{(t)}}\left[
        \frac{d }{d h_u^{(t)}}\partial_2 l_U(\tilde{h}_u^{(t)}, h_u^{(t)} + \usf{t};\Gamma\sqrt{q^{(t-1)}}) 
    \right],
    \\
    \hat{R}^{(t)} &= -\alpha_U^{(t)}\Delta_U^{(t)}\E_{\xi_{u,1}^{(t)}, \xi_{u,2}^{(t)}\sim\gN(0,\Delta_U^{(t)}), y^{(t)}\sim p_{y}^{(t)}}\left[
        \frac{d }{d \tilde{h}_u^{(t)}}\partial_2 l_U(\tilde{h}_u^{(t)}, h_u^{(t)} + \usf{t};\Gamma\sqrt{q^{(t-1)}})
    \right]
    \\
    \hat{\chi}^{(t)} &=  \alpha_U^{(t)}\Delta_U^{(t)} \E_{\xi_{u,1}^{(t)}, \xi_{u,2}^{(t)}\sim\gN(0,\Delta_U^{(t)}), y^{(t)}\sim p_{y}^{(t)}}\left[
            \left(
                \partial_2 l_U(\tilde{h}_u^{(t)}, h_u^{(t)} + \usf{t};\Gamma\sqrt{q^{(t-1)}})
            \right)^2
         \right],
    \\
    \hat{m}^{(t)} &= \alpha_U^{(t)} \E_{\xi_{u,1}^{(t)}, \xi_{u,2}^{(t)}\sim\gN(0,\Delta_U^{(t)}), y^{(t)}\sim p_{y}^{(t)}}\left[
            (2y^{(t)}-1)\partial_2 l_U(\tilde{h}_u^{(t)}, h_u^{(t)} + \usf{t};\Gamma\sqrt{q^{(t-1)}})
         \right],
    \\
    0 &= \E_{\xi_{u,1}^{(t)}, \xi_{u,2}^{(t)}\sim\gN(0,\Delta_U^{(t)}), y^{(t)}\sim p_{y}^{(t)}}\left[
            \partial_2 l_U(\tilde{h}_u^{(t)}, h_u^{(t)} + \usf{t};\Gamma\sqrt{q^{(t-1)}})
         \right],
\end{align}
for $t\ge1$.  Finally, taking the average over $\xi_w^{(t)}$ yields the self-consistent equation defined in \ref{def: self-consistent equations}. The parameters are determined so that they self-consistently satisfy the saddle-point condition. Hence the above saddle point equations are termed {\it self-consistent equations}.

\subsection{RS generating functional}
\label{appsubeq: rs generating functional}
As we have already seen, the generating functional can be written as \eqref{appeq: generating functional rs 1}-\eqref{appeq: effective w 2}.  Imposing that $\alpha_U^{(t)}=\alpha_U, \Delta_U^{(t)}=\Delta_U, \alpha_U^{(t)}=\alpha_U$ and $\rho^{(t)}_U=\rho_U$ yields the Prediction.

As discussed in the main text, $\{\wsf{t}\}$ in \eqref{appeq: effective w 1}-\eqref{appeq: effective w 2} effectively describes the statistical properties of the weight vectors. Although the weight vectors $\hat{\bm{w}}^{(t)}$ are the high-dimensional vectors in $\R^N$, the empirical distributions of the component of the vectors $\{\hat{w}_i^{(t)}\}$ are described by the one-dimensional Gaussian process. Similarly, we can consider another generating functional regarding $y_\nu^{(t)}$, $\tilde{u}_{\nu}=\bm{x}_{\nu}^{(t)}\cdot\bm{w}^{(t-1)}/\sqrt{N} + B^{(t-1)}$ and ${u}_{\nu}=\bm{x}_{\nu}^{(t)}\cdot\bm{w}^{(t)}/\sqrt{N}+B^{(t)}$:
\begin{equation}
    \Xi_{\rm ST}(\epsilon_u) = \lim_{N,\beta\to\infty}  \E_{\{\bm{\theta}^{(t)}\}_{t=0}^T\sim p_{\rm ST}, D}\left[
        e^{
            \epsilon_u g_u(\{(y_\nu^{(t)}, \tilde{u}_{\nu}^{(t)}, u_{\nu}^{(t)})\}_{t=1}^{T})
        }
    \right], 
\end{equation}
where $\nu \in [M_U]$ and $t\ge1$. The computation of this generating functional is completely analogous to those in the case of $\Xi_{\rm ST}(\epsilon_w, \epsilon_B)$. Specifically, the same form of the replica trick can be used by replacing the factor $e^{\epsilon_w g_w(\{w_{1,i}^{(t)}\}_{t=0}^{T})+ \epsilon_Bg_B(\{B_1^{(t)}\}_{t=0}^{T})} $ by $e^{\epsilon_u g_u(\{(y_{\nu}^{(t)}, \tilde{u}_{1,\nu}^{(t)}, u_{1,\nu}^{(t)})\}_{t=1}^{T})}$. Again, the index $1$ in the latter expression is the replica index. The following calculation procedures are also the same. Then, the counterpart to equation \eqref{appeq: generating functional general} is obtained as follows:
\begin{align}
    \phi_{n_0,\dots,n_T}^{(T)} &= \lim_{N, \beta\to\infty}\int 
        e^{N\gS_u(\Theta,\hat{\Theta})}
        \E_{\{y^{(t)}, \bm{u}^{(t)}\}_{t=1}^T}\left[
            e^{
                \epsilon_u g_u(\{(y^{(t)}, \tilde{u}_1^{(t)}, u_1^{(t)})\}_{t=1}^{T})
            }
        \right]
    d\Theta d\hat{\Theta},
    \label{appeq: generating functional general u}
    \\
    y^{(t)} &\sim p_y^{(t)}, \bm{u}^{(t)}\sim \tilde{p}_{{\rm eff}, u|y},
    \\
    \gS_u(\Theta, \hat{\Theta}) &= \alpha_L \varphi_u^{(0)} + \sum_{t=1}^T(1-1/N)\alpha_U^{(t)} \varphi_u^{(t)} + \varphi_w.
\end{align}
Recall that $\tilde{p}_{{\rm eff}, u}$ is the density defined in \eqref{appeq: effective density u integer n}. Also, $\gS_u$ is equal to $\gS$ when $N\gg1$. The integral over $\Theta, \hat{\Theta}$ can be approximated by the saddle point at $N\gg 1$, which yields the same saddle point condition. After that, one can obtain the following result:
\begin{align}
    \Xi_{\rm ST}(\tilde{\epsilon}_u, \epsilon_u) &= \E_{\xi_u^{(0)}, \xi_{u,1}^{(t)}, \xi_{u,2}^{(t)}\sim_{\rm iid}\gN(0,\Delta_U^{(t)}), y^{(t)}\sim p_y^{(t)}}[
        e^{
            \tilde{\epsilon}_u g_{\tilde{u}}(\{\tilde{h}_u^{(t)}\}_{t=1}^T)
            + \epsilon_u g_u(\{h_u^{(t)} + \usf{t}\}_{t=0}^{T})
        }
    ].
\end{align}
By a similar arguments with $\Xi_{\rm ST}(\epsilon_w, \epsilon_B)$, it can be understood that $(\tilde{h}_u^{(t)}, \usf{t}+h_u^{(t)})$ effectively describes the statistical properties of the logits. It effectively describes the statistical properties of the empirical distributions of the logits. The minimization problems \eqref{appeq: effective problem u 0}-\eqref{appeq: effective problem u t} can be understood as effective one-dimensional problems to determine the logits. There, $\xi_{u,1}^{(t)}$ and $\xi_{u,2}^{(t)}$ effectively play the role of $D_U^{(t)}$.

%
\begin{figure}[p!]
    \centering
    \begin{subfigure}[b]{0.322\textwidth}
        \centering
        \includegraphics[width=\textwidth]{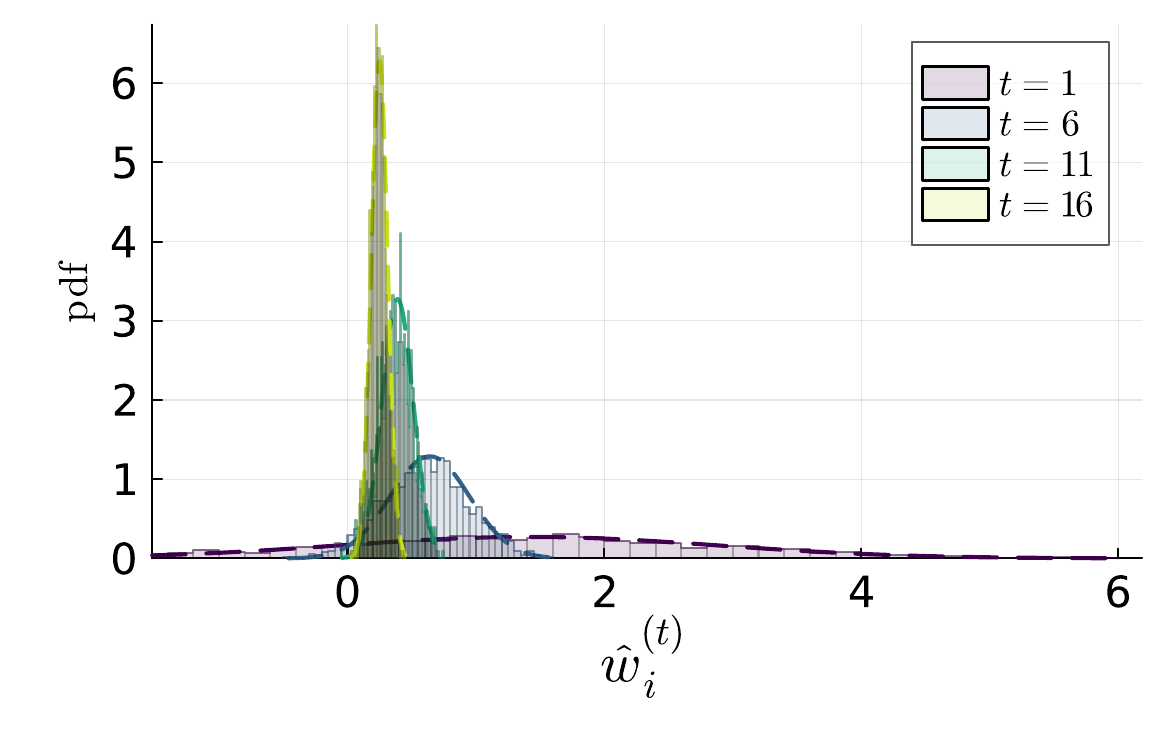}
        \caption{$\{\hat{w}_i^{(t)}\}: (N,\rho)=(2^{10}, 0.5)$}
    \end{subfigure}
    \hfill
    \begin{subfigure}[b]{0.322\textwidth}
        \centering
        \includegraphics[width=\textwidth]{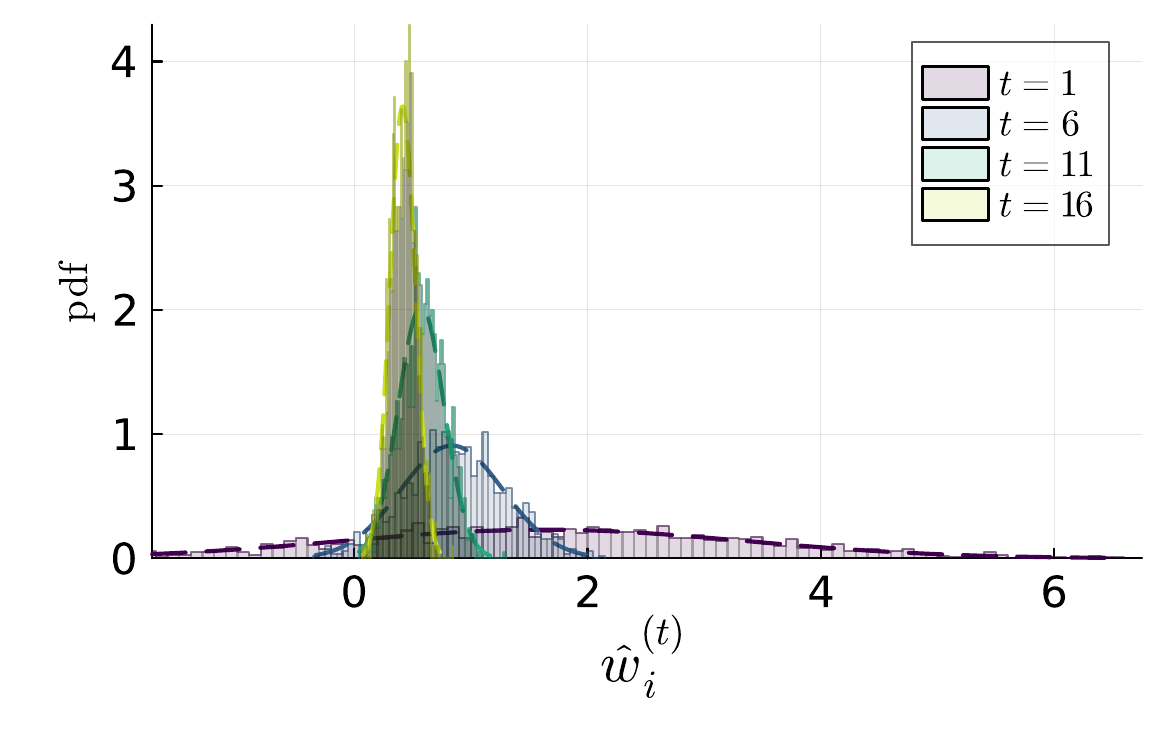}
        \caption{$\{\hat{w}_i^{(t)}\}: (N,\rho)=(2^{10}, 0.495)$}
    \end{subfigure}
    \hfill
    \begin{subfigure}[b]{0.322\textwidth}
        \centering
        \includegraphics[width=\textwidth]{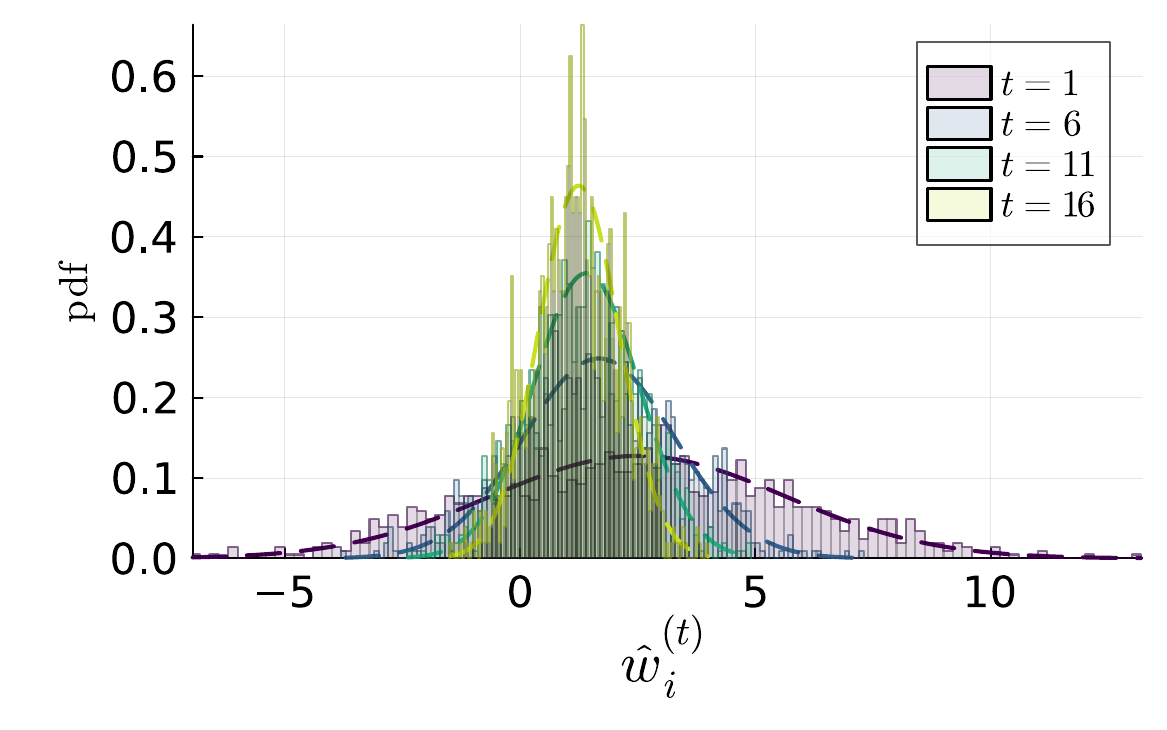}
        \caption{$\{\hat{w}_i^{(t)}\}: (N,\rho)=(2^{10}, 0.495)$}
    \end{subfigure}
    \begin{subfigure}[b]{0.322\textwidth}
        \centering
        \includegraphics[width=\textwidth]{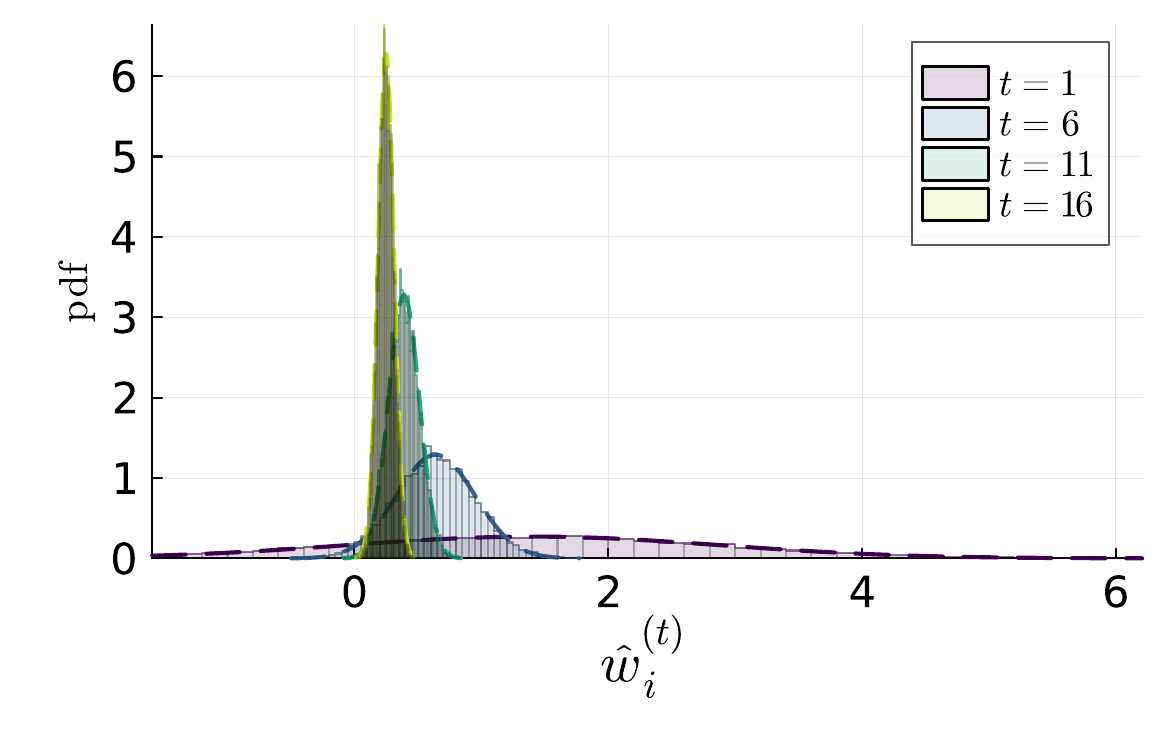}
        \caption{$\{\hat{w}_i^{(t)}\}: (N,\rho)=(2^{13}, 0.5)$}
    \end{subfigure}
    \hfill
    \begin{subfigure}[b]{0.322\textwidth}
        \centering
        \includegraphics[width=\textwidth]{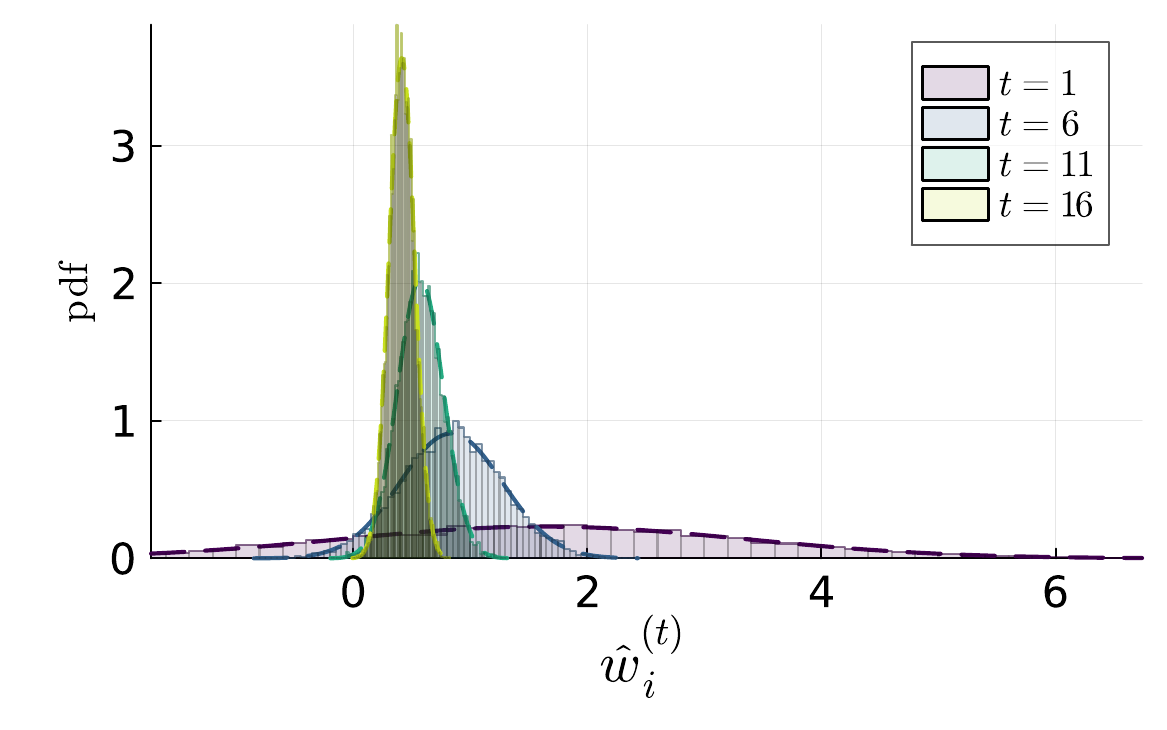}
        \caption{$\{\hat{w}_i^{(t)}\}: (N,\rho)=(2^{13}, 0.495)$}
    \end{subfigure}
    \hfill
    \begin{subfigure}[b]{0.322\textwidth}
        \centering
        \includegraphics[width=\textwidth]{figures/experiment_result_PLS_rho0.4_delta0.5625_alphaL0.5_alphaU2.0_T16_N8192_wdist.pdf}
        \caption{$\{\hat{w}_i^{(t)}\}: (N,\rho)=(2^{13}, 0.4)$}
    \end{subfigure}
    %
    \begin{subfigure}[b]{0.322\textwidth}
        \centering
        \includegraphics[width=\textwidth]{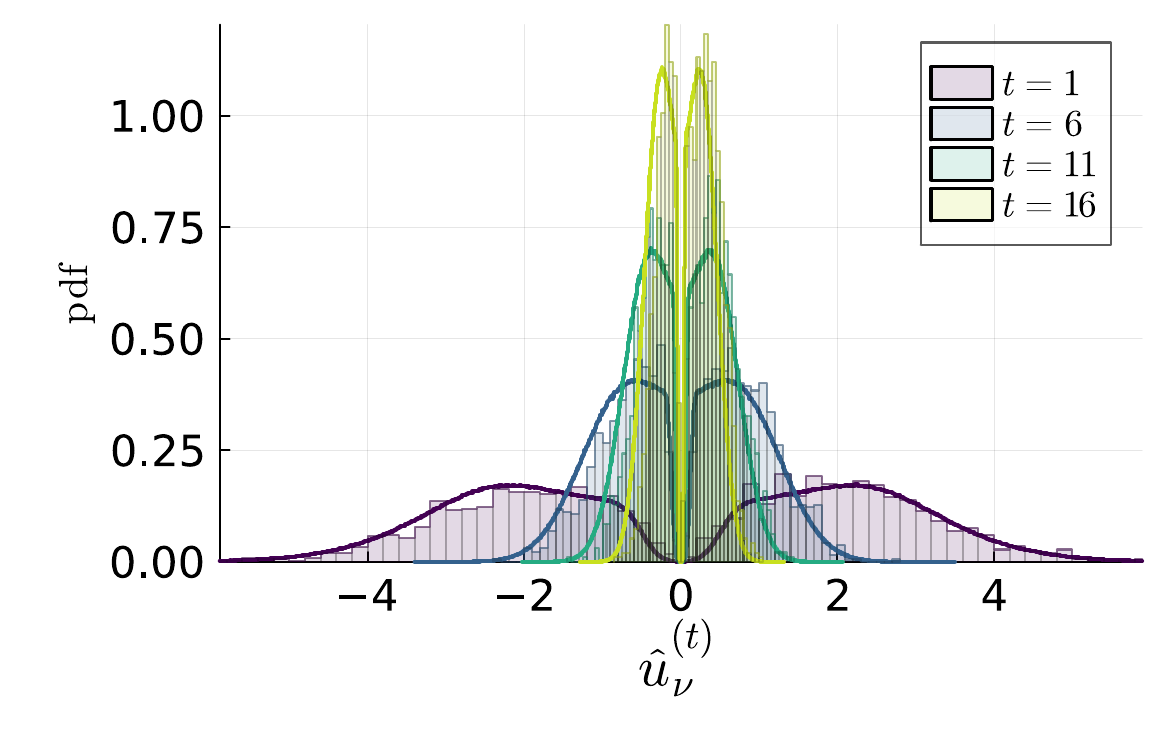}
        \caption{$\{\hat{u}_\nu^{(t)}\}: (N,\rho)=(2^{10}, 0.5)$}
    \end{subfigure}
    \hfill
    \begin{subfigure}[b]{0.322\textwidth}
        \centering
        \includegraphics[width=\textwidth]{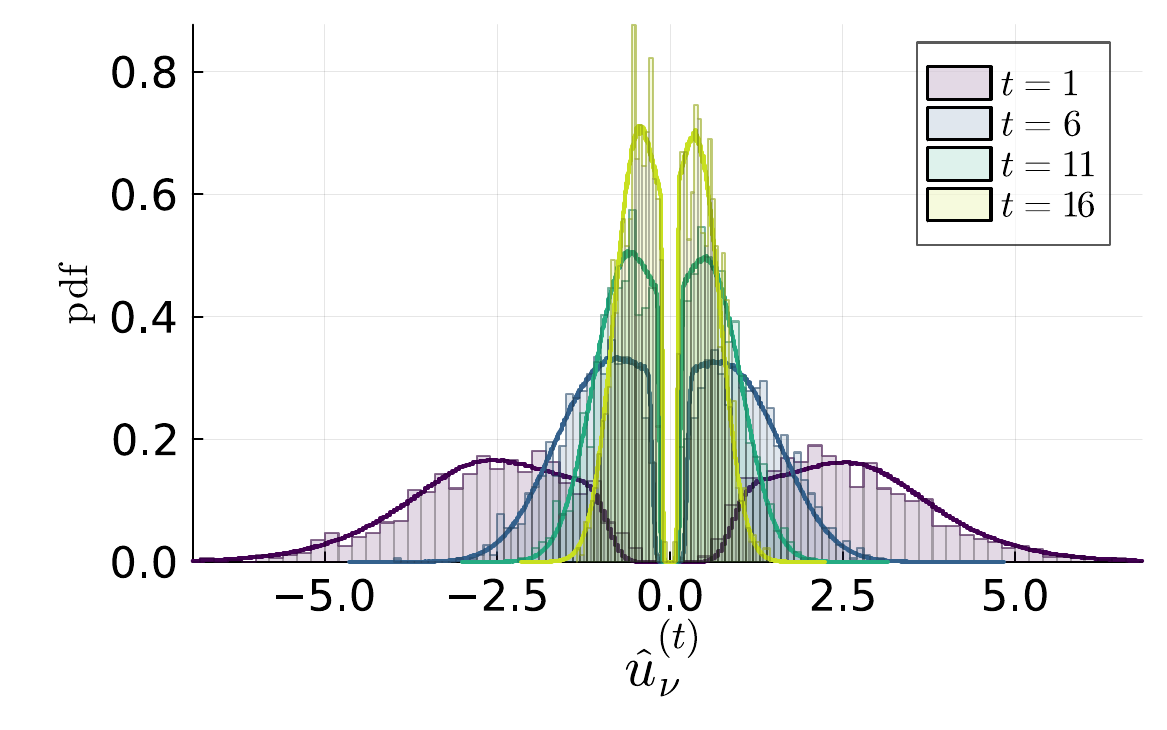}
        \caption{$\{\hat{u}_\nu^{(t)}\}: (N,\rho)=(2^{10}, 0.495)$}
    \end{subfigure}
    \hfill
    \begin{subfigure}[b]{0.322\textwidth}
        \centering
        \includegraphics[width=\textwidth]{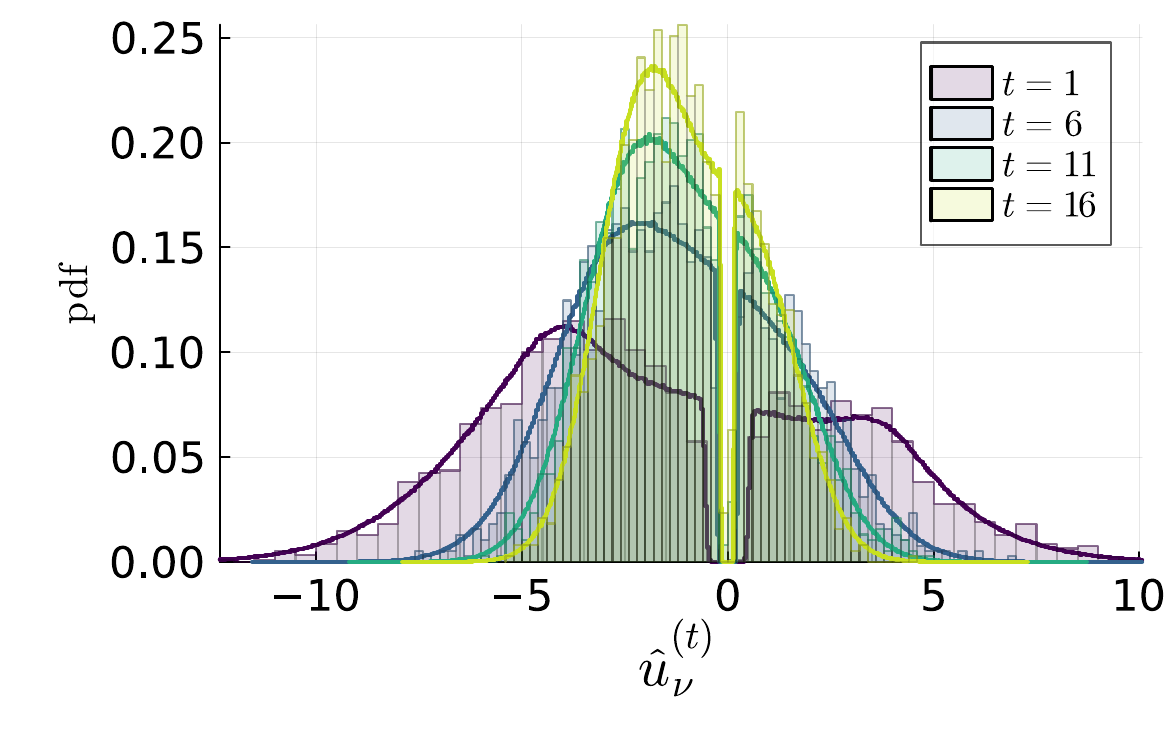}
        \caption{$\{\hat{u}_\nu^{(t)}\}: (N,\rho)=(2^{10}, 0.4)$}
    \end{subfigure}
    \begin{subfigure}[b]{0.322\textwidth}
        \centering
        \includegraphics[width=\textwidth]{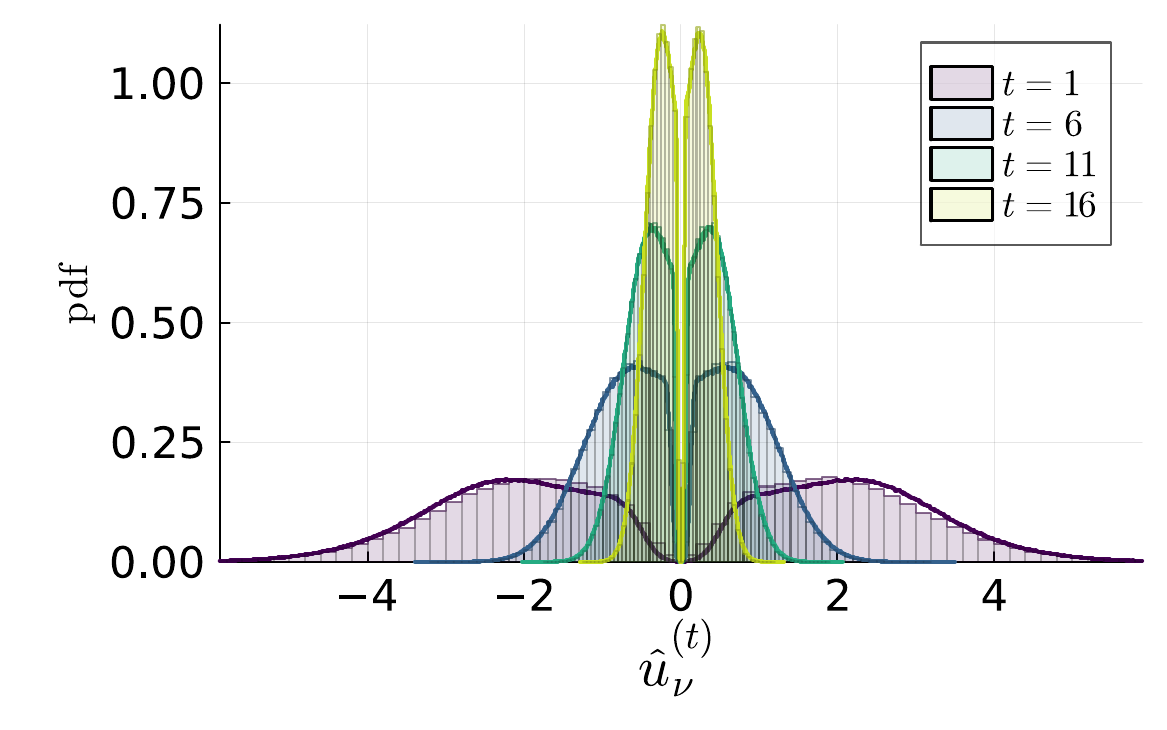}
        \caption{$\{\hat{u}_\nu^{(t)}\}: (N,\rho)=(2^{13}, 0.5)$}
    \end{subfigure}
    \hfill
    \begin{subfigure}[b]{0.322\textwidth}
        \centering
        \includegraphics[width=\textwidth]{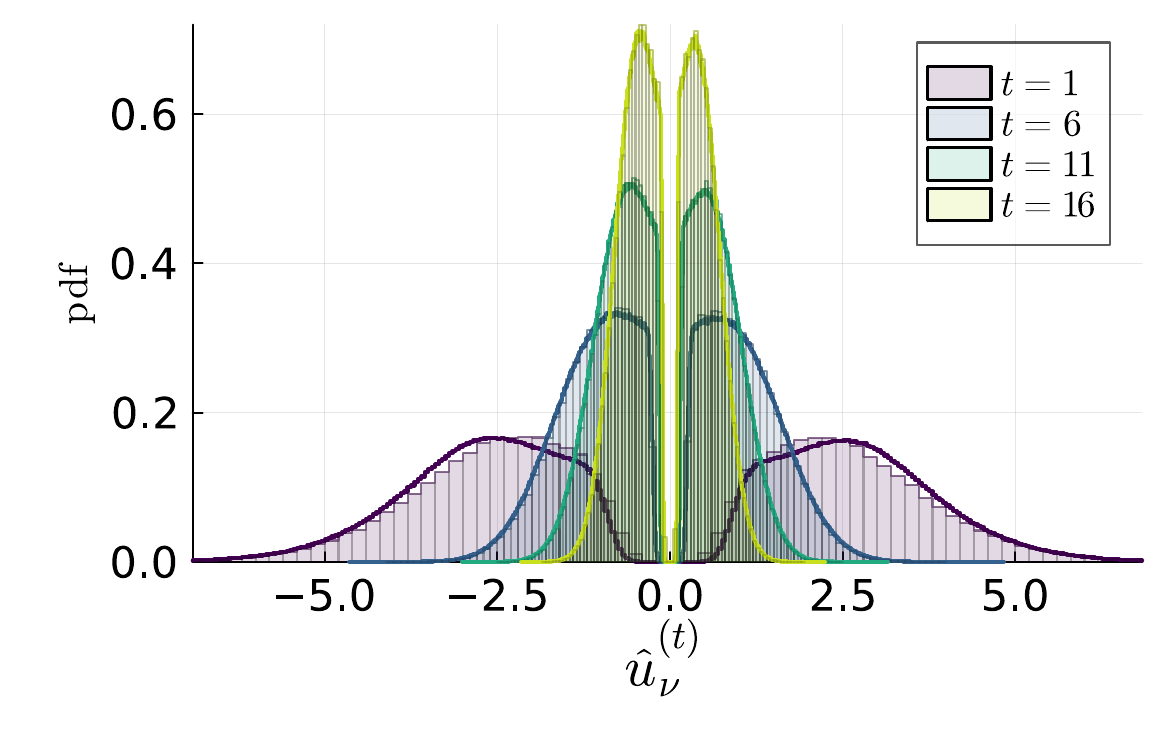}
        \caption{$\{\hat{u}_\nu^{(t)}\}: (N,\rho)=(2^{13}, 0.495)$}
    \end{subfigure}
    \hfill
    \begin{subfigure}[b]{0.322\textwidth}
        \centering
        \includegraphics[width=\textwidth]{figures/experiment_result_PLS_rho0.4_delta0.5625_alphaL0.5_alphaU2.0_T16_N8192_udist.pdf}
        \caption{$\{\hat{u}_\nu^{(t)}\}: (N,\rho)=(2^{13}, 0.4)$}
    \end{subfigure}
    \caption{
        (a)-(f): Comparison between the empirical distribution of the elements of $\hat{\bm{w}}^{(t)}$ (histogram), which is obtained by single-shot numerical experiment of finite size, and the theoretical prediction given by the Gaussian process $\wsf t$ in \eqref{eq: GP of w} (solid line). 
        (g)-(l): Comparison between the empirical distribution of the elements of $\{\hat{u}_\nu^{(t)}=\hat{\bm{w}}^{(t)}\cdot \bm{x}_\nu^{(t)}/\sqrt{N}+\hat{B}^{(t)}\}_{\nu=1}^{M_U}$ (histogram), which is obtained by single-shot numerical experiment of finite size, and the theoretical prediction given by $\usf{t} + h_u^{(t)}$ in \eqref{eq:rs-saddle-uhat} and \eqref{eq: intuitive hu2} (solid line). 
        Different colors represent different iteration steps of ST with total number of iterations $T=16$. For the details of the settings, refer to the main text.
    }
    \label{fig: empirical dist with experiment}
\end{figure}

\begin{figure}[p!]
    \centering
    \begin{subfigure}[b]{0.322\textwidth}
        \centering
        \includegraphics[width=\textwidth]{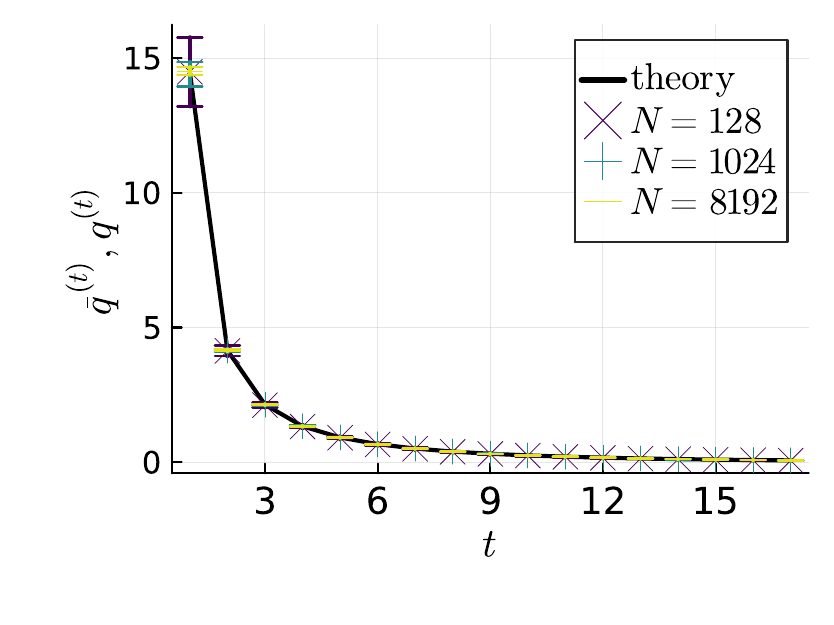}
        \caption{$q^{(t)}: \rho=0.5$}
    \end{subfigure}
    \hfill
    \begin{subfigure}[b]{0.322\textwidth}
        \centering
        \includegraphics[width=\textwidth]{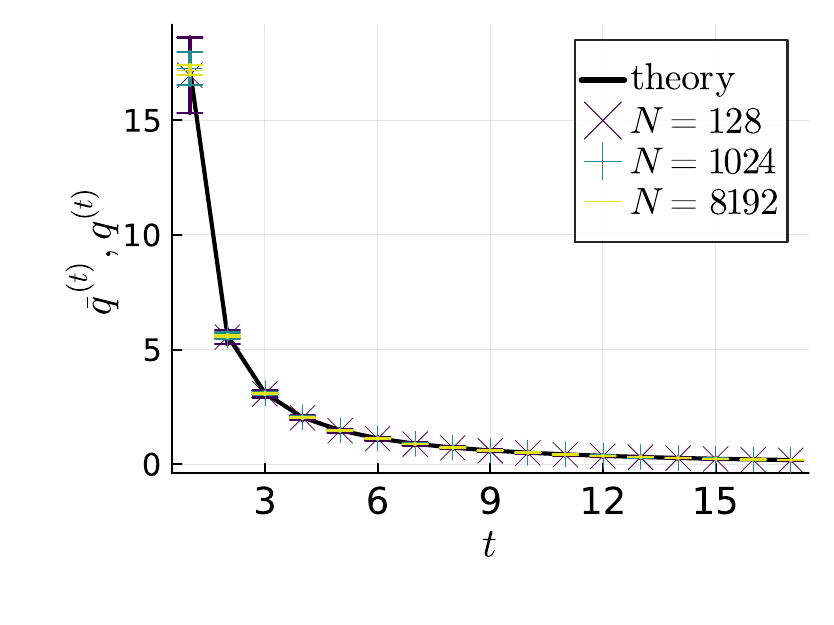}
        \caption{$q^{(t)}: \rho=0.495$}
    \end{subfigure}
    \hfill
    \begin{subfigure}[b]{0.322\textwidth}
        \centering
        \includegraphics[width=\textwidth]{figures/experiment_result_PLS_rho0.4_delta0.5625_alphaL0.5_alphaU2.0_T16_q.pdf}
        \caption{$q^{(t)}: \rho=0.4$}
    \end{subfigure}
    %
    \begin{subfigure}[b]{0.322\textwidth}
        \centering
        \includegraphics[width=\textwidth]{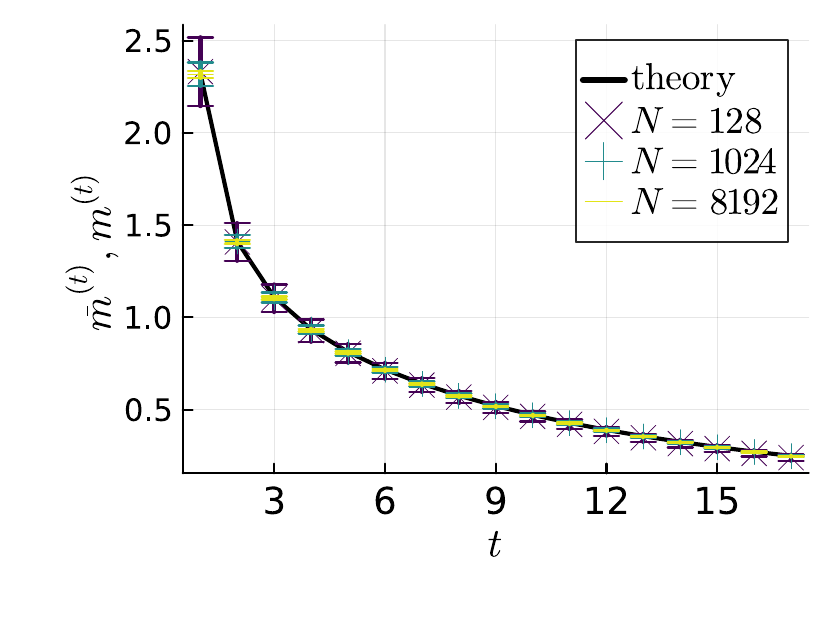}
        \caption{$m^{(t)}: \rho=0.5$}
    \end{subfigure}
    \hfill
    \begin{subfigure}[b]{0.322\textwidth}
        \centering
        \includegraphics[width=\textwidth]{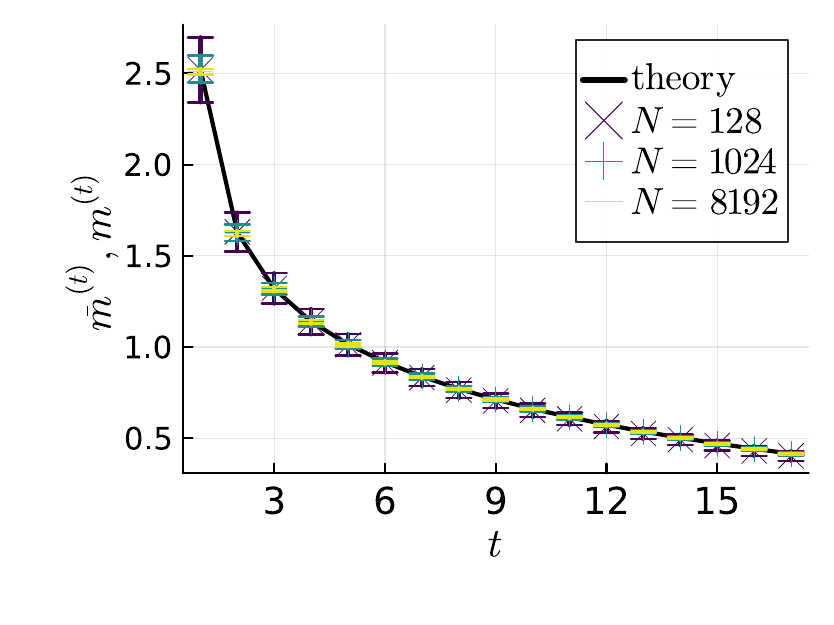}
        \caption{$m^{(t)}: \rho=0.495$}
    \end{subfigure}
    \hfill
    \begin{subfigure}[b]{0.322\textwidth}
        \centering
        \includegraphics[width=\textwidth]{figures/experiment_result_PLS_rho0.4_delta0.5625_alphaL0.5_alphaU2.0_T16_m.pdf}
        \caption{$m^{(t)}: \rho=0.4$}
    \end{subfigure}
    %
    \begin{subfigure}[b]{0.322\textwidth}
        \centering
        \includegraphics[width=\textwidth]{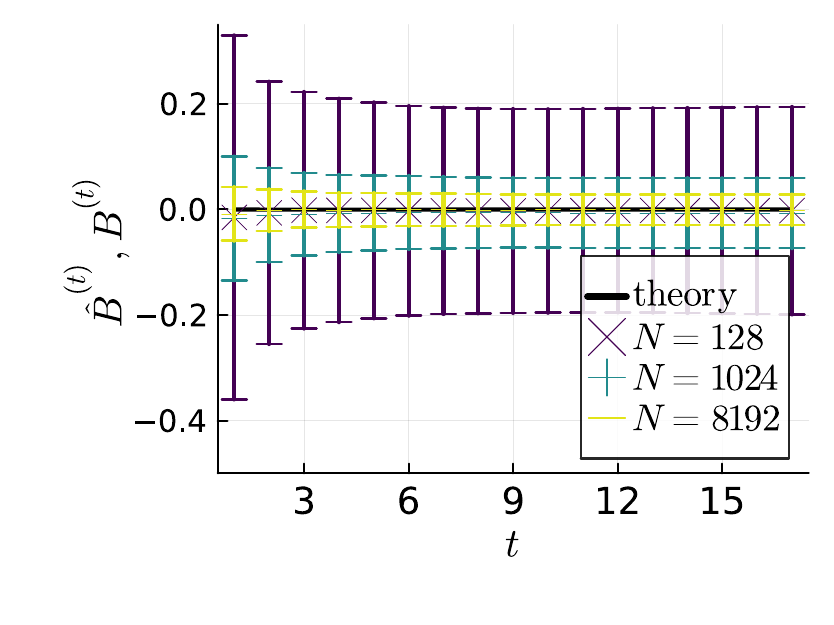}
        \caption{$B^{(t)}: \rho=0.5$}
    \end{subfigure}
    \hfill
    \begin{subfigure}[b]{0.322\textwidth}
        \centering
        \includegraphics[width=\textwidth]{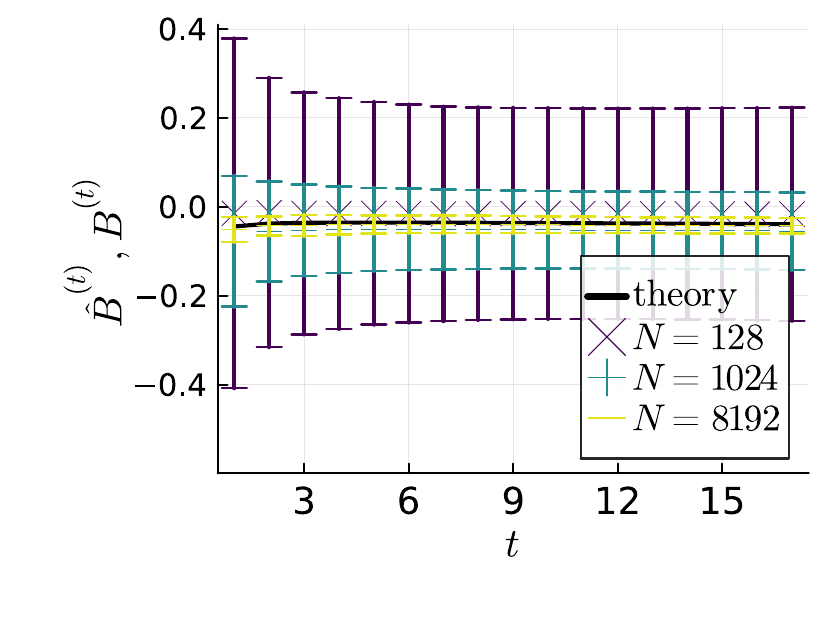}
        \caption{$B^{(t)}: \rho=0.495$}
    \end{subfigure}
    \hfill
    \begin{subfigure}[b]{0.322\textwidth}
        \centering
        \includegraphics[width=\textwidth]{figures/experiment_result_PLS_rho0.4_delta0.5625_alphaL0.5_alphaU2.0_T16_B.pdf}
        \caption{$B^{(t)}: \rho=0.4$}
    \end{subfigure}
    %
    \begin{subfigure}[b]{0.322\textwidth}
        \centering
        \includegraphics[width=\textwidth]{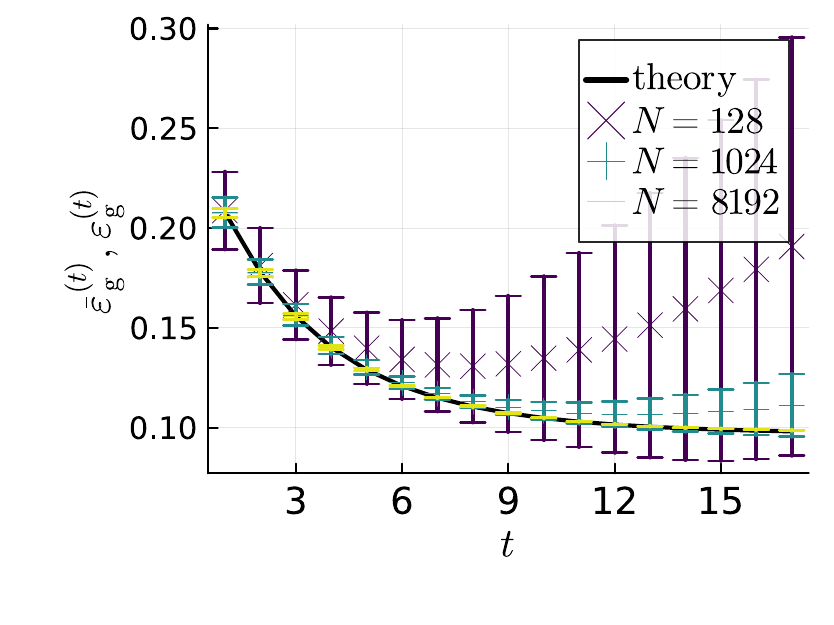}
        \caption{$\epsilon_{\rm g}^{(t)}: \rho=0.5$}
    \end{subfigure}
    \hfill
    \begin{subfigure}[b]{0.322\textwidth}
        \centering
        \includegraphics[width=\textwidth]{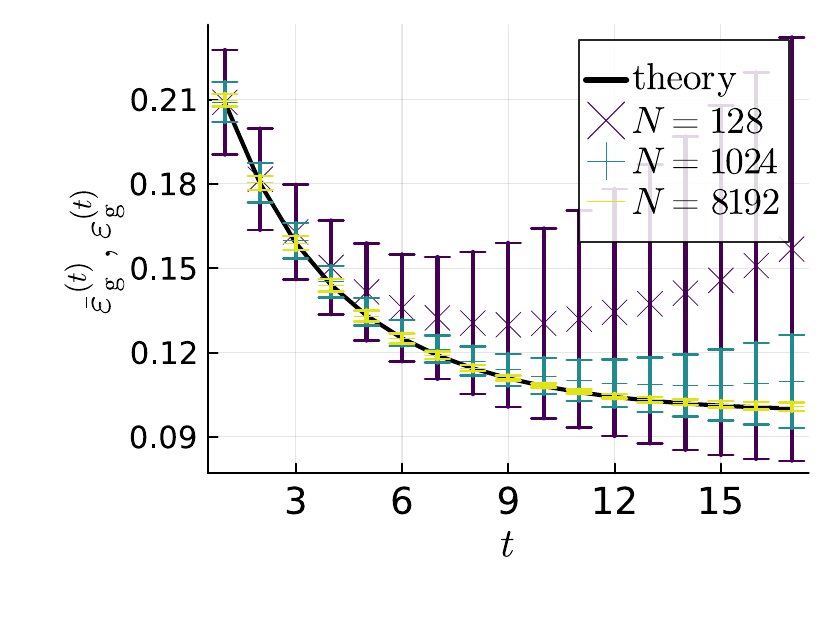}
        \caption{$\epsilon_{\rm g}^{(t)}: \rho=0.495$}
    \end{subfigure}
    \hfill
    \begin{subfigure}[b]{0.322\textwidth}
        \centering
        \includegraphics[width=\textwidth]{figures/experiment_result_PLS_rho0.4_delta0.5625_alphaL0.5_alphaU2.0_T16_egen.pdf}
        \caption{$\epsilon_{\rm g}^{(t)}: \rho=0.4$}
    \end{subfigure}
    \caption{
        Comparison of the macroscopic quantities obtained by experiments of finite-size systems (markers with error bars) and the theoretical prediction (black solid line).  The error bars represent standard deviations.  
        (a)-(c): The squared norm of the weight vector: $\|\hat{\bm{w}}^{(t)}\|_2^2/N$ and  $q^{(t)}$ in \eqref{eq: meaning of q}.
        (d)-(f): The inner product between the cluster center and the weight vector: $\bm{v}\cdot \hat{\bm{w}}^{(t)}/N$ and $m^{(t)}$ in \eqref{eq: meaning of m}.
        (g)-(i): The bias: $\hat{B}^{(t)}$ and $B^{(t)}$ in \eqref{eq: meaning of B}.
        (j)-(l): Generalization error: $\epsilon_{\rm g}^{(t)}$ defined in \eqref{eq: gen_err} and that in \eqref{eq: rs generalization error}.  
    }
    \label{fig: macro with experiment}
\end{figure}

\section{Additional finite size experiments}
\label{app: additional finite size checks}

\inred{
In this appendix, we provide additional comparisons between the RS predictions and finite-size numerical experiments. The main text shows representative examples in
Figures~\ref{fig: representative empirical dist with experiment} and \ref{fig: representative macro with experiment}. Here, we report the remaining plots for other values of the label bias $\rho$, system size $N$, and macroscopic quantities. These additional results support the same conclusion as in the main text: the RS predictions accurately capture both the empirical distributions and the macroscopic order parameters in the finite-size simulations.
}

\paragraph{Distribution of the weights and logits:} The panels (a)-(f) in Figure \ref{fig: empirical dist with experiment} show the comparison between the empirical distribution of the elements of the weight vector $\{\hat{w}_i^{(t)}\}_{i=1}^N$, obtained by the numerical experiments of finite size systems, and the theoretical prediction given by the Gaussian process $\wsf{t}$ in \eqref{eq: GP of w}. Each panel shows the result of different label bias $\rho$ and the system size $N$. Also, the panels (g)-(l) in Figure \ref{fig: empirical dist with experiment} show the comparison between the empirical distribution of the logits $\{\hat{u}_\nu^{(t)}=\hat{\bm{w}}^{(t)}\cdot\bm{x}_\nu/\sqrt{N} +\hat{B}^{(t)}\}$ and the theoretical prediction given by $\usf{t} + h_u^{(t)}$ in \eqref{eq:rs-saddle-uhat} and \eqref{eq: intuitive hu2}. Since the logits on the data points excluded by PLS are not detemined, the contribution from these points is excluded from the figures. For numerical experiments, two different system sizes $N=1024(=2^{10})$ and $8192(=2^{13})$ are used. The figure shows that the empirical distributions of both the weight vector and the logits at each iteration step are in good agreement even at $N=1024$. Although the results of $N=1024$ have a slight fluctuation due to the lack of sample size, the results of $N=8192$ show almost complete agreement with the RS solution.

\paragraph{Macroscopic quantities:} Figure \ref{fig: macro with experiment} presents the results of a comparison between experiment of finite size systems and the theoretical predictions for macroscopic quantities such as the norm $\|\hat{\bm{w}}^{(t)}\|_2^2/N$, the inner product with the cluster center $\bm{v}\cdot\hat{\bm{w}}^{(t)}/N$, the bias $\hat{B}^{(t)}$, and the generalization error \eqref{eq: gen_err}. Each panel shows the result of different label bias $\rho$. For numerical experiments, three different system sizes $N=128, 1024$, and $8192$ are used. Reported experimental results are averaged over several realization of $D$ depending on the size $N$. The error bars represent standard deviations. The panels of $q^{(t)}, m^{(t)}$ and $B^{(t)}$ show that the experimental results and RS solutions are in good agreement even at $N=1024$. Also, the standard deviation decreases as the system size grows, indicating the concentration of these quantities. However, in the panels of generalization error, there is still a discrepancy between experiment and theory at $N=1024$. This is because $q^{(t)}$ becomes very small when $t$ is large, causing even slight fluctuations of $\bar{q}^{(t)}$ and $\hat{B}^{(t)}$ to lead to significant deviations in the generalization error that depends on these quantities in the form of $\hat{B}^{(t)}/\sqrt{\bar{q}^{(t)}}$ as in \eqref{eq: macro generalization error}. Hence a large system size such as $N=8192$ is required to obtain a good agreement in the generalization error.

\section{Increase of the linear suceptiblity in small unlabeled data scenario}
\label{app: linear susceptibility}
\begin{figure}[t!]
    \centering
    \includegraphics[width=0.9\textwidth]{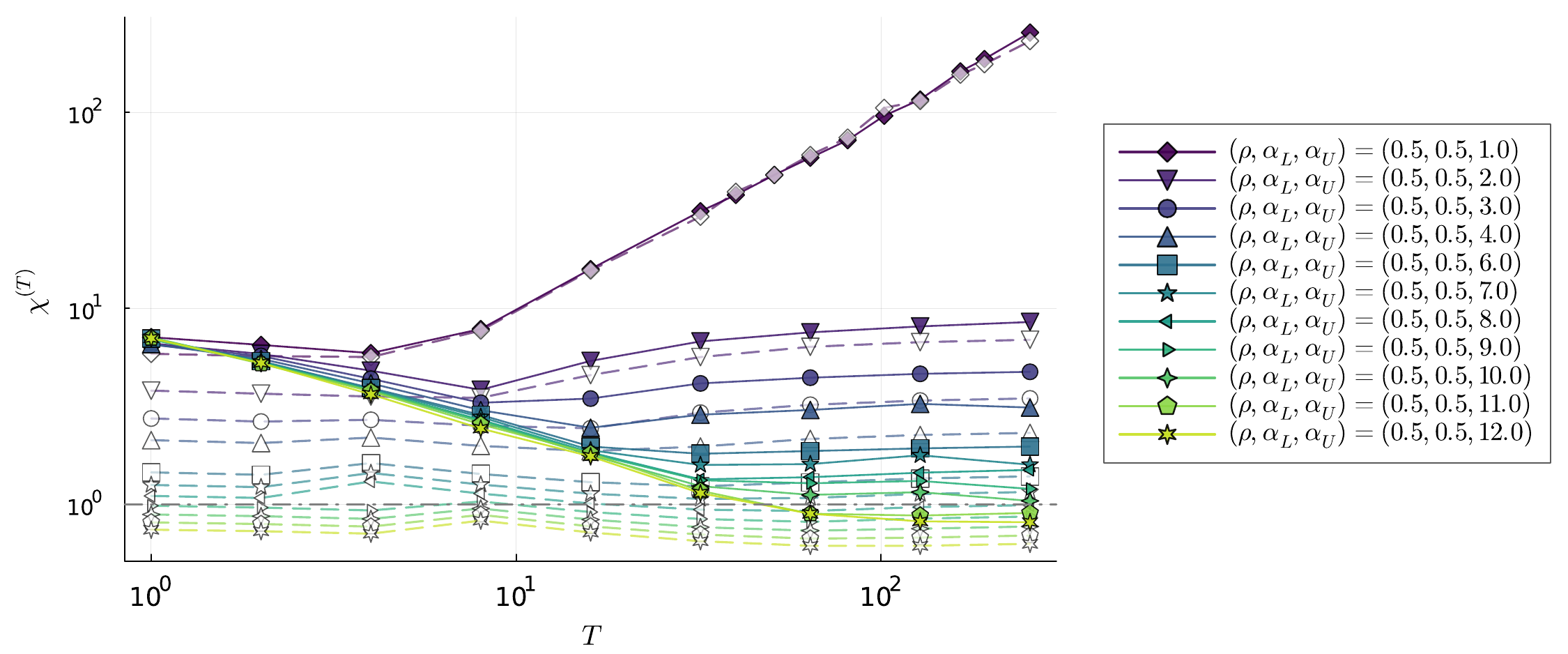}
    \caption{The behavior of the quantity $\chi^{(T)}$ when $(\rho,\Delta)=(1/2,0.75^2)$. Different colors represent different $\alpha_U$.}
    \label{fig: partition dependence balanced susceptibility}
\end{figure}

\inred{
    In this appendix, we complement the observations in subsection \ref{subsec: numerical inspection}, which shows that the ST does not perform better when $\alpha_U$ is small. Figure \ref{fig: partition dependence balanced susceptibility} shows that as $\alpha_U$ decreases, the linear susceptibility $\chi^{(T)}$ becomes greater at large iteration $T$, which indicates that the training result is getting unstable towers a small perturbation to the loss function. These observations indicate that ST is unstable when it repeatedly fits to pseudo-labels in overparameterized settings. 
}

\section{Derivation of Prediction \ref{pred: smoothing}}
\label{app: derivation of smoothing}
In this section, we outline the derivation of Prediction \ref{pred: differential equation}.  Also, let $\lambda_U^{(t)}, \alpha_U^{(t)}, \rho_U^{(t)}$ and $\Delta_U^{(t)}$ be the values of the regularization parameter, the relative size of the unlabeled data, and the label imbalance and the size of the cluster at step $t\ge1$. Although we mainly focus on the case of $\lambda_U^{(t)}=\lambda_U, \alpha_U^{(t)}=\alpha_U, \rho_U^{(t)}=\rho_U$ and $\Delta_U^{(t)}=\Delta_U$ in the main text paper, the following derivation allows this step-dependent cases. 

The basic strategy is to expand $\Theta^{(t)}$ and $\hat{\Theta}$ as $\Theta^{(t)}=\Theta_0^{(t)} + \Theta_1^{(t)}\lambda_U^{(t)}+\dots$, and $ \hat{\Theta}^{(t)}=\hat{\Theta}_0^{(t)} + \hat{\Theta}_1^{(t)}\lambda_U^{(t)} + \dots$, respectively,  and then to consider the self-consistent equations order by order in $\lambda_U$. In the following, lower subscript represents the order of the expansion. For example, $q_0^{(t)}$ is the zero-th order term of $q^{(t)}$.

Before going into the details, we make two remarks here. The first point is regarding the sign of $\chi^{(t)}$. Since $\chi^{(t)}$ is defined as a limit of the variance as in \eqref{appeq: chi t}, it is positive:
\begin{equation}
    \chi^{(t)} > 0.
    \label{appeq: chi t is positive}
\end{equation}
The second point is about $\usf{t}$. Due to the convexity of the loss functions, $\usf{t}, t\ge1$ in \eqref{eq:rs-saddle-uhat} is determined by the following extreme condition:
\begin{equation}
    \frac{\usf{t}}{\chi^{(t)}\Delta_U^{(t)}} + \partial_2 \tilde{l}(\tilde{h}_{u}^{(t)},h_u^{(t)} + \usf{t}) = 0.
    \label{appeq: extreme u}
\end{equation}

\subsection{Zero-th order}
First, Prediction \ref{pred: w effective average} and Assumption \ref{assumption: stability} immediately imply the following:
\begin{equation}
    q_0^{(t)} = q^{(t-1)}, 
    \\
    R_0^{(t)} = q^{(t-1)},
    \\
    m_0^{(t)} = m^{(t-1)}.
    \label{appeq: stability of order parameter zeroth order}
\end{equation}
Inserting these conditions into the expression of $h_{u}^{(t)}$, it is shown that the lowest order of $h_u^{(t)}$ is equal to $\tilde{h}_{u}^{(t)}$:
\begin{equation}
    h_{u,0}^{(t)} = \tilde{h}_{u}^{(t)}.
    \label{appeq: h_u 0}
\end{equation}
Then, the zero-th order of the condition \eqref{appeq: extreme u} for determining $\usf{t}$ is given as follows
\begin{equation}
	\frac{\usf{t}_0}{\chi_0^{(t)}\Delta_U^{(t)}} + \partial_2 \tilde{l}(\tilde{h}_{u,t}, \tilde{h}_{u,t} + \usf{t}_0) = 0.
\end{equation}
From \eqref{eq: at extremum} in Assumption \ref{assumption: loss}, this equation is satisfied trivially when $\usf{t}_0 = 0$. Thus, it is concluded that the zero-th order of $\usf{t}_0$ vanishes:
\begin{equation}
    \usf{t}_0 = 0.
    \label{appeq: u0}
\end{equation}
By inserting this condition into the self-consistent equations \eqref{eq:rs-saddle-hatm} and \eqref{eq:rs-saddle-hatchi}, it is shown that the zero-th order terms of $\hat{m}^{(t)}$ and $\hat{\chi}^{(t)}$ also vanish:
\begin{align}
   \hat{m}_0^{(t)} &= 0,
    \label{appeq: mhat0}
    \\
    \hat{\chi}_0^{(t)} &= 0.
    \label{appeq: chihat0}
\end{align}

On the other hand, the zero-th order of the self-consistent equations for $q^{(t)}, m^{(t)}$ and $R^{(t)}$ are given as follows:
\begin{align}
	q_0^{(t)} &= \left(\frac{\hat{R}_0^{(t)}}{\hat{Q}_0^{(t)}}\right)^2q^{(t-1)},
	\\
	m_0^{(t)} &= \frac{\hat{R}_0^{(t)}}{\hat{Q}_0^{(t)}}m^{(t-1)},
	\\
	R_0^{(t)} &=  \frac{\hat{R}_0^{(t)}}{\hat{Q}_0^{(t)}}q^{(t-1)}.
\end{align}
Because the condition \eqref{appeq: stability of order parameter zeroth order} should be satisfied self-consistently, the zero-th order of $\hat{Q}^{(t)}$ and $\hat{R}^{(t)}$ are equal:
\begin{equation}	
	\hat{Q}_0^{(t)} = \hat{R}_0^{(t)}.
\end{equation}

Finally, the zero-th order expression of $\chi^{(t)}$ yields the condition
\begin{equation}
	\chi_0^{(t)} = \frac{1}{\hat{Q}_0^{(t)}}.
\end{equation}
Since $\chi_0^{(t)}$ is a value of $\chi^{(t)}$ at $\lambda_U=0$, it is non-negative from \eqref{appeq: chi t is positive}. Hence $\hat{Q}_0^{(t)}$ is also non-negative.

\subsection{First order}
First, we show that $h_u^{(t)}$ does not contain non-integer order terms at least up to the first order, i.e., it can be expanded as $h_u^{(t)} = h_{u,0}^{(t)} + h_{u,1}^{(t)}\lambda_U^{(t)} + \dots$. By inserting \eqref{appeq: mhat0} and \eqref{appeq: chihat0} into \eqref{eq:rs-saddle-q} and \eqref{eq:rs-saddle-r}, $q_1^{(t)}-2R_1^{(t)} =\hat{\chi}_1^{(t)}/(\hat{Q}_0^{(t)})^2$ is obtained. Since $h_u^{(t)}$ can be written as 
\begin{align}
    h_u^{(t)} &= \tilde{h}_u^{(t)} + \left(
        B_1^{(t)} + (2y^{(t)}-1)m_1^{(t)} + \frac{R_1^{(t)}}{\sqrt{q^{(t-1)}}}\xi_{u,1}^{(t)}
    \right)\lambda_U^{(t)}
    \nonumber \\
    &+ \sqrt{(q_1^{(t)} - 2R_1^{(t)})\lambda_U^{(t)} + \gO((\lambda_U^{(t)})^2)}\xi_{u,2}^{(t)} + \gO((\lambda_U^{(t)})^2),
    \label{appeq: h expansion general}
\end{align}
$\usf{t}$ that is determined by the condition \eqref{appeq: extreme u} may be expanded as 
\begin{equation}
	\usf{t} = \usf{t}_{1/2}(\lambda_U^{(t)})^{1/2} + \usf{t}_1\lambda_U + \dots.
\end{equation}
However, by inserting this expression and \eqref{appeq: h expansion general} into the condition \eqref{appeq: extreme u} and expanding by $\lambda_U$, one can show that $\usf{t}_{1/2}\propto\sqrt{\hat{\chi}_1^{(t)}}$. This implies that the RHS of \eqref{eq:rs-saddle-hatchi} is also proportional to $\hat{\chi}_1^{(t)}$ at the first order of $\lambda_U^{(t)}$, which leads to the condition $\hat{\chi}_1^{(t)}\propto\hat{\chi}_1^{(t)}$. Thus, we obtain  the following:
\begin{equation}
	\hat{\chi}_0^{(t)} = \hat{\chi}_1^{(t)} = 0.
	\label{appeq: chihat1}
\end{equation}
From this, it is evident that $h_u^{(t)}$ can be expanded as $h_u^{(t)} = h_{u,0}^{(t)} + h_{u,1}^{(t)}\lambda_U^{(t)} + \dots$ , where 
\begin{equation}
    h_{u,1}^{(t)} = B_1^{(t)} + (2y^{(t)} -1)m_1^{(t)} + \frac{R_1^{(t)}}{\sqrt{q^{(t-1)}}}\xi_{u,1}^{(t)} + \sqrt{q_2^{(t)} - \frac{(R_1^{(t)})^2}{q^{(t-1)}}-R_2^{(t)}}\xi_{u, 2}^{(t)}.
    \label{appeq: h_expansion first order}
\end{equation}
From the above result, it also follows that the lowest order of the expansion of $\usf{t}$ is the first order: $\usf{t}=\usf{t}_1\lambda_U^{(t)} + \dots$.

The consequence of the above result is that the contributions from the noise terms $\sqrt{\hat{\chi}^{(t)}}\xi_w^{(t)}$ in \eqref{eq: effective w intuitive} and $\sqrt{q^{(t)} - \frac{(R^{(t)})^2}{q^{(t-1)}}}\xi_{u,2}^{(t)}$ in \eqref{eq: intuitive hu2} are higher order in $\lambda_U^{(t)}$ because these terms appear only in the form of a square (second order in $\lambda_U^{(t)}$) or the raw average (equals to zero).


\section{Derivation of Prediction \ref{pred: differential equation}}
\label{app: derivation of differential equation}
The settings are the same with Appendix \ref{app: derivation of smoothing}. Let us start from the first order of the perturbative expansion considered in Appendix \ref{app: derivation of smoothing}.
Let us consider the self-consistent equations determining $q^{(t)}, R^{(t)}, m^{(t)}, B^{(t)}$. Inserting \eqref{appeq: mhat0} and \eqref{appeq: chihat1} into \eqref{eq:rs-saddle-q}-\eqref{eq:rs-saddle-r}, we obtain the first order expansion of $m^{(t)}, R^{(t)}$ and $q^{(t)}$ as follows:
\begin{align}
    m_1^{(t)} &= \frac{1}{\hat{Q}_0^{(t)}}\left(
        \hat{m}_1^{(t)} - (1- \hat{\delta}_1^{(t)})m^{(t-1)}
    \right),
    \label{appeq: m1}
    \\
    R_1^{(t)} &= \frac{1}{\hat{Q}_0^{(t)}}\left(
        \hat{m}_1^{(t)}m^{(t-1)} - (1- \hat{\delta}_1^{(t)})q^{(t-1)}
    \right),
    \label{appeq: r1}
    \\
    q_1^{(t)} &= 2R_1^{(t)},
    \label{appeq: q1}
    \\
    \hat{\delta}_1^{(t)} &= \hat{R}_1^{(t)} - \hat{Q}_1^{(t)},
    \label{appeq: deltahat1}
\end{align}
Importantly, the noise term 
\begin{equation}
    \sqrt{q_2^{(t)} - \frac{(R_1^{(t)})^2}{q^{(t-1)}}-R_2^{(t)}}\xi_{u, 2}^{(t)},
\end{equation}
in $h_{u}^{(t)}$ does not contribute to the above result at all because this term appears only in the form of a square (higher order in $\lambda_U^{(t)}$) or raw average (equals to zero). Therefore, combined with the result of $\hat{\chi}^{(t)}$ in \eqref{appeq: chihat1}, the update of the order parameters are intrinsically noiseless up to the first order of $\lambda_U^{(t)}$. 

By combining equations \eqref{appeq: m1}-\eqref{appeq: q1}, the next equations are obtained: 
\begin{align}
    M^{(t)} &= M^{(t-1)} + M_1^{(t)}\lambda_U^{(t)} + \dots,
    \\
    M_1^{(t)} &= \remark{\equiv C^{(t)}}{\frac{2}{\hat{Q}_0^{(t)}} \frac{\hat{m}_1^{(t)}}{m^{(t-1)}}}M^{(t-1)}(1-M^{(t-1)}).
\end{align}
Note that $\hat{Q}_0^{(t)} = 1/\chi_0^{(t)}>0$ as already described above. Since $\hat{m}_1^{(t)}$ is the leading order of $\hat{m}^{(t)}$ that represents the signal component of $\wsf{t}$ accumulated at step $t$ as depicted in \eqref{eq: effective w intuitive}, $C^{(t)}$ is positive if the training yields a positive accumulation of the signal component to $\wsf{t}$ when $m^{(t-1)}=\bm{v}\cdot \hat{\bm{w}}^{(t-1)}/N>0$, i.e., the classification plane at step $t-1$ is positively correlated with the cluster center, which seems to naturally hold when using a legitimate loss function. This is a natural result of the elimination of noise contributions as seen above.

We also remark that the condition $0=\E[\usf{t}]$ that determines $B^{(t)}$ is not used to derive the above result.

\section{Derivation of Prediction \ref{pred: squared loss}}
\label{app: derivation of differential equation squared loss}
In this section, we outline the derivation of Prediction \ref{pred: squared loss}. As in Appendix \ref{app: derivation of differential equation}, let $\lambda_U^{(t)}, \alpha_U^{(t)}, \rho_U^{(t)}$ and $\Delta_U^{(t)}$ be the values of the regularization parameter, the relative size of the unlabeled data, and the label imbalance and the size of the cluster at step $t\ge1$, although we mainly focus on the case of $\lambda_U^{(t)}=\lambda_U, \alpha_U^{(t)}=\alpha_U, \rho_U^{(t)}=\rho_U$ and $\Delta_U^{(t)}=\Delta_U$ in the main text. Dropping the time indices yield the Prediction \ref{pred: squared loss}.

When $\tilde{l}_{\tilde{\Gamma}}(y,x)=1/2(y-x)^2$, we can obtain an explicit expression of $\usf{t}$ as follows:
\begin{equation}
    \usf{t} = \frac{\tilde{h}_{u}^{(t)} - h_u^{(t)}}{\frac{1}{\chi^{(t)}\Delta_U^{(t)}}+1}.
\end{equation}
From this, the self-consistent equations that determine $\hat{Q}^{(t)}, \hat{R}^{(t)}, \hat{m}^{(t)}, B^{(t)}$ also have the following explicit expressions:
\begin{align}
    \hat{Q}^{(t)} &= \hat{R}^{(t)} = \frac{\alpha_U^{(t)}\Delta_U^{(t)}}{1 + \chi^{(t)}\Delta_U^{(t)}},
    \\
    \hat{m}^{(t)} &= \frac{\alpha_U^{(t)}\Delta_U^{(t)}}{1+\chi^{(t)}\Delta_U^{(t)}}\left((2\rho_U^{(t)}-1)(B^{(t-1)}-B^{(t)}) + (m^{(t-1)} - m^{(t)})\right),
    \\
    B^{(t)} &= B^{(t-1)} + (2\rho_U^{(t)} - 1) (m^{(t-1)}-m^{(t)}).
\end{align}
Since $\hat{Q}^{(t)}$ does not involve average, the expansion coefficient of $\hat{Q}^{(t)}$ and $\chi^{(t)}$ at the zero-th order can be explicitly solved as
\begin{align}
    \hat{Q}_0^{(t)} &= \Delta_U^{(t)}(\alpha_U^{(t)}-1),
    \\
    \chi_0^{(t)} &= \frac{1}{\Delta_U^{(t)}(\alpha_U^{(t)}-1)}.
\end{align}

By using the above explicit expressions, similar perturbative analysis made in Appendix \ref{app: derivation of differential equation} yields the expansion coefficients at the first order as follows:
\begin{align}
    m_1^{(t)} &= -\frac{1}{\Delta_U^{(t)} + V_U^{(t)}}\frac{1}{\alpha_U^{(t)}-1} m^{(t-1)},
    \\
    M_1^{(t)} &= C^{(t)}M^{(t-1)}(1-M^{(t-1)}), \quad C^{(t)} = \frac{2}{\alpha_U^{(t)}-1}\frac{V_U^{(t)}}{\Delta_U^{(t)}+V_U^{(t)}},
    \\
    B_1^{(t)} &= -(2\rho_U^{(t)}-1)m_1^{(t)}.
\end{align}
Solving these equations at the continuum limit yields Prediction \ref{pred: squared loss}. 

Since an explicit expression of $\hat{m}_1^{(t)}$ as $\hat{m}_1^{(t)} = V_U^{(t)}/(\Delta_U^{(t)} + V_U^{(t)})m^{(t-1)}$ is obtained, the time evolution of $q^{(t)}$ can also be derived, which indicates the exponential shrinkage of the norm directly. However, the resultant expression is rather complex.

\section{Numerical procedure for evaluating RS saddle point}
\label{app: numerical recipe}
In this section, we sketch the numerical treatment of the self-consistent equations in Definition \ref{def: self-consistent equations}. Recall that $\Theta^{(t)}$ and $\hat{\Theta}^{(t)}$ are defined as follows:
\begin{align}
    \Theta^{(0)} &= (q^{(0)}, \chi^{(0)}, m^{(0)}, B^{0}),
    \\
    \hat{\Theta}^{(0)} &= (\hat{Q}^{(0)}, \hat{\chi}^{(0)}, \hat{m}^{(0)}),
    \\
    \Theta^{(t)} &= (q^{(t)}, \chi^{(t)}, R^{(t)}, m^{(t)}, B^{t}),
    \\
    \hat{\Theta}^{(t)} &= (\hat{Q}^{(t)}, \hat{\chi}^{(t)}, \hat{R}^{(t)}, \hat{m}^{(t)}).
\end{align}

\begin{algorithm}[t]
    \caption{Fixed-point iteration of solving the self-consistent equations}
    \begin{algorithmic}[1]
    \label{algo: fixed-point iteration}
    \REQUIRE{
        The dumping parameter $\eta_{\rm d}\in(0,1)$, the convergence criterion $\epsilon_{\rm tol}$ and the maximum number of iterations $s_{\rm max}$.
    }
    \STATE{// solving for $t=0$}
        \STATE{Select initial value $\hat{\Theta}_0^{(0)}$ of $\hat{\Theta}^{(0)}$.}
        \STATE{$\Theta_0^{(0)}\leftarrow \gF^{(0)}(\hat{\Theta}_0^{(0)})$}
        \FOR{$s=1,2,\dots, s_{\rm max}$}
            \STATE{$\Theta_s^{(0)}\leftarrow \gF^{(0)}(\hat{\Theta}_{s-1}^{(0)})$} 
            \STATE{$\hat{\Theta}_s^{(0)}\leftarrow (1-\eta_{\rm d})\hat{\Theta}_{s-1}^{(0)} + \eta_{\rm d}\hat{\gF}^{(0)}(\Theta_{s-1}^{(0)})$} 
            \IF{$\max{|\Theta_s^{(0)}- \Theta_{s-1}^{(0)}|} < \epsilon_{\rm tol}$}
                \STATE{$s\leftarrow s_{\rm max}$}
                \STATE{break}
            \ENDIF
        \ENDFOR
    \STATE{// solving for $t>1$}
        \FOR{$t=1,\dots,T$}
            \STATE{$(\Theta_{0}^{(t)}, \hat{\Theta}_0^{(t)})\leftarrow(\Theta_{s_{\rm max}}^{(t-1)}, \hat{\Theta}_{s_{\rm max}}^{(t-1)})$}
            \FOR{$s=1,2,\dots, s_{\rm max}$}
                \STATE{$\Theta_s^{(t)}\leftarrow \gF^{(t)}({\Theta}_{s-1}^{(t-1)}, \hat{\Theta}_{s-1}^{(t)})$} 
                \STATE{$\hat{\Theta}_s^{(t)}\leftarrow (1-\eta_{\rm d})\hat{\Theta}_{s-1}^{(t)} + \eta_{\rm d}\hat{\gF}^{(t)}(\Theta_{s-1}^{(t)})$} 
                \IF{$\max{|\Theta_s^{(t)}- \Theta_{s-1}^{(t)}|} < \epsilon_{\rm tol}$}
                    \STATE{$s\leftarrow s_{\rm max}$}
                    \STATE{break}
                \ENDIF
            \ENDFOR
        \ENDFOR
    \end{algorithmic}
\end{algorithm}

\subsection{Solving the self-consistent equations}
To solve the self-consistent equations in Definition \ref{def: self-consistent equations}, the fixed-point iteration algorithm are used. Since the self-consistent equations are already written in a form of $(\Theta, \hat{\Theta}) = \gF(\Theta, \hat{\Theta})$, the definition of the fixed-point iteration is straightforward. In particular, $(\Theta^{(t)}, \hat{\Theta}^{(t)})$ only depends on the parameters at $t$ and $t-1$, the equations are solved successively. Let us define the right-hand side of the self-consistent equations at step $t$ as $\Theta^{(t)} = \gF^{(t)}(\Theta^{(t-1)}, \hat{\Theta}^{(t)})$ and $\hat{\Theta}^{(t)}=\hat{\gF}^{(t)}(\Theta^{(t)})$. Then our fixed-point iteration can be summarized as in Algorithm \ref{algo: fixed-point iteration}. For evaluating the integral in updating $\hat{\Theta}^{(t)}$,  the Gauss-Hermite quadrature is used when $\Gamma=0$ and the Monte-Carlo integral is used in other cases. In Monte-Carlo integral, the samples of size $10^5-10^7$ are used depending on the cases. For solving the nonlinear equations in \eqref{appeq: effective problem u 0} and \eqref{appeq: effective problem u t}, the Newton method is used. Note that the derivative in \eqref{eq:rs-saddle-hatr} includes the derivative of the indicator function $\1(|\tilde{h}_{u}^{(t)}|>\Gamma\sqrt{q^{(t-1)}})$, which yields the delta functions at $\tilde{h}_{u}^{(t)}=\pm\Gamma\sqrt{q^{(t-1)}}$. Although including this contribution by Gauss-Hermite or Monte-Carlo integral is difficult, we can easily include it by hand because in only requires the values of the integrand at two points, i.e, the boundaries of the indicator.

We remark that the fixed-point iteration of the self-consistent equations of the type considered above may be closely related to the efficient approximate inference algorithms. Indeed, when considering a simple empirical risk minimization or the Bayesian inference in linear models, it corresponds to the state-evolution formula of the approximate message passing (AMP) algorithm \citep{zdeborova2016, takahashi2020, takahashi2022}, which is an efficient algorithm with fast convergence. Hence we expect that the above apparently naive fixed-point iteration yields a moderately good convergence.

\subsection{Optimization of the regularization parameter}
Since the generalization error can be numerically evaluated for each value of the hyper parameters by solving the self-consistent equation using Algorithm \ref{algo: fixed-point iteration}, we can treat the generalization error \eqref{eq: rs generalization error} as a function of the hyper parameters. However, the explicit form of such dependence is unknown. Thus, to optimize the generalization error, the Nelder-Mead algorithm, which is a black-box optimization algorithm, implemented in Julia language \citep{mogensen2018optim} is used. By solving the self-consistent equation at each step of optimization, it can be straightforwardly implemented.

\section{Pathological finite size effect at large regularization}
\label{app: pathological finite size effect}
As already reported in the previous literature \citep{Dobribal2018high, mignacco2020role}, the ridge-regularized logistic regression with true labels yields the Bayes-optimal classifier when $\rho=1/2$ and the infinitely large regularization parameter is used. However, this result is valid only in the asymptotic regime, and it is known that the experimental results of finite-size systems significantly deviate from theoretical predictions as reported in \citep{mignacco2020role}. This is because the norm of the weight vector becomes extremely small for large regularization parameters, making the numerical computation challenging and even small fluctuations can lead to catastrophic results. Therefore, predictions from sharp asymptotics may be irrelevant to practical results if we unbound the regularization parameters in our ST as well. 

To check this point, we conducted preliminary experiments. For simplicity, we consider the single-shot case with $T=1$. In this case, we can find the optimal regularization parameters $(\lambda_\ast^{(0)}, \lambda_\ast^{(1)})=(\lambda_L^\ast, \lambda_U^\ast)$ that minimize the generalization error $\epsilon_{\rm g}^{(1)}$ by a brute force optimization since numerically evaluating the generalizaeion. We optimize the regularization parameters in $(0,10^4]^2$ and observed the behavior of the optimal regularization parameters and the generalization error $\epsilon_{\rm g}^{(1)}$. Here $\Delta_U=\Delta_L=\Delta, \rho_U=\rho_L=\rho$.

\begin{figure}[t!]
     \centering
     \begin{subfigure}[b]{0.3\textwidth}
         \centering
         \includegraphics[width=\textwidth]{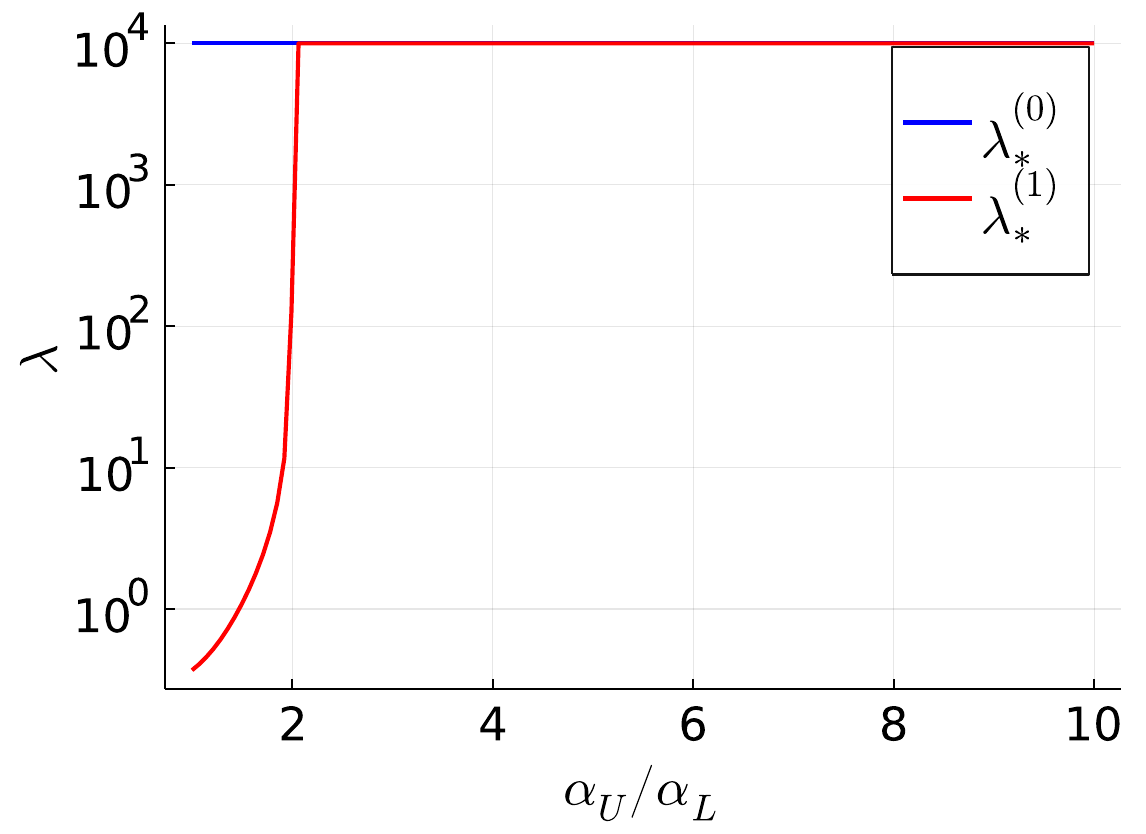}
         \caption{$(\rho, \Delta) = (0.5, 1.0)$}
     \end{subfigure}
     \hfill
     \begin{subfigure}[b]{0.3\textwidth}
         \centering
         \includegraphics[width=\textwidth]{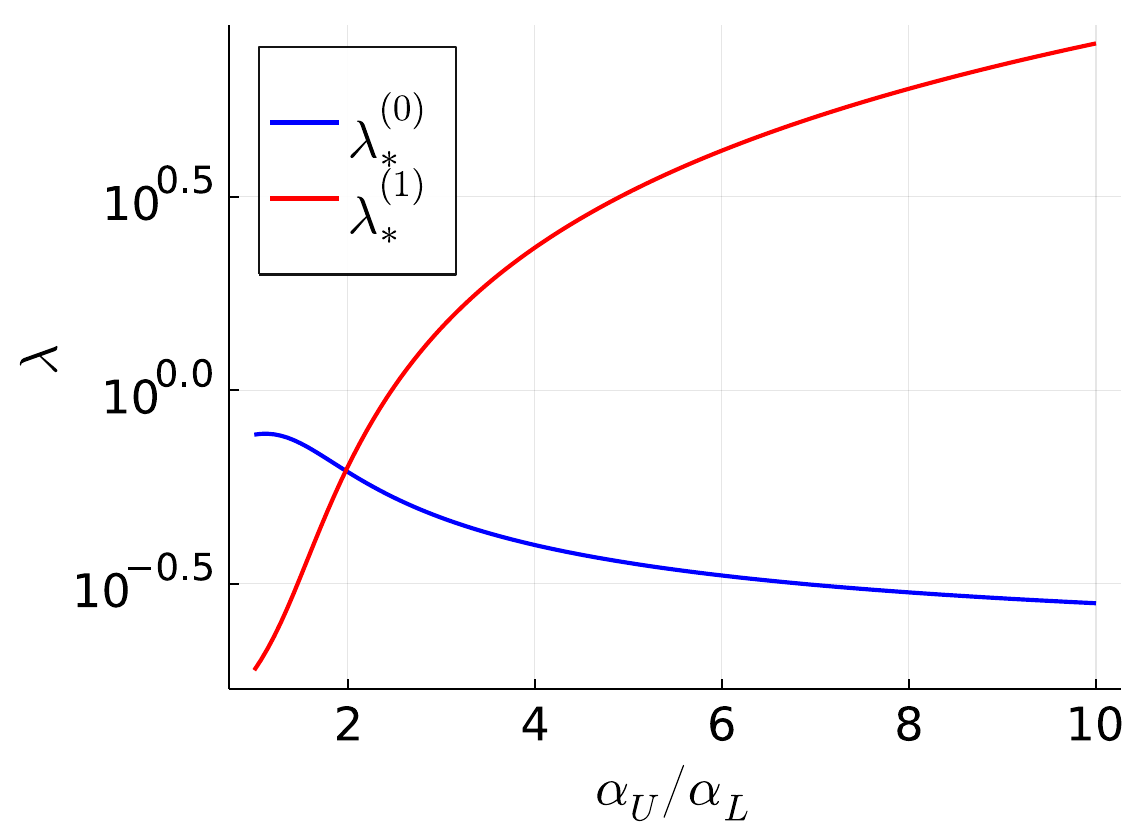}
         \caption{$(\rho, \Delta) = (0.495, 1.0)$}
     \end{subfigure}
     \hfill
     \begin{subfigure}[b]{0.3\textwidth}
         \centering
         \includegraphics[width=\textwidth]{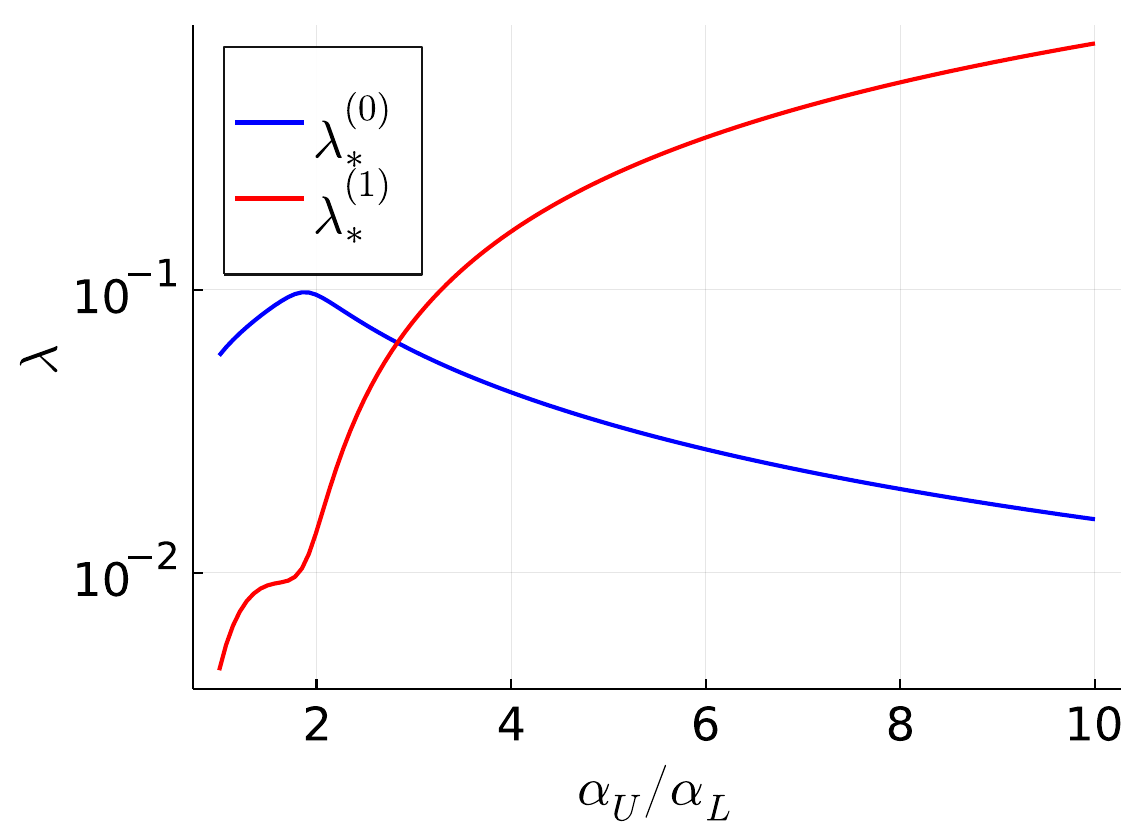}
         \caption{$(\rho, \Delta) = (0.4, 1.0)$}
     \end{subfigure}
     \hfill
     \begin{subfigure}[b]{0.3\textwidth}
         \centering
         \includegraphics[width=\textwidth]{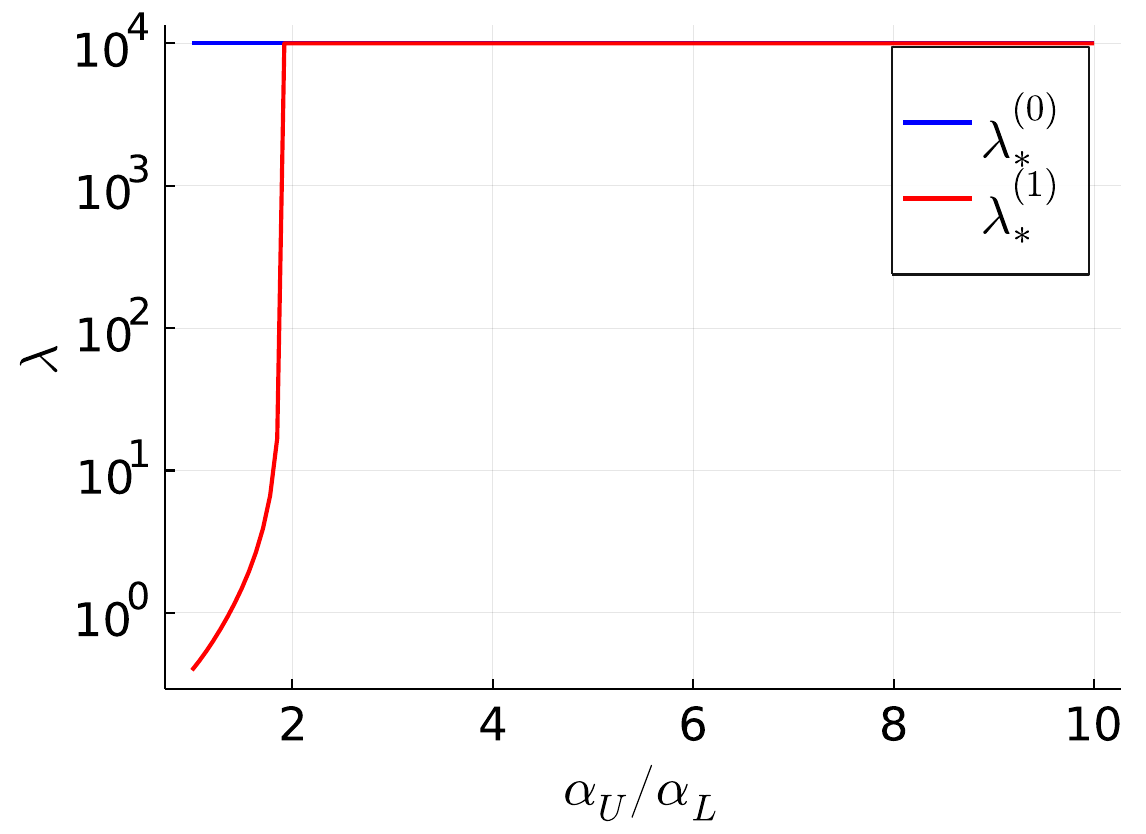}
         \caption{$(\rho, \Delta) = (0.5, 0.75^2)$}
     \end{subfigure}
     \hfill
     \begin{subfigure}[b]{0.3\textwidth}
         \centering
         \includegraphics[width=\textwidth]{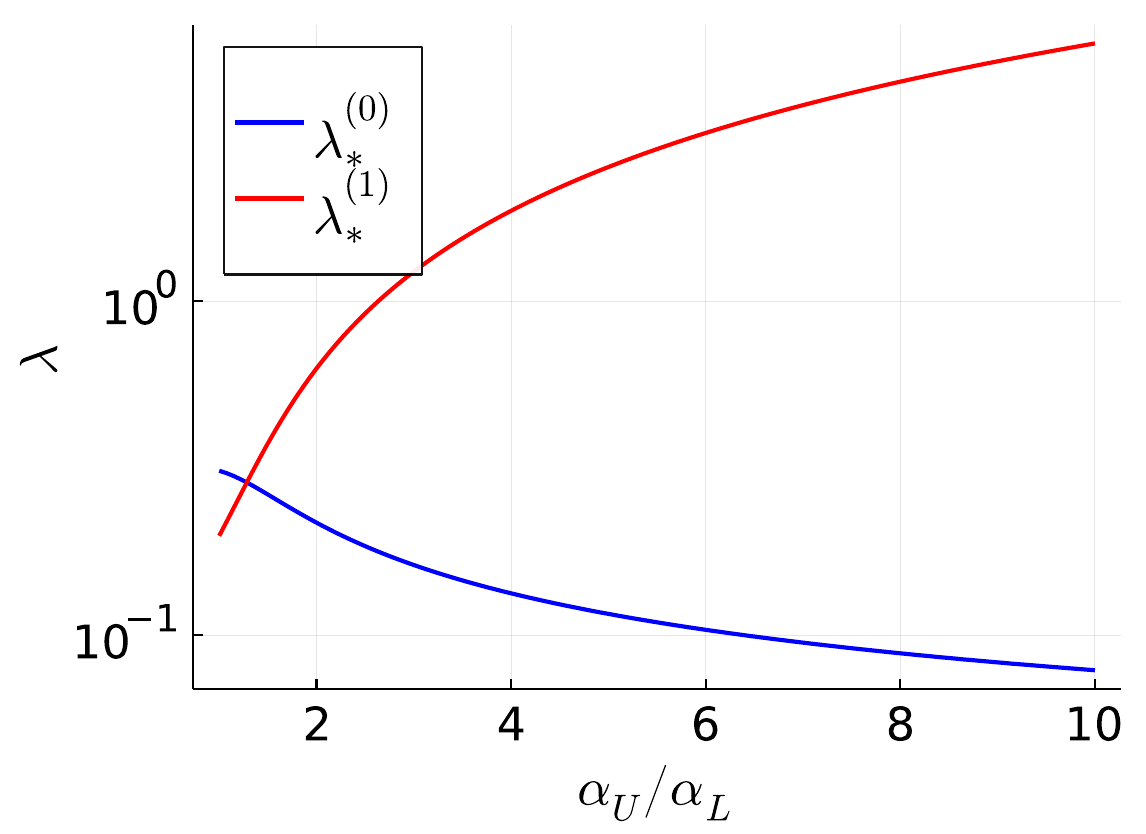}
         \caption{$(\rho, \Delta) = (0.495, 0.75^2)$}
     \end{subfigure}
     \hfill
     \begin{subfigure}[b]{0.3\textwidth}
         \centering
         \includegraphics[width=\textwidth]{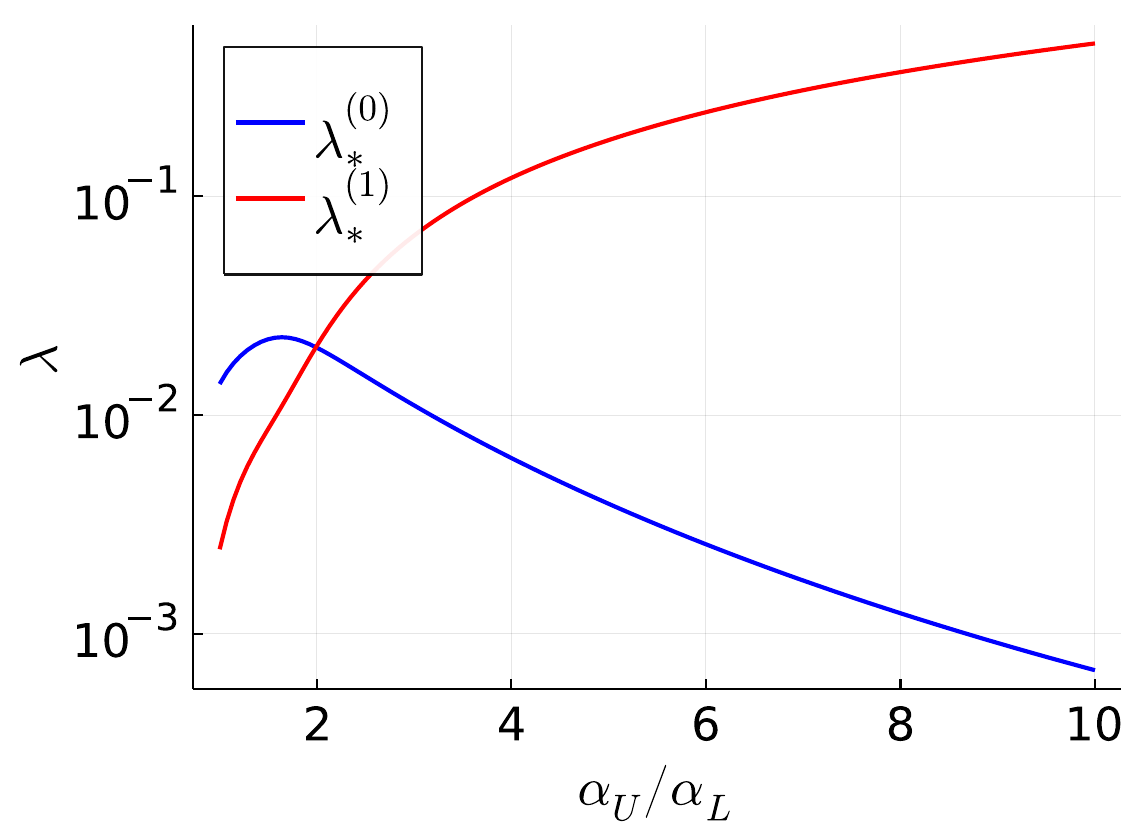}
         \caption{$(\rho, \Delta) = (0.4, 0.75^2)$}
     \end{subfigure}
    \caption{The optimal regularization parameters $(\lambda^{(0)}_\ast, \lambda^{(1)}_\ast)$, which minimize the generalization error $\epsilon_{\rm g}^{(1)}$, are plotted against the relative number of unlabeled data points $\alpha_U/\alpha_L$. Each panel corresponds to the different values of $\rho$ and $\Delta$.}
    \label{fig: optimal lambda alphaU dependence single shot}
\end{figure}

\begin{figure}[t!]
    \centering
    \begin{subfigure}[b]{0.45\textwidth}
        \centering
        \includegraphics[width=\textwidth]{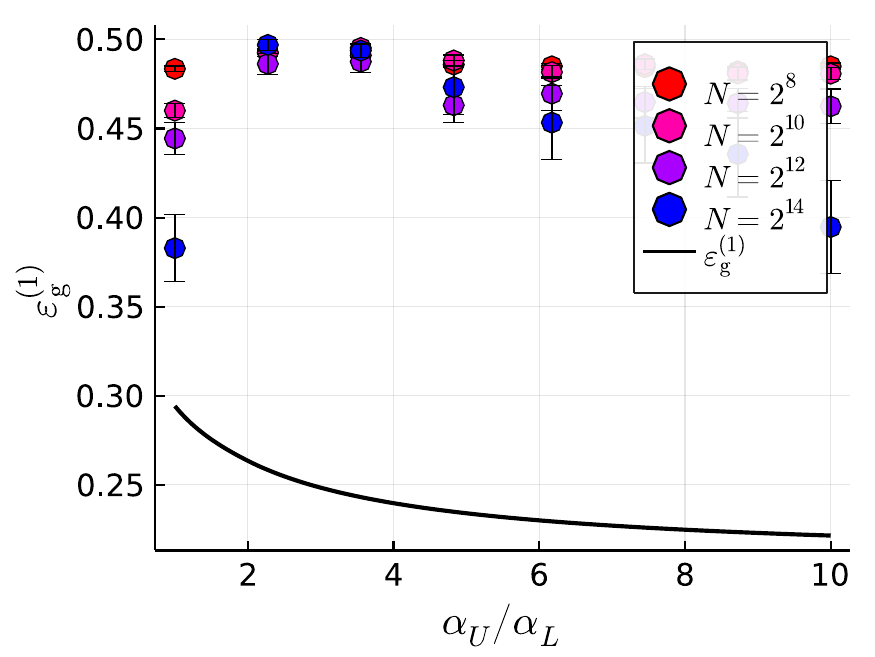}
        \caption{$(\rho, \Delta) =(0.5,1.0)$}
    \end{subfigure}
    \hfill
    \begin{subfigure}[b]{0.45\textwidth}
        \centering
        \includegraphics[width=\textwidth]{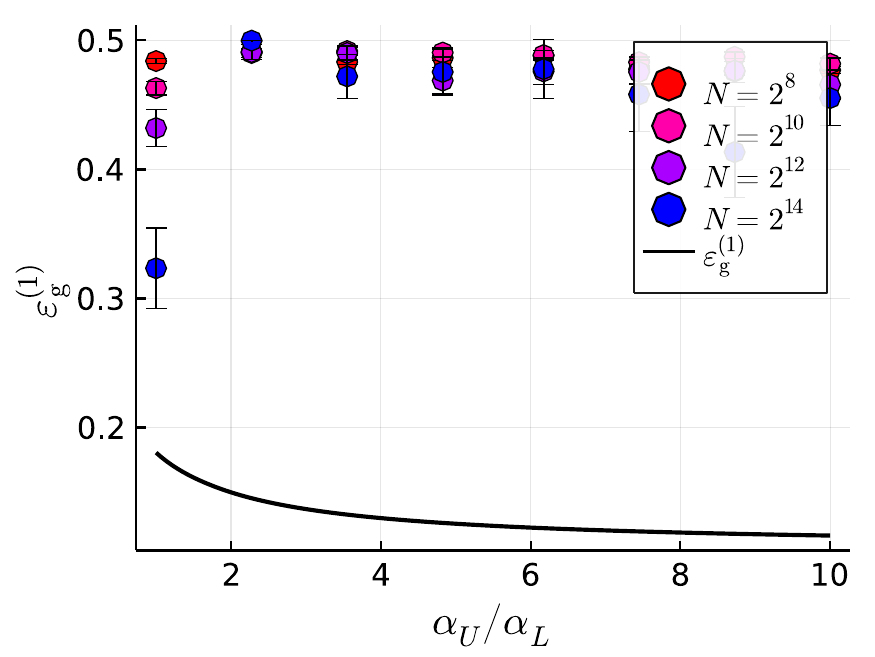}
        \caption{$(\rho, \Delta)=(0.5,0.75^2)$}
    \end{subfigure}
    \caption{Theoretical estimate for the optimal generalization error $\epsilon_{\rm g}^{(1)}$ for $\rho=0.5$ is compared with the experiments. The black solid line is the theoretical estimate in LSL, and the symbols represent the experimental results. Here the regularization parameter is limited as $(\lambda^{(0)},\lambda^{(1)})\in(0,10)^2$. For moderately large $\alpha_U$, $(\lambda^{(0)}, \lambda^{(1)})=(10,10)$, however, for such large regularization parameters, the experimental results heavily suffer from large finite-size effects.}
    \label{fig: large regularization single-shot ST}
\end{figure}

Figure \ref{fig: optimal lambda alphaU dependence single shot} summarizes how the optimal regularization parameters $(\lambda^{(0)}_\ast, \lambda^{(1)}_\ast)$, which minimize the generalization error $\epsilon_{\rm g}^{(1)}$, depend on the size of the unlabeled dataset $\alpha_U$, the relative size of clusters $\rho$ and the size of each cluster $\Delta$. Each panel corresponds to different values of $\rho$ and $\Delta$.  When $\rho=1/2$, both $\lambda_\ast^{(0)}$ and $\lambda_\ast^{(1)}$ show diverging tendency as expected. Figure \ref{fig: large regularization single-shot ST} shows the comparison with the experiments of the regularization parameter found in the above procedure. Similarly to the previous studies, the experimental results severely deviates from the prediction of the asymptotic result. Hence, we need to limit the range of the regularization parameters as in the main text.

\vskip 0.2in
\bibliography{references}

\end{document}